\documentclass[journal]{IEEEtran}
\usepackage{graphicx}
\usepackage{amsmath,amssymb} 
\usepackage{color}
\usepackage{subfig}
\usepackage{graphicx}
\usepackage{multirow}
\usepackage{multicol}
\usepackage{url}
\usepackage{amsmath}
\usepackage{amssymb}
\usepackage{mathrsfs}
\usepackage{bm}
\newcommand{\etal}{\textit{et al.~}}
\def\hspacefigure{\hspace{-0.8mm}}
\def\widtheight{0.13\linewidth}
\def\widtheighttwo{0.23\linewidth}
\def\widthnine{0.092\linewidth}
\def\widthtwelve{0.086\linewidth}
\def\widthNne{0.086\linewidth}
\usepackage{array}
\newcolumntype{I}{!{\vrule width 1.12pt}}

\ifCLASSINFOpdf
\else
\fi

\usepackage[bookmarks=false]{hyperref}

\usepackage{amsmath}

\usepackage{algorithmic}

\begin{document}
%
\title{RGBD Salient Object Detection via Deep Fusion}
%
%
%

\author{Liangqiong~Qu,
        Shengfeng~He,
        Jiawei~Zhang,
        Jiandong~Tian,
        Yandong~Tang,
        and~Qingxiong~Yang
        \\{\url{http://www.cs.cityu.edu.hk/~jiawzhang8/saliency/TIP_saliency.htm}} 
\thanks{L. Qu, S. He, J. Zhang, and Y. Qing are with the Department of Computer Science, City University of Hong Kong, Hong Kong. L. Qu
is also with the State Key Laboratory of Robotics, Shenyang Institute of Automation, Chinese Academy of Sciences, Shenyang, 110016, and the University of Chinese Academy of Sciences, Beijing, China, 100049. (E-mail: quliangqiong@sia.cn; shengfeng\_he@yahoo.com; jiawzhang8-c@my.cityu.edu.hk; qiyang@cityu.edu.hk).}
\thanks{J. Tian and Y. Tang are with the State Key Laboratory of Robotics,
Shenyang Institute of Automation, Chinese Academy of Sciences, Shenyang, 110016 (E-mail:
tianjd@sia.cn; ytang@sia.cn).}}

%
%

\markboth{Journal of \LaTeX\ Class Files,~Vol.~14, No.~8, August~2015}%
{Shell \MakeLowercase{\textit{et al.}}: Bare Demo of IEEEtran.cls for IEEE Journals}
%



\maketitle

\begin{abstract}
Numerous efforts have been made to design different low level saliency cues for the RGBD saliency detection, such as color or depth contrast features, background and color compactness priors. However, how these saliency cues interact with each other and how to incorporate these low level saliency cues effectively to generate a master saliency map remain a challenging problem.
In this paper, we design a new convolutional neural network (CNN) to fuse different low level saliency cues into hierarchical features for automatically detecting salient objects in RGBD images. In contrast to the existing works that directly feed raw image pixels to the CNN, the proposed method takes advantage of the knowledge in traditional saliency detection by adopting various meaningful and well-designed saliency feature vectors as input. This can guide the training of CNN towards detecting salient object more effectively due to the reduced learning ambiguity. We then integrate a Laplacian propagation framework with the learned CNN to extract a spatially consistent saliency map by exploiting the intrinsic structure of the input image. Extensive quantitative and qualitative experimental evaluations on three datasets demonstrate that the proposed method consistently outperforms state-of-the-art methods.
\end{abstract}

\begin{IEEEkeywords}
RGBD saliency detection, Convolutional neural network, Laplacian propagation.
\end{IEEEkeywords}

%
\IEEEpeerreviewmaketitle

\section{Introduction}
%
%
%
%
\IEEEPARstart{S}ALIENCY detection, which is to predict where human looks in the image,
has attracted a lot of research interests in recent years. It serves as an important pre-processing step
in many problems such as image classification, image retargeting and
object recognition \cite{rutishauser2004bottom,mahadevan2009saliency,sharma2012discriminative,luo2014switchable}.
Unlike RGB saliency detection which receives much research attention, there are not many exploration on RGBD cases. The recently emerged sensing technologies, such as Time-of-flight sensor and Microsoft Kinect, provides excellent ability and flexibility to capture RGBD image \cite{zhang2012microsoft,gokturk2004time}.
Detecting RGBD saliency becomes essential for many applications such as 3D content surveillance, retrieval, and image recognition \cite{mishra2012segmenting,fu2015object,banica2015second}. In this paper, we focus on how to integrate RGB and the additional depth information for RGBD saliency detection.

\begin{figure*}
\centering
\captionsetup[subfigure]{labelformat=empty}
\subfloat[(a)]{ \label{fig:Intro:a}\includegraphics[width=\widthnine]{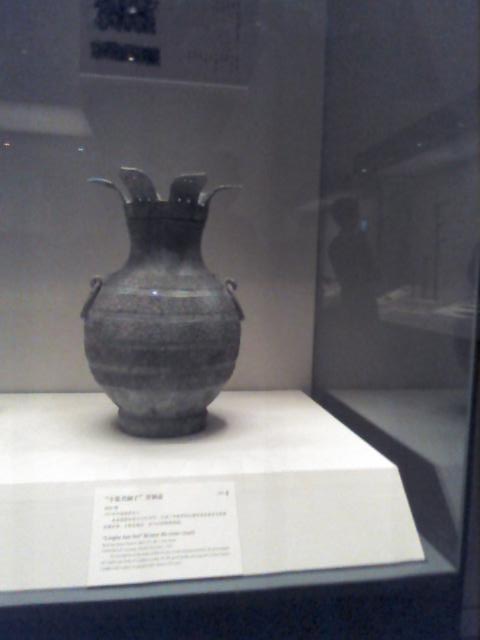}} \hspacefigure
\subfloat[(b)]{ \label{fig:Intro:b}\includegraphics[width=\widthnine]{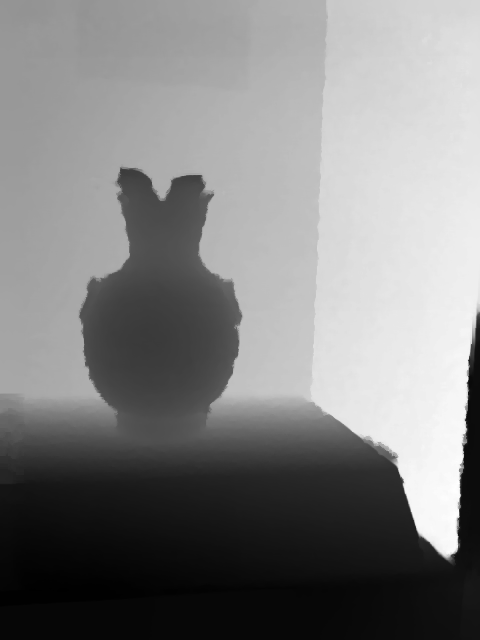}}\hspacefigure
\subfloat[(c)]{ \label{fig:Intro:c}\includegraphics[width=\widthnine]{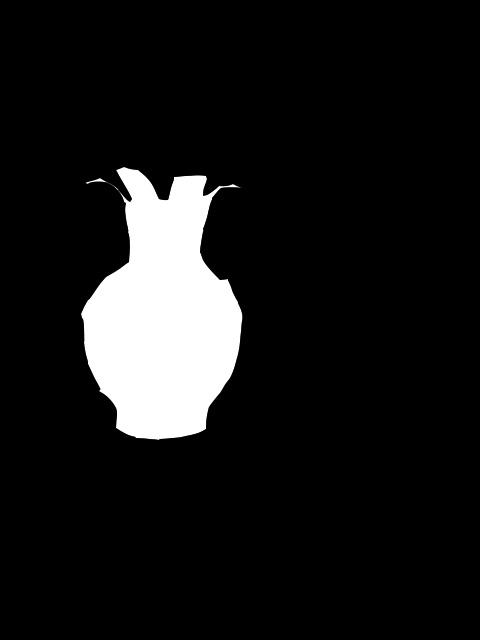}}\hspacefigure
\subfloat[(d)]{ \label{fig:Intro:d}\includegraphics[width=\widthnine]{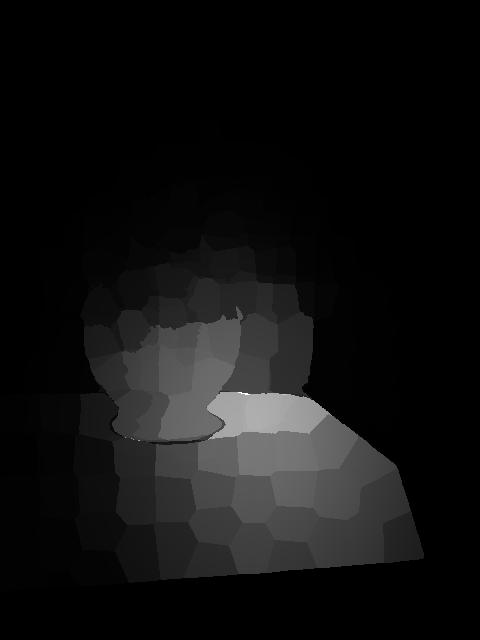}} \hspacefigure
\subfloat[(e)]{ \label{fig:Intro:e}\includegraphics[width=\widthnine]{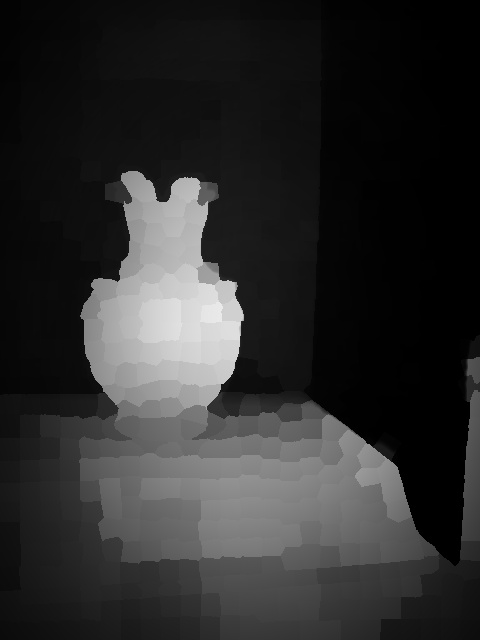}} \hspacefigure
\subfloat[(f)]{ \label{fig:Intro:f}\includegraphics[width=\widthnine]{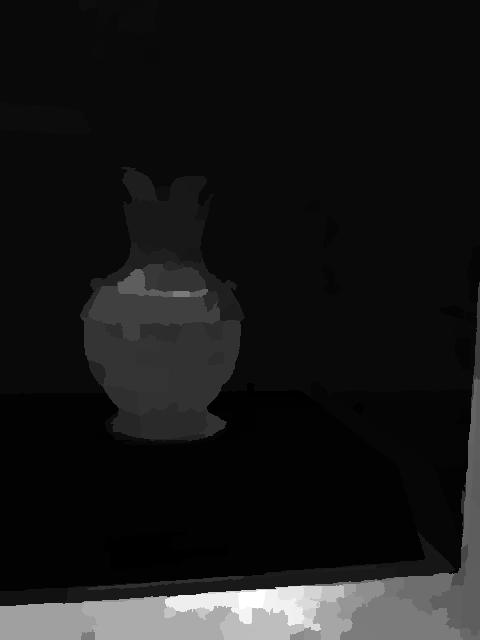}} \hspacefigure
\subfloat[(g)]{ \label{fig:Intro:g}\includegraphics[width=\widthnine]{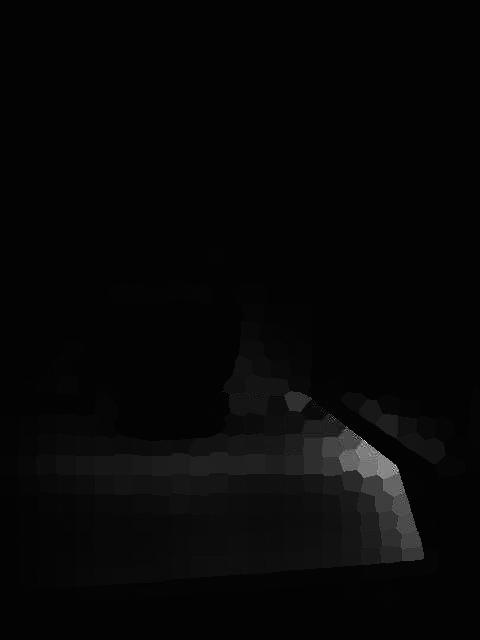}} \hspacefigure
\subfloat[(h)]{ \label{fig:Intro:h}\includegraphics[width=\widthnine]{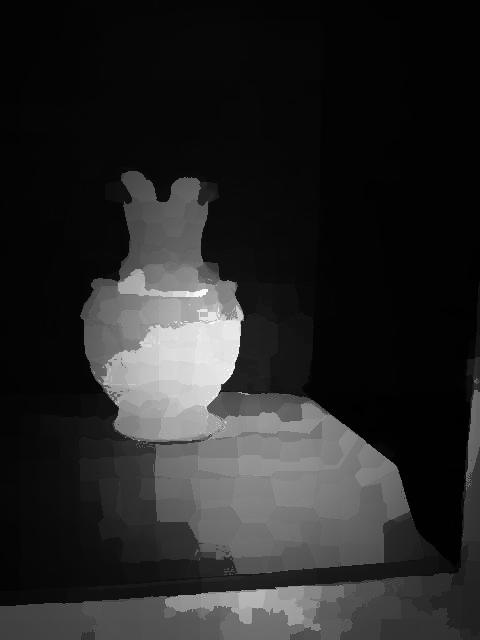}}\hspacefigure
\subfloat[(i)]{ \label{fig:Intro:i}\includegraphics[width=\widthnine]{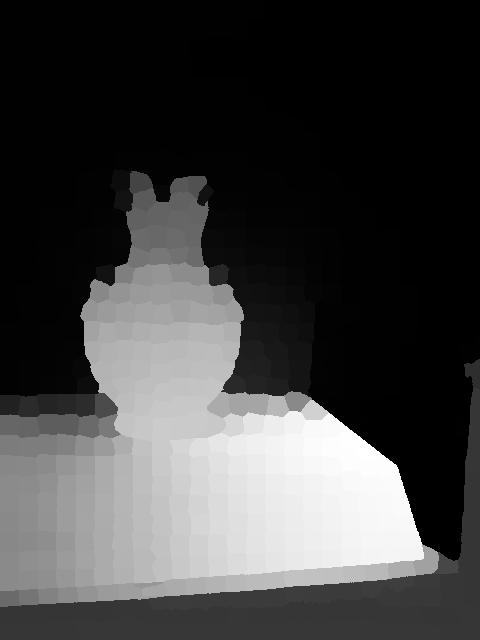}}\hspacefigure
\subfloat[(j)]{ \label{fig:Intro:j}\includegraphics[width=\widthnine]{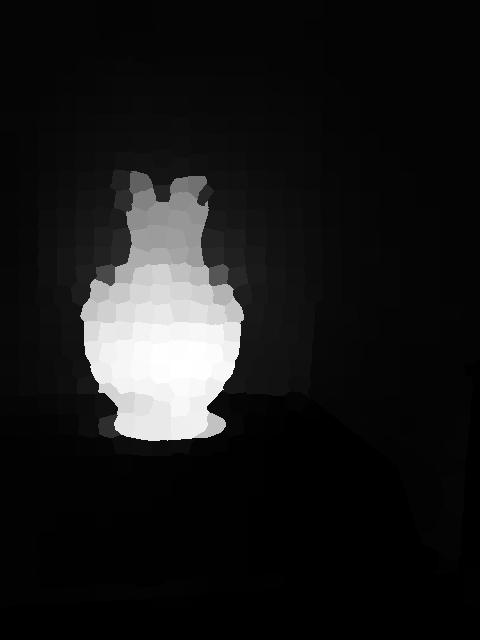}} \hspacefigure\\
\vspace{-1.5mm}
  \caption{Example to show the problem of saliency map merging methods.
  (a) Original RGB image. (b) Original depth image. (c) Ground truth saliency map.
  (d) Saliency map by LMH \cite{peng2014rgbd}. (e) Saliency map by ACSD \cite{ju2014depth}. (f) Saliency map by GP \cite{ren2015exploiting}. (g) to (i) are the saliency map integration results of (d), (e), and (f). (g) Linear combination (i.e., averaging). (h) MCA integration \cite{qin2015saliency}. (i) CNN based fusion. (j) Saliency map by the proposed hyper-feature fusion.}\vspace{-1.5mm}
\label{fig:Intro}
\vspace{-2mm}
\end{figure*}

According to how saliency is defined, saliency detection methods can be classified into two categories: top-down approach and bottom-up approach \cite{itti1998model,ma2003contrast}. Top-down saliency detection is task-dependent that incorporates high level features to locate the salient object. On the other hand, bottom-up approach is task-free, and it utilizes low level features that are biologically motivated to estimate salient regions. Most of the existing bottom-up saliency detection methods focus on designing different low-level cues to represent salient objects. The saliency maps of these low-level features are then fused to become a master saliency map. As human attention are preferentially attracted by the high contrast regions with their surrounding, contrast-based features (like the color, edge orientation or texture contrasts) make a crucial role to derive the salient objects. Background \cite{wei2012geodesic} and color compactness priors~\cite{keyang2013} consider salient object in different perspectives. The first one leverages the fact that most of the salient objects are far from image boundaries, the latter one utilizes the color compactness of the salient object. In addition to RGB information, depth has been shown to be one of the practical cue to extract saliency \cite{maki1996computational,lang2012depth,desingh2013depth,zhang2010stereoscopic}. Most existing approaches for 3D saliency detection either treat the depth map as an indicator to weight the RGB saliency map \cite{maki1996computational,zhang2010stereoscopic} or consider depth cues as an independent image channel \cite{lang2012depth,desingh2013depth}.

Notwithstanding the demonstrated success of these features, whether these features complement to each other remains a question. The interaction mechanism of different saliency features is not well explored, and it is not clear how to integrate 2D saliency features with depth-induced saliency feature in a better way. Linearly combining the saliency maps produced by these features cannot guarantee better result (as shown in Figure~\ref{fig:Intro:g}). Some other more complex combination algorithms have been proposed in \cite{liu2011learning,yan2013hierarchical,qin2015saliency,zhou2015salient,peng2014rgbd,ren2015exploiting}.
Qin \emph{et al.} \cite{qin2015saliency} propose a Multi-layer Cellular Automata (MCA, a Bayesian framework) to merge different saliency maps by taking advantage of the superiority of each saliency detection methods. Recently, several heuristic algorithms are designed to combine the 2D related saliency maps and depth-induced saliency map \cite{peng2014rgbd,ren2015exploiting}.
However, as restricted by the computed saliency values, these saliency map combination methods are not able to correct wrongly estimated salient regions. For example in Figure \ref{fig:Intro}, heuristic based algorithms (Figure \ref{fig:Intro:d} to \ref{fig:Intro:f}) cannot detect the salient object correctly. Adopting these saliency maps for further fusion, neither simple linear fusion (Figure \ref{fig:Intro:g}) nor MCA integration (Figure \ref{fig:Intro:h}) are able to recover the salient object. We wonder whether a good integration can address this problem by further adopting Convolutional Neural Network technique to train a saliency map integration model. The resulted image shown in Figure \ref{fig:Intro:i} indicates that saliency map integration is hugely influenced by the quality of the input saliency maps. Based on the these observations, we take one step back to handle more raw and flexible saliency features.

In this paper, we propose a deep fusion framework for RGBD saliency detection. The proposed method takes advantage of the representation learning power of CNN to extract the hyper-feature by fusing different hand-designed saliency features to detect salient object (as shown in Figure~\ref{fig:Intro:j}). We first compute several feature vectors from original RGBD image, which include local and global contrast, background prior, and color compactness. We then propose a CNN architecture to incorporate these regional feature vectors into a more representative and unified features. Compared with feeding raw image pixels, these extracted saliency features are well-designed and they can guide the learning of CNN towards saliency-optimized more effectively. As the resulted saliency map may suffer from local inconsistency and noisy false positive, we further integrate a Laplacian propagation framework with the proposed CNN. This approach propagates high confidence saliency to the other regions by taking account of the color and depth consistency and the intrinsic structure of the input image \cite{zhou2004learning}, which is able to remove noisy values and produce smooth saliency map. The Laplacian propagation is solved with fast convergence by the adoption of Conjugate gradient and preconditioner. Experimental evaluations demonstrate that, once our deep fusion framework are properly trained, it generalizes well to different datasets without any additional training and outperforms the state-of-the-art approaches.

The main contributions of this paper are summarized as follows.

1. We propose a simple yet effective deep learning model to explore the interaction mechanism of RGB and depth-induced saliency features for RGBD saliency detection. This deep model is able to generate representative and discriminative hyper-features automatically rather than hand-designing heuristical features for saliency. 

2. We adopt Laplacian Propagation to refine the resulted saliency map and solve it with fast convergence.
Different from CRF model, our Laplacian Propagation not only considers the spatial consistency but also exploits the intrinsic structure of the input image \cite{zhou2004learning}. Extensive experiments further demonstrate that this proposed Laplacian Propagation is able to refine the saliency maps of existing approaches, which can be widely adopted as a post processing step.

3. We investigate the limitations of saliency map integration, and demonstrate that simple features fusion are able to obtain superior performance.
\section{Related work}
\begin{figure*}
\centering
\includegraphics[width=0.99\linewidth]{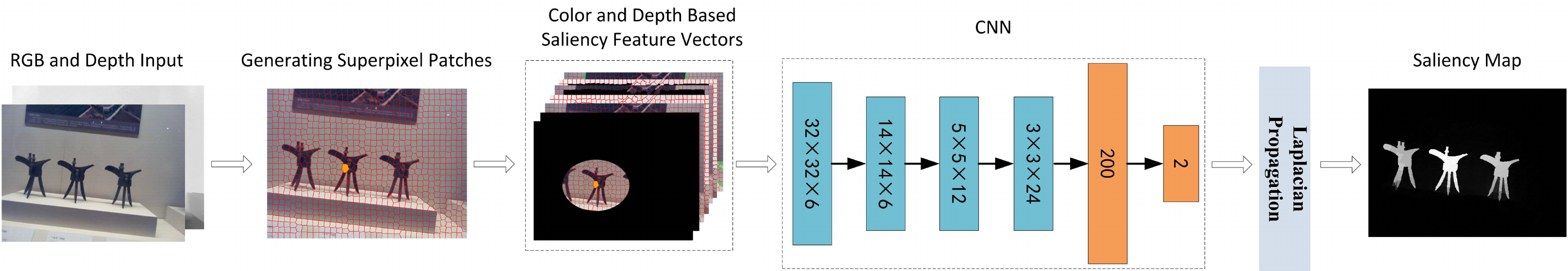}
\vspace{-1.0mm}
   \caption{The pipeline of the proposed method. Our method composes of three modules. First, it generates different RGB and depth based saliency feature vectors from the RGBD input image. These generated saliency feature vectors are then fed to the CNN. The CNN takes an input of size $32\times32\times6$ and generates the saliency confidence value (the probability of this patch belonging to salient). Finally, a Laplacian propagation is performed on the resulted probabilities to extract the final spatially consistent saliency map.
  }
\label{fig:Convnet}
\vspace{-1.5mm}
\end{figure*}

In this section, we give a brief survey and review of RGB and RGBD saliency detection methods, respectively. Comprehensive literature reviews on these saliency detection
methods can be found in ~\cite{borji2014salient,peng2014rgbd}.

\textbf{RGB saliency detection:} As suggested by the studies of cognitive science~\cite{itti1998model}, bottom-up saliency is driven by low-level stimulus features. This concept is also adopted in computer vision to model saliency. Contrast-based cues, especially color contrast, are the most widely adopted features in previous works. These contrast-based methods can be roughly classified into two categories: local and global approaches. Local method calculates color, edge orientation or texture contrast of a pixel/region with respect to a local window to measure saliency \cite{itti1998model,bruce2005saliency}. In \cite{itti1998model}, they develop an early local based visual saliency detection method by computing center surrounding differences across multi-scale image features to estimate saliency. Bruce \emph{et al.} \cite{bruce2005saliency} propose to apply sparse representation on local image patches. However, based only on local contrast, these methods may highlight the boundaries of salient object \cite{keyang2013} and be sensitive to high frequency content \cite{shenfeng2014}. In contrast to local approach, the global approach measures salient region by estimating the contrast over the entire image. Achanta \etal \cite{achanta2009frequency} model saliency by computing color difference to the mean image color. Cheng \etal \cite{ChengPAMI} propose a histogram-based global contrast saliency method by considering the spatial weighted coherence. Although these global methods achieve superior performances, they may suffer from distractions when background shares similar color to the salient object. Background and color compactness priors are proposed as a complement to contrast-based methods \cite{wei2012geodesic,shen2012unified,keyang2013}. These methods are built on strong assumptions, which may invalid in some scenarios.

As each feature has different strengths, some works focus on designing the integration mechanism for different saliency features \cite{liu2011learning,shen2012unified,yan2013hierarchical,jiang2013salient}. Liu \etal \cite{liu2011learning} use CRF to integrate three different features from both local and global point of views. Yan \etal \cite{yan2013hierarchical} propose a hierarchical framework to integrate saliency maps in different scales, which can handle small high contrast regions well. Unlike these methods that directly combine the saliency maps obtained from different saliency cues, the proposed method records low-level saliency feature in vector forms and jointly learns the interaction mechanism to become a hyper-feature with CNN.

Similar to the proposed method, CNN has been adopted in some other works to extract hierarchical feature representations for detecting salient regions \cite{vig2014large,zhao2015saliency,li2015visual,he2015supercnn,wang2015deep,LiYu16}. In contrast to most of these deep networks that take raw image pixels as input, the proposed method aims at designing a unified CNN framework to learn the interaction mechanism of different saliency cues.

\textbf{RGBD saliency detection:} Unlike RGB saliency detection, RGBD saliency receives less research attention \cite{maki1996computational,zhang2010stereoscopic,desingh2013depth,lang2012depth,wang2013computational}. Maki \etal \cite{maki1996computational} propose an early computational model on depth-based attention by measuring disparity, flow and motion. Similar to color contrast, Zhang \etal design a stereoscopic visual attention algorithm based on depth and motion contrast for 3D video \cite{zhang2010stereoscopic}. Desingh \etal \cite{desingh2013depth} estimate saliency regions by fusing the saliency maps produced by appearance and depth cues independently. These methods either treat the depth map as an indicator to weight the RGB saliency map \cite{maki1996computational,zhang2010stereoscopic} or consider depth map as an independent
image channel for saliency detection \cite{lang2012depth,desingh2013depth}. On the other hand, Peng \etal \cite{peng2014rgbd} propose a multi-stage RGBD model to combine both depth and appearance cues to detect saliency. Ren \etal \cite{ren2015exploiting} integrate the normalized depth prior and the surface orientation prior with RGB saliency cues directly for the RGBD saliency detection. These methods combine the depth-induced saliency map with RGB saliency map either directly  \cite{ju2014depth,ren2015exploiting} or in a hierarchy way to calculate the final RGBD saliency map \cite{peng2014rgbd}. However, these saliency map level integration is not optimal as it is restricted by the determined saliency values. On the contrary, we incorporate different saliency cues and fuse them with CNN in feature level.


\section{Proposed method}
As shown in Figure \ref{fig:Convnet}, the proposed deep fusion framework for RGBD salient object detection composes of three modules. The first module generates various saliency feature vectors for each superpixel region. The second module is to extract hyper-feature representation from the obtained saliency feature vectors. The third module is the Laplacian propagation framework which helps to detect a spatially consistent saliency map.

\subsection{Saliency feature vectors extraction}\label{title_3_1}
\begin{figure*}[t]
\centering
\includegraphics[width=0.99\linewidth]{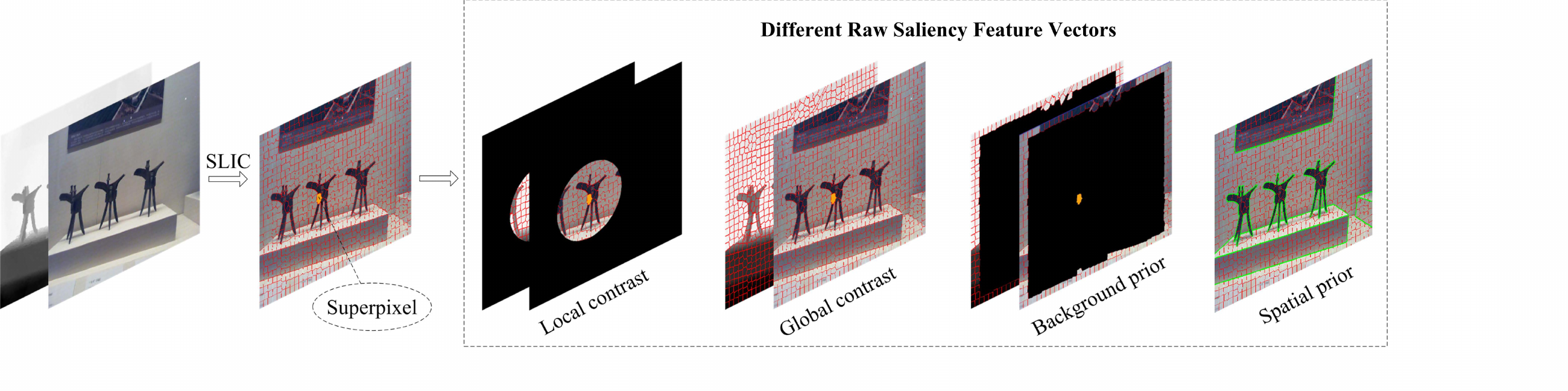}

   \caption{Saliency feature extraction.}
\label{fig:saliency_cue}
\vspace{-1.0mm}
\end{figure*}

Given an image, we aim to represent saliency by some demonstrated effective saliency features.
Figure \ref{fig:saliency_cue} gives an illustration on the proposed saliency feature extraction. We first segment the image into \emph{N} superpixels using SLIC method \cite{achanta2012slic}. Given a RGB image $\mathcal{I}$, we denote the segmented \emph{N} regions as $\mathcal{P}  = \{ {P_1},{P_2},...,{P_i},...{P_N}\} $. For each superpixel ${P_i}$, we denote the calculated saliency features as a vector ${\Gamma _{{P_i}}}$.
In the following, we will take region $P_i$ (the region that marked in orange in Figure \ref{fig:saliency_cue}) as an example to show how we calculate different saliency feature vectors.

Different from the classical saliency detection methods that directly calculate the saliency values for each superpixel, we record the saliency features for each image region and no further operation is performed to make saliency features as raw as possible. For region $P_i$, there are seven types of feature vectors: ${\Gamma _{{P_i}}}  = \left\{ {{\Theta_i ^{CL}},{\Theta_i ^{CG}},{\Theta_i ^{DL}},{\Theta_i ^{DG}},{\Theta_i ^{CB}},\\{\Theta_i ^{DB}},{\Theta_i ^{CS}}} \right\}$, where $C$ and $D$ represent color and depth information respectively, $L$ indicates that saliency is determined in the local scope and $G$ indicates the global scope, $B$ and $S$ represent the background and color compactness priors respectively. More specifically, the color based feature vectors are recorded in the following formula,
\begin{equation}\label{Color_LG}
  \left\{ \begin{array}{l}
{\Theta_i ^{CL}} = \{ P_{i,1}^{CL},...,P_{i,j}^{CL},...,P_{i,N}^{CL}\} \\
{\Theta_i ^{CG}} = \{ P_{i,1}^{CG},...,P_{i,j}^{CG},...,P_{i,N}^{CG}\} \\
{\Theta_i ^{CB}} = \{ P_{i,1}^{CB},...,P_{i,j}^{CB},...,P_{i,N_b}^{CB}\} \\
{\Theta_i ^{CS}} = \{ P_{i,1}^{CS},...,P_{i,j}^{CS},...,P_{i,N}^{CS}\} \\
\end{array} \right.,
\end{equation}
and the depth based feature vectors are defined similarly. We compute the color-based features in $Lab$ color space.

The local color contrast $P_{i,j}^{CL}$ is calculated as:
\begin{equation}\label{color_contrast local}
  P_{i,j}^{CL} = t({{{j}}})\phi_L ({i},{j})\left\| {{\bm{c}_{{i}}} - {\bm{c}_{{j}}}} \right\|_2,
\end{equation}
where $t({j})$ is the total number of pixels in region $P_j$, and a larger superpixel contributes more to the saliency. ${\bm{c}_{i}}$ and ${\bm{c}_{j}}$ are the mean color values of the region $P_i$ and $P_j$. $\phi_L ({i},{j})$ is used to control the spatial influential distance. This weight is defined as $\exp ( - \frac{{{{\left\| {{{\bm{x}_i}} - {{\bm{x}_j}}} \right\|_2^2}}}}{{2\sigma _{Lr}^2}})$, and $\bm{x}_i$ and $\bm{x}_j$ are the centers of corresponding regions. In our experiment, the parameter $\sigma _{Lr}$ = 0.15 is set to make the neighbors have higher influence on the calculated contrast values, while the influence of other regions are negligible. Similar to the local color contrast vector, the global color contrast vector is defined as,
\begin{equation}\label{color_contrast_global}
 P_{i,j}^{CG} = t({{j}}){\phi _G}({i},{j})\left\| {{\bm{c}_{{i}}} - {\bm{c}_{{j}}}} \right\|_2.
\end{equation}
The difference between the global contrast and local contrast lies in the spatial weight $\phi_G ({i},{j})$, where in the global contrast the parameter $\sigma _{Gr}$ is set to 0.45 to cover the entire image.

Likewise, the depth contrast between region $P_j$ and region $P_i$ can be calculated as in Eq. \ref{Depth_local} and Eq. \ref{Depth_global}.
\begin{equation}\label{Depth_local}
  P_{i,j}^{DL} = t({{j}}){\phi _L}({i},{j})\left| {{{d}_i} - {{d}_j}} \right|,
\end{equation}
\begin{equation}\label{Depth_global}
  P_{i,j}^{DG} = t(j){\phi _G}({i},{j})\left| {{{d}_i} - {{d}_j}} \right|,
\end{equation}
where ${d_{i}}$ and ${d_{j}}$ are the mean depth values of the region $P_i$ and $P_j$ respectively.

Generally speaking, the colors of an object are compacted together whereas the colors belong to the background are widely distributed in the entire image. The element ${P_{i,j}^{CS}}$ in the color compactness based feature vector is calculated as following.
\begin{equation}\label{col_spatial}
  {P_{i,j}^{CS}} =  {\phi ({\bm{c}_{i}},{\bm{c}_{j}})\left\| {{\bm{x}_j} - \bm{u}_i^{cs}} \right\|_2},
\end{equation}
where the function $\phi(\bm{c}_{i},\bm{c}_{j})$ is used to calculate the similarity of two colors $\bm{c}_{i}$ and $\bm{c}_{j}$, and is defined as $\exp ( - \frac{{\left\| {{\bm{c}_{i}} - {\bm{c}_{j}}} \right\|_2^2}}{{2\delta _c^2}})$. $\bm{u}_i^{cs} = \sum\limits_{j = 1}^M {\phi ({\bm{c}_{i}},{\bm{c}_{j}})} {\bm{x}_j}$ defines the weighted mean position of color $\bm{c}_{i}$. The parameter $\delta _c$ is set to 20 in our implementation. We omit the depth compactness prior in our method since the depth map contains only dozens of depth levels and their spatial distributions can be very random. The experiment results also show that whether adding the depth compactness or not does not affect the final results too much.

Beside color compactness prior, we further introduce the background prior, which leverages the fact that salient object is less possible to be arranged to close to the image boundaries. We first extract $N_b$ regions along the image boundary as pseudo-background regions. Then the color or depth contrast to the pseudo-background regions will be calculated similar to Eq. \ref{color_contrast_global} and Eq. \ref{Depth_global}. In our experiment, the number of superpixels $N$ is set to 1024 and $N_b$ is set to 160.

\subsection{Hyper-feature extraction with CNN}\label{title_3_2}

Given the obtained saliency feature vectors, we then propose a CNN architecture to automatically incorporate them into unified and representative features. We formulate saliency detection as a binary logistic regression problem, which takes a patch as input and output the probabilities of two classes. Our CNN takes an input of size $32 \times 32 \times 6$, and generates a prediction as saliency output. For each superpixel $P_i$, all the seven saliency feature vectors are integrated into a multiple channel image as follows:

(1) Reshape the $N$ length vector (${\Theta ^{CL}}$, ${\Theta ^{CG}}$, ${\Theta ^{DL}}$, ${\Theta ^{DG}}$ and ${\Theta ^{CS}}$) to size $32 \times 32$ to form the first five channels, respectively;

(2) Perform zero padding to the $N_b$ length vector ${\Theta ^{CB}}$ and ${\Theta ^{DB}} $ to length $N/2$ and then concatenate and reshape them into size $32 \times 32$ to form the sixth channel.

As shown in Figure \ref{fig:Convnet}, our network consists of three convolutional layers followed by a fully connected layer and a logistic regression output layer with sigmoid nonlinear function. Following the first and second convolutional layers, we add an average pooling layer for translation invariance. We adopt the sigmoid function as the nonlinear mapping function for the three convolutional layers, while Rectified Linear Unites (ReLUs) is applied in the last two layers. Dropout procedure is applied after the first fully connected layers to avoid overfitting.

For simplification, we use $conv(N,K)$ and $fc(N)$ to indicate the convolutional layer and the fully connected layer with $N$ output and kernel size $K$. $pool(T,K)$ indicates the pooling layer with type $T$ and kernel size $K$. $sig$ and $relu$ represent the sigmoid function and ReLUs. Then the architecture of our CNN can be described as $conv1(6,5)-sig1-pool1(MEAN,2)-conv2(12,5)-sig2-pool2(MEAN,2)-conv3(24,3)-sig3-fc4(200)-relu4-dropout4-fc5(2)$. This proposed CNN was trained with back-propagation using stochastic gradient descent (SGD).

\subsection{Laplacian propagation}
\begin{figure*}[t]
\centering
\captionsetup[subfigure]{labelformat=empty}
\subfloat{ \label{fig:init_refine:a}\includegraphics[width=\widtheight,height=0.12\linewidth]{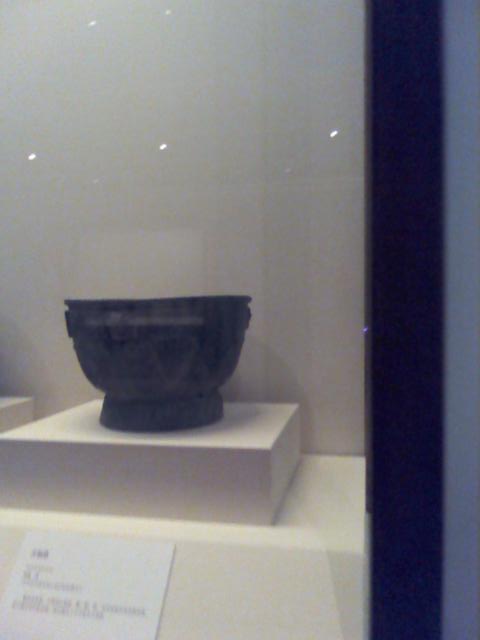}} \hspacefigure
\subfloat{ \label{fig:init_refine:b}\includegraphics[width=\widtheight,height=0.12\linewidth]{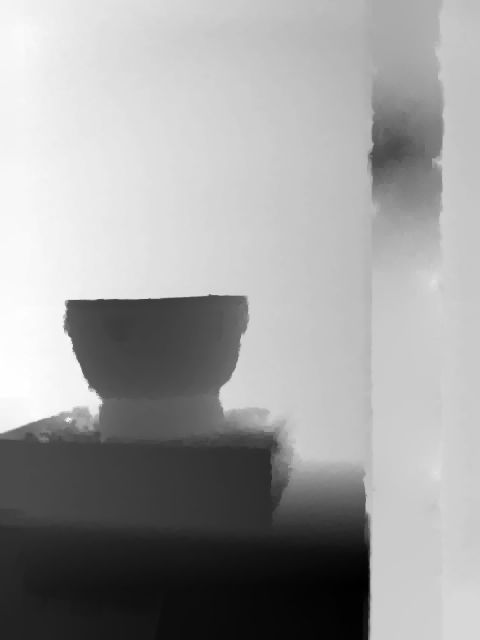}}\hspacefigure
\subfloat{ \label{fig:init_refine:c}\includegraphics[width=\widtheight,height=0.12\linewidth]{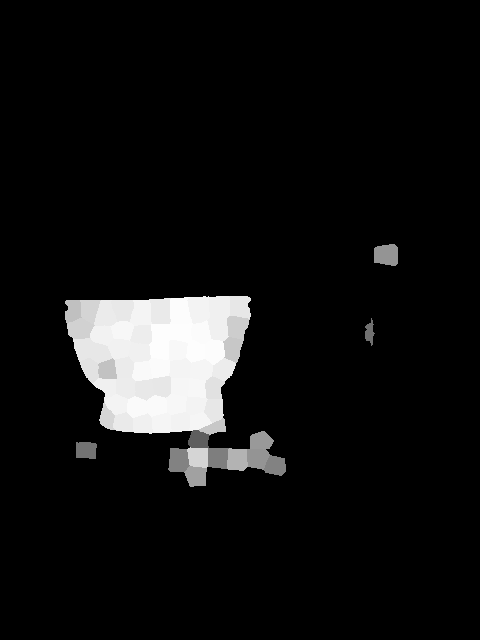}}\hspacefigure
\subfloat{ \label{fig:init_refine:d}\includegraphics[width=\widtheight,height=0.12\linewidth]{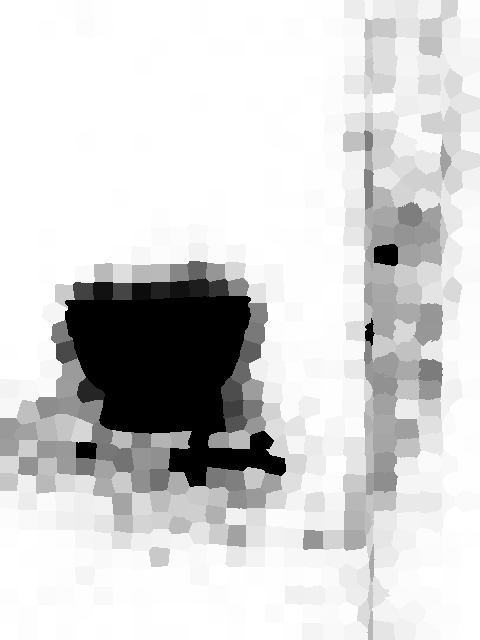}} \hspacefigure
\subfloat{ \label{fig:init_refine:e}\includegraphics[width=\widtheight,height=0.12\linewidth]{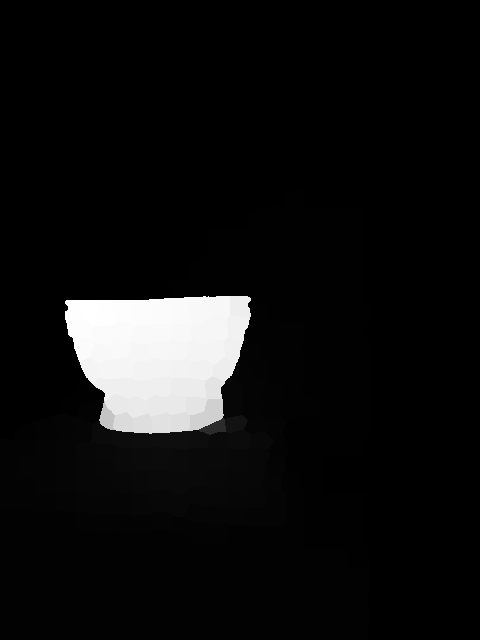}} \hspacefigure
\subfloat{ \label{fig:init_refine:f}\includegraphics[width=\widtheight,height=0.12\linewidth]{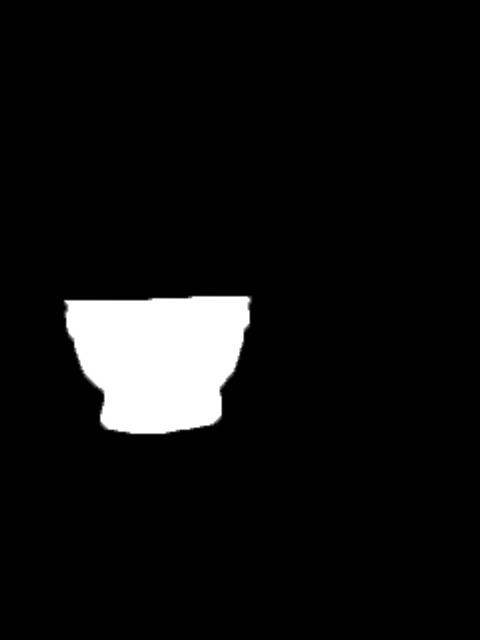}} \hspacefigure\\
\vspace{-1.5mm} 
 \subfloat[(a)]{ \includegraphics[width=\widtheight]{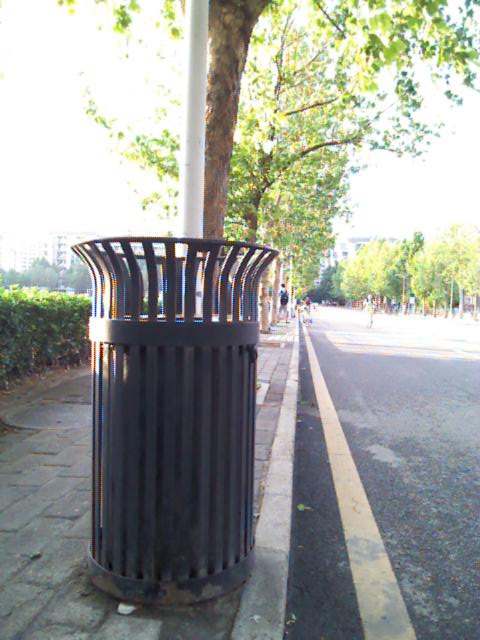}}\hspacefigure
 \subfloat[(b)]{ \includegraphics[width=\widtheight]{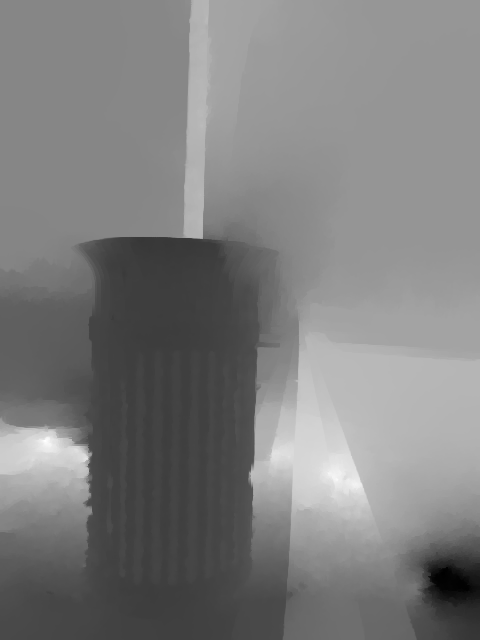}}\hspacefigure
 \subfloat[(c)]{ \includegraphics[width=\widtheight]{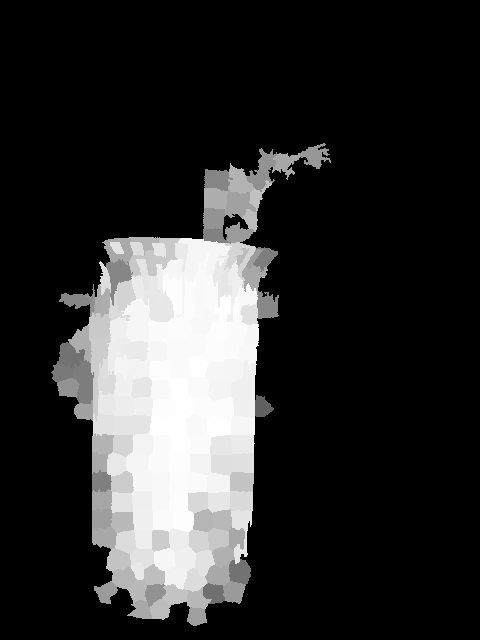}}\hspacefigure
 \subfloat[(d)]{ \includegraphics[width=\widtheight]{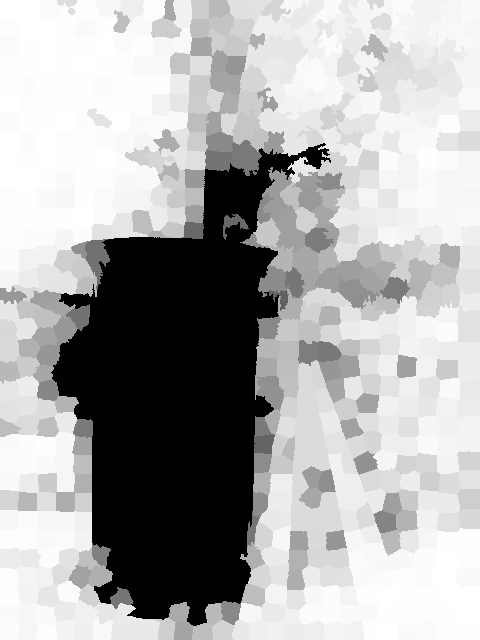}}\hspacefigure
 \subfloat[(e)]{ \includegraphics[width=\widtheight]{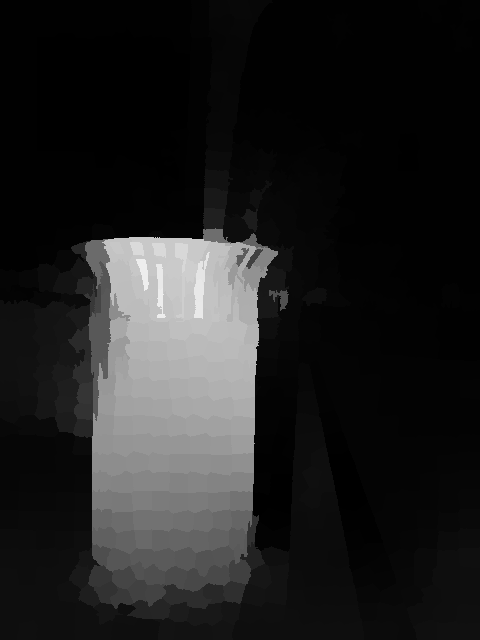}}\hspacefigure
 \subfloat[(f)]{ \includegraphics[width=\widtheight]{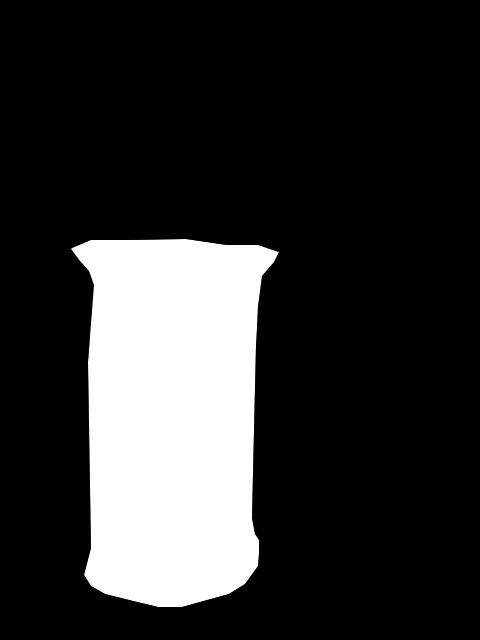}}\hspacefigure\\
  \caption{Examples of the proposed Laplacian propagation. (a) RGB image. (b) Depth image. (c) Saliency probability produced by the proposed CNN. (d) Background (non-salient) probability produced by the proposed CNN. (e) Refined saliency map using (c) and (d) as guidance. (f) The ground truth saliency map.}
  \vspace{-4mm}
\label{fig:init_refine}
\end{figure*}

As saliency values are estimated for each superpixel individually, the proposed CNN in Section \ref{title_3_2} may fail to retain the spatial consistency and lead to noisy output. Figure \ref{fig:init_refine:c} shows two examples of the saliency maps produced by our CNN for RGBD image. It indicates that our CNN omits some salient regions and wrongly detects some background regions as salient. Despite these misdetected regions, most of the regions with high probability to be salient are correct, robust, and reliable. The same situation also occurs for non-salient probability in the background (Figure \ref{fig:init_refine:d}). As a consequence, these high confident regions are used as guidance, and they are employed in a Laplacian propagation framework \cite{zhou2004learning} to obtain a more spatially consistent saliency map. The key of the Laplacian propagation lies in propagating the saliency from the regions with high probability to those ambiguous regions by considering two criteria:
(1) neighboring regions are more likely to have similar saliency values; and (2) regions within the same manifold are more likely to have similar saliency values.

Given a set of superpixels $\mathcal{P}  = \{ {P_1},{P_2},...,{P_N}\}$ of an input image $\mathcal{I}$ and a label set $\mathcal{L} = \{1,2\}$, we denote the salient and non-salient probability generated by the proposed CNN as $w^{sal}$ and $w^{non\_sal}$. The superpixels in $\mathcal{P}$ are labeled as 1 if $w^{sal} > \tau _1 $,  or as 2 if $w^{non\_sal} > \tau _2$. The goal of Laplacian propagation is to predict the labels of the remaining regions.

Let $F = {[\bm{f}_1^T,\bm{f}_2^T,...,\bm{f}_N^T]^T}$ denotes a $N\times2$ non-negative matrix which corresponds to the binary classification results of $\mathcal{P}$,
and each region $P_i$ is assigned with a label $y_i= \arg \max _{k= \{1, 2\}}f_{ik}$, where $\bm{f}_i = \{f_{i1},f_{i2}\}$. An indicator matrix is defined as $Y =  {[{y_{ik}}]_{N \times 2}}$ with $y_{ik}=1$ if region $P_i$ is labeled as $k$, otherwise $y_{ik} = 0$.
We further adopt the color and depth information to form the affinity matrix $A = {[{a_{ij}}]_{N \times N}}$:
\begin{equation}\label{w_ij}
  {a_{ij}} = \exp ( - \frac{{\left\| {{\bm{c}_{{i}}} - {\bm{c}_{{j}}}} \right\|_2^2}}{{2\delta _{_{1}}^2}})\exp ( - \frac{{\left| {{d_{{i}}} - {d_{{j}}}} \right|^2}}{{2\delta _{_{2}}^2}}),
\end{equation}
where the first term defines the color distance of superpixel region $P_i$ and $P_j$, and the second term defines the relative depth distance. Most of the elements of the affinity matrix $A$ are zero except for those neighbouring $P_i$ and $P_j$ pairs. In order to better leverage the local smoothness, we use a two-hierarchy neighboring connection model, i.e., each region is not only connected to its neighboring regions but also connected to the regions that share the same boundaries with its neighboring regions. We set $a_{ii} = 0$ to avoid self-reinforcement.
Then the Laplacian propagation can be formulated to solve the following optimization functions:
\begin{equation}\label{sal_optimization}
F^*=\arg\mathop {\min }\limits_F \frac{\mathcal{Q}(F)}{2},
\end{equation}
\begin{equation}\label{sal_optimization2}
\mathcal{Q}(F)=\sum\limits_{i,j = 1}^N {{a_{ij}}} {\left\| {\frac{{{\bm{f}_i}}}{{\sqrt {{m_{ii}}} }} - \frac{{{\bm{f}_j}}}{{\sqrt {{m_{jj}}} }}} \right\|_2^2} + \mu {\sum\limits_{i = 1}^N {\left\| {{\bm{f}_i} - {\bm{y}_i}} \right\|}_2 ^2},
\end{equation}
where parameter $\mu$ controls the balance between the smoothness constraint (the first term) and the fitting constraint (the second term). $m_{ii}$ is the element of the degree matrix $M$ derived from affinity matrix $A$, and $m_{ii} = \sum\limits_{j}{a_{ij}}$.
This designed smoothness constraint not only considers local smoothness but also confines the regions within the same manifold to have the same label by
constructing a smooth classifying function. This classifying function can change sufficiently slow along the coherent structure revealed by the original image \cite{zhou2004learning}.

This optimization function Eq. \ref{sal_optimization} can be solved using an iteration algorithm as shown in \cite{zhou2004learning}, or it can be reformulated into a linear system. For efficiency, we set the derivative of the $\mathcal{Q}({F})$ to zero and the optimal solution of Eq. \ref{sal_optimization} can be obtained by solving the following linear equation:
\begin{equation}\label{opt_solution}
  (I - \alpha S){{F}^*} = {Y},
\end{equation}
where $I$ is an identity matrix and $\alpha = 1/(1+\mu)$. We further adopt Conjugate Gradient and  preconditioner to solve this linear equation for fast convergence.

After propagating from the high probability salient and non-salient regions, the final saliency map is normalized to [0,1] and it is denoted as ${S} = \overline {{F}^*}$. Two examples of the proposed propagation are shown in Figure \ref{fig:init_refine}. Those wrongly estimated regions in Figure \ref{fig:init_refine:b} and Figure \ref{fig:init_refine:c} are corrected in the final saliency maps produced by the Laplacian propagation. In our implementation, parameters $\tau_1$ and $\tau_2$ are adaptively determined by Otsu method \cite{otsu1975threshold}.
\begin{figure*}[t]
\centering
\captionsetup[subfigure]{labelformat=empty}
\subfloat{ \label{fig:saliency2:a}\includegraphics[width=\widthtwelve,height=0.091\linewidth]{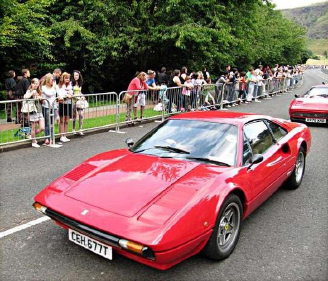}} \hspacefigure
\subfloat{ \label{fig:saliency2:b}\includegraphics[width=\widthtwelve,height=0.091\linewidth]{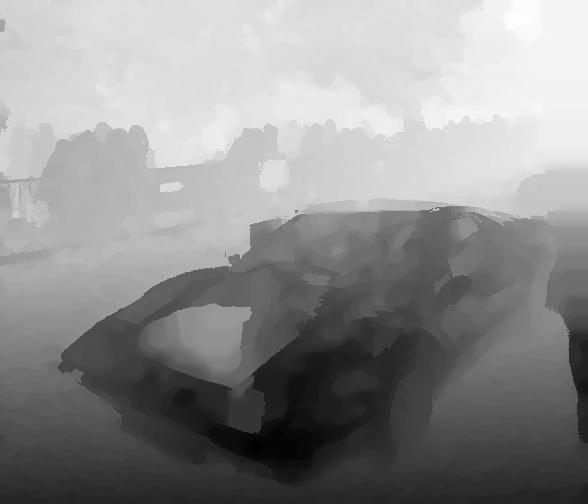}}\hspacefigure
\subfloat{ \label{fig:saliency2:c}\includegraphics[width=\widthtwelve,height=0.091\linewidth]{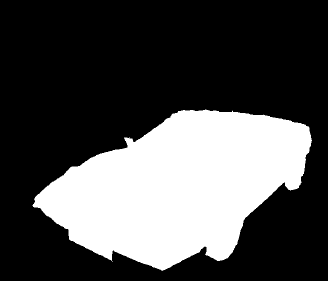}}\hspacefigure
\subfloat{ \label{fig:saliency2:e}\includegraphics[width=\widthtwelve,height=0.091\linewidth]{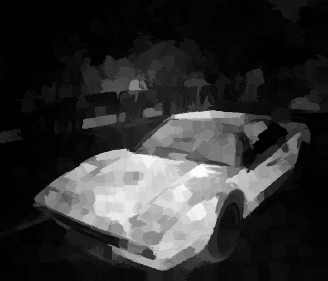}} \hspacefigure
\subfloat{ \label{fig:saliency2:f}\includegraphics[width=\widthtwelve,height=0.091\linewidth]{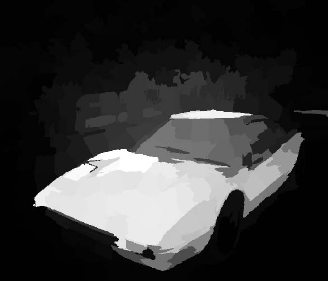}} \hspacefigure
\subfloat{ \label{fig:saliency2:g}\includegraphics[width=\widthtwelve,height=0.091\linewidth]{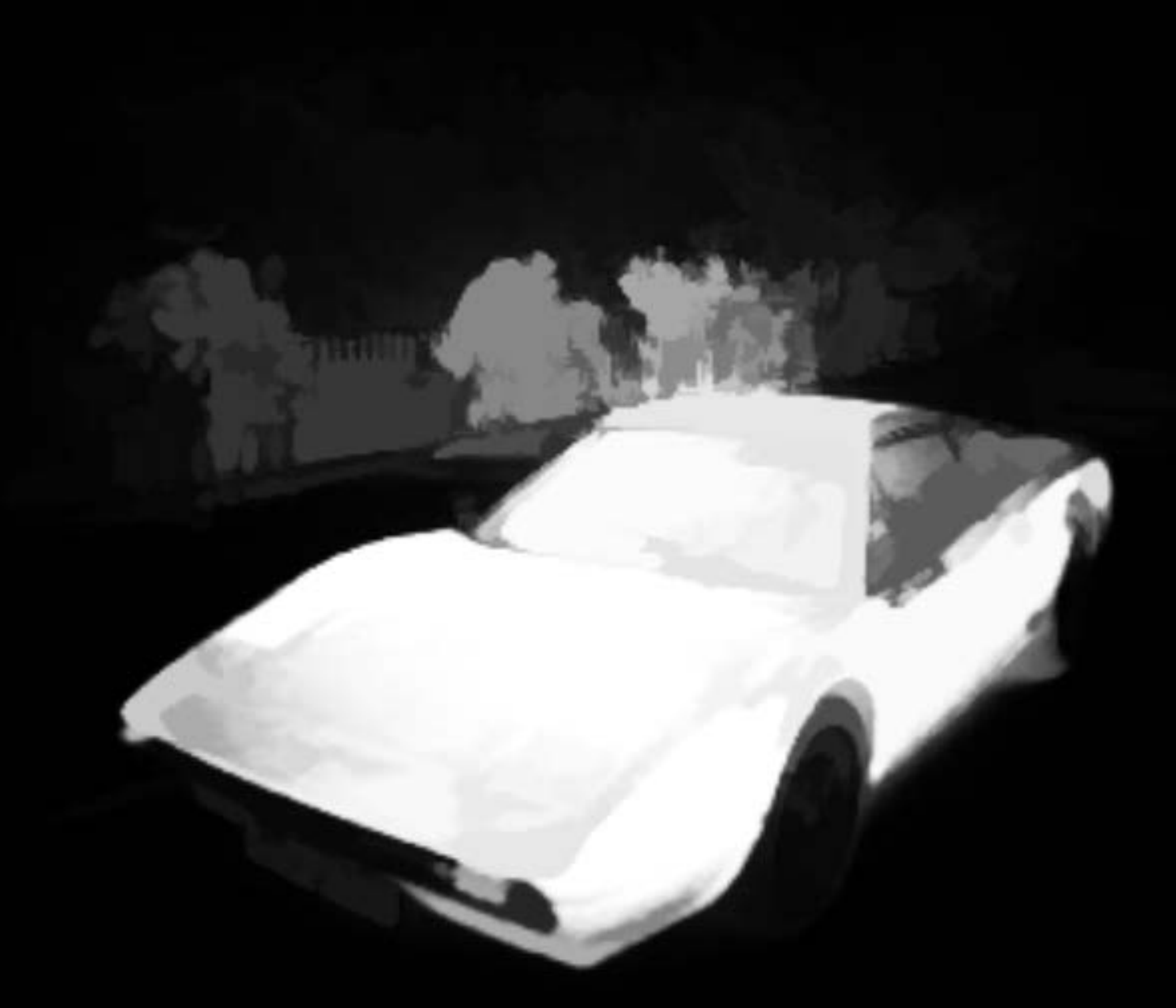}} \hspacefigure
\subfloat{ \label{fig:saliency2:g}\includegraphics[width=\widthtwelve,height=0.091\linewidth]{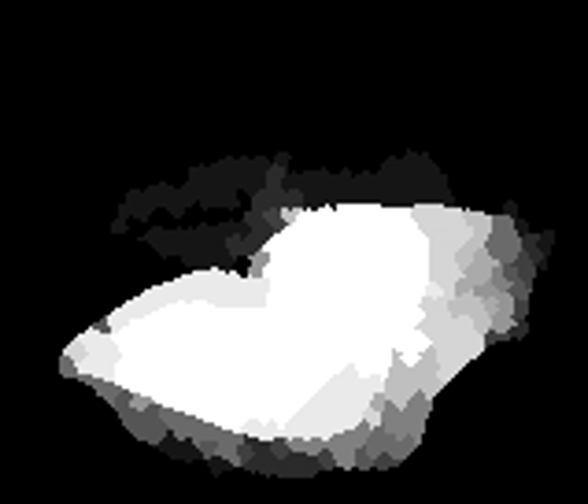}} \hspacefigure
\subfloat{ \label{fig:saliency2:i}\includegraphics[width=\widthtwelve,height=0.091\linewidth]{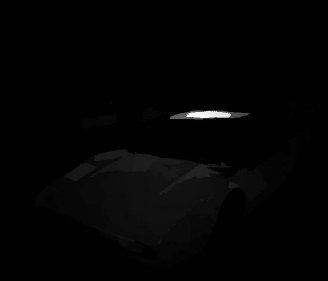}} \hspacefigure
\subfloat{ \label{fig:saliency2:j}\includegraphics[width=\widthtwelve,height=0.091\linewidth]{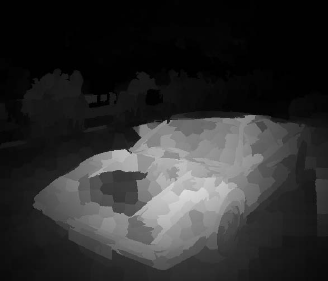}} \hspacefigure
\subfloat{ \label{fig:saliency2:k}\includegraphics[width=\widthtwelve,height=0.091\linewidth]{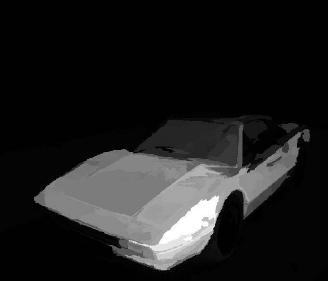}} \hspacefigure
\subfloat{ \label{fig:saliency2:l}\includegraphics[width=\widthtwelve,height=0.091\linewidth]{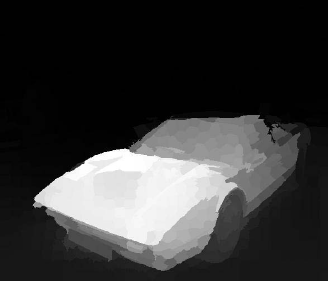}} \hspacefigure\\
\vspace{-1.5mm} 
\subfloat{ \includegraphics[width=\widthtwelve,height=0.075\linewidth]{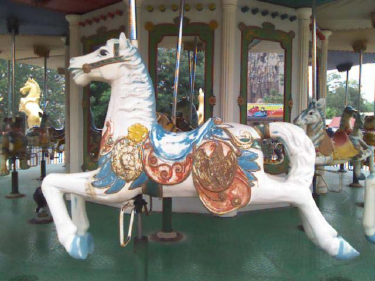}} \hspacefigure
\subfloat{ \includegraphics[width=\widthtwelve,height=0.075\linewidth]{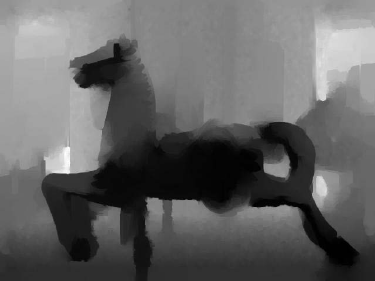}}\hspacefigure
\subfloat{ \includegraphics[width=\widthtwelve,height=0.075\linewidth]{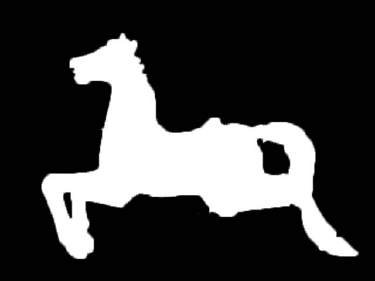}}\hspacefigure
\subfloat{ \includegraphics[width=\widthtwelve,height=0.075\linewidth]{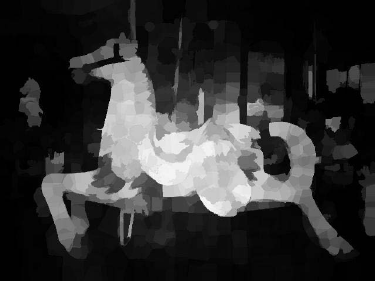}} \hspacefigure
\subfloat{ \includegraphics[width=\widthtwelve,height=0.075\linewidth]{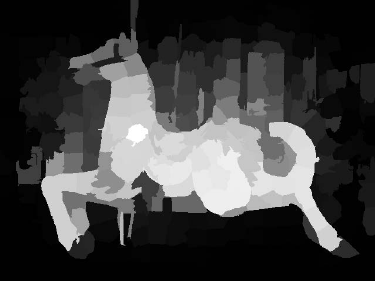}} \hspacefigure
\subfloat{ \includegraphics[width=\widthtwelve,height=0.075\linewidth]{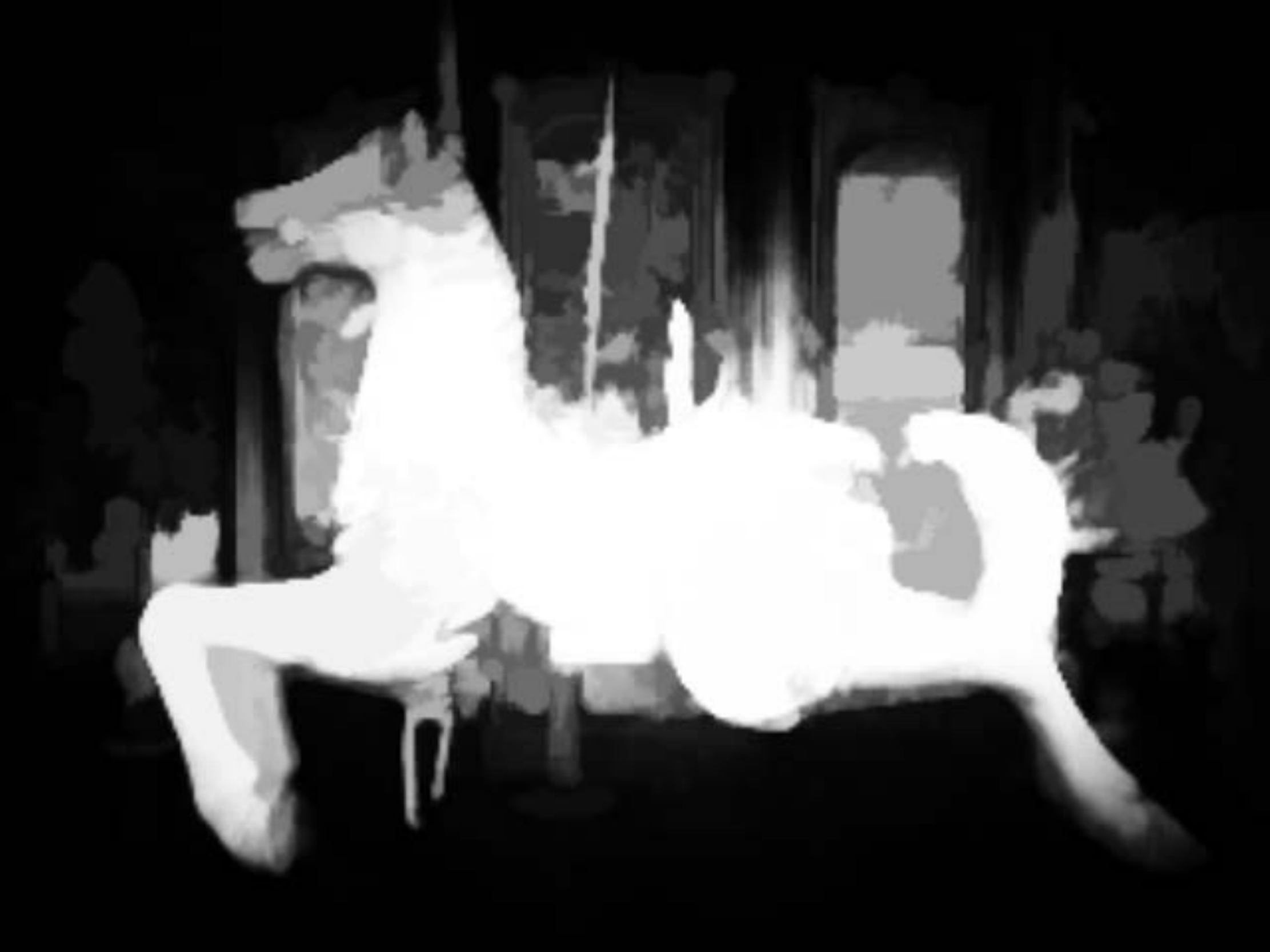}} \hspacefigure
\subfloat{ \includegraphics[width=\widthtwelve,height=0.075\linewidth]{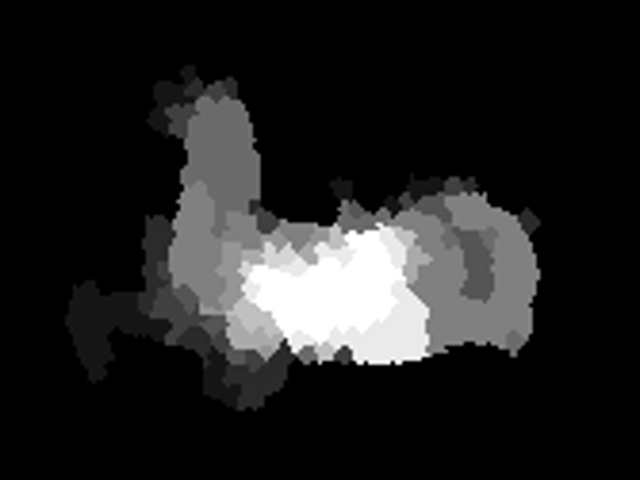}} \hspacefigure
\subfloat{ \includegraphics[width=\widthtwelve,height=0.075\linewidth]{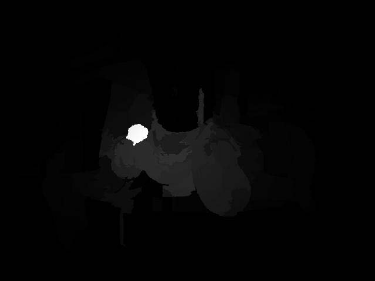}} \hspacefigure
\subfloat{ \includegraphics[width=\widthtwelve,height=0.075\linewidth]{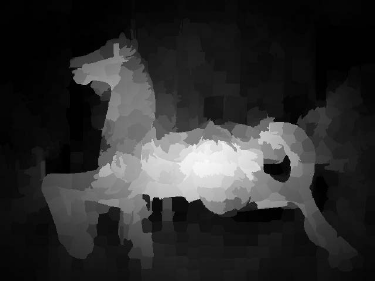}} \hspacefigure
\subfloat{ \includegraphics[width=\widthtwelve,height=0.075\linewidth]{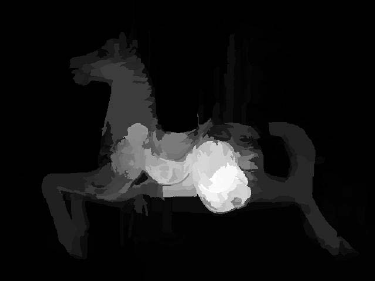}} \hspacefigure
\subfloat{ \includegraphics[width=\widthtwelve,height=0.075\linewidth]{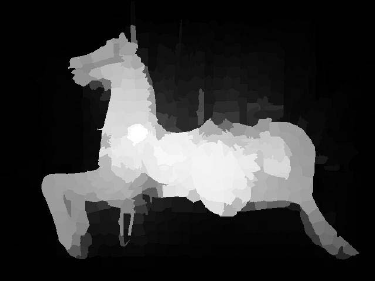}} \hspacefigure\\
 \vspace{-1.5mm}
\subfloat{ \includegraphics[width=\widthtwelve]{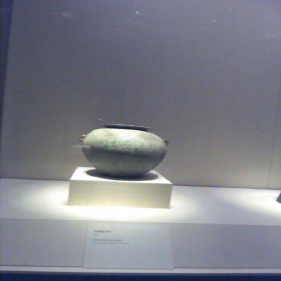}} \hspacefigure
\subfloat{ \includegraphics[width=\widthtwelve]{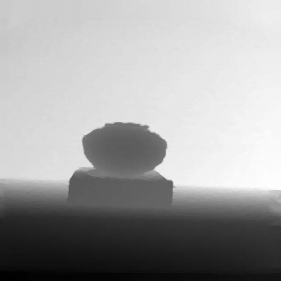}}\hspacefigure
\subfloat{ \includegraphics[width=\widthtwelve]{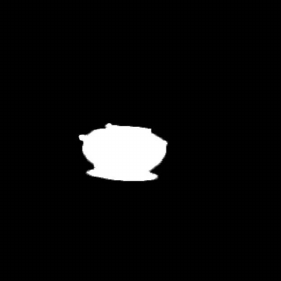}}\hspacefigure
\subfloat{ \includegraphics[width=\widthtwelve]{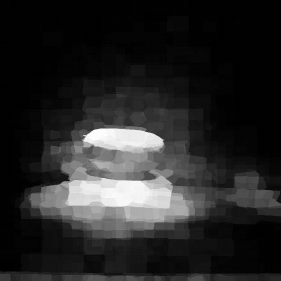}} \hspacefigure
\subfloat{ \includegraphics[width=\widthtwelve]{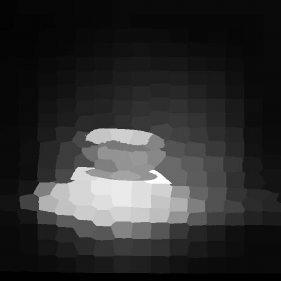}} \hspacefigure
\subfloat{ \includegraphics[width=\widthtwelve]{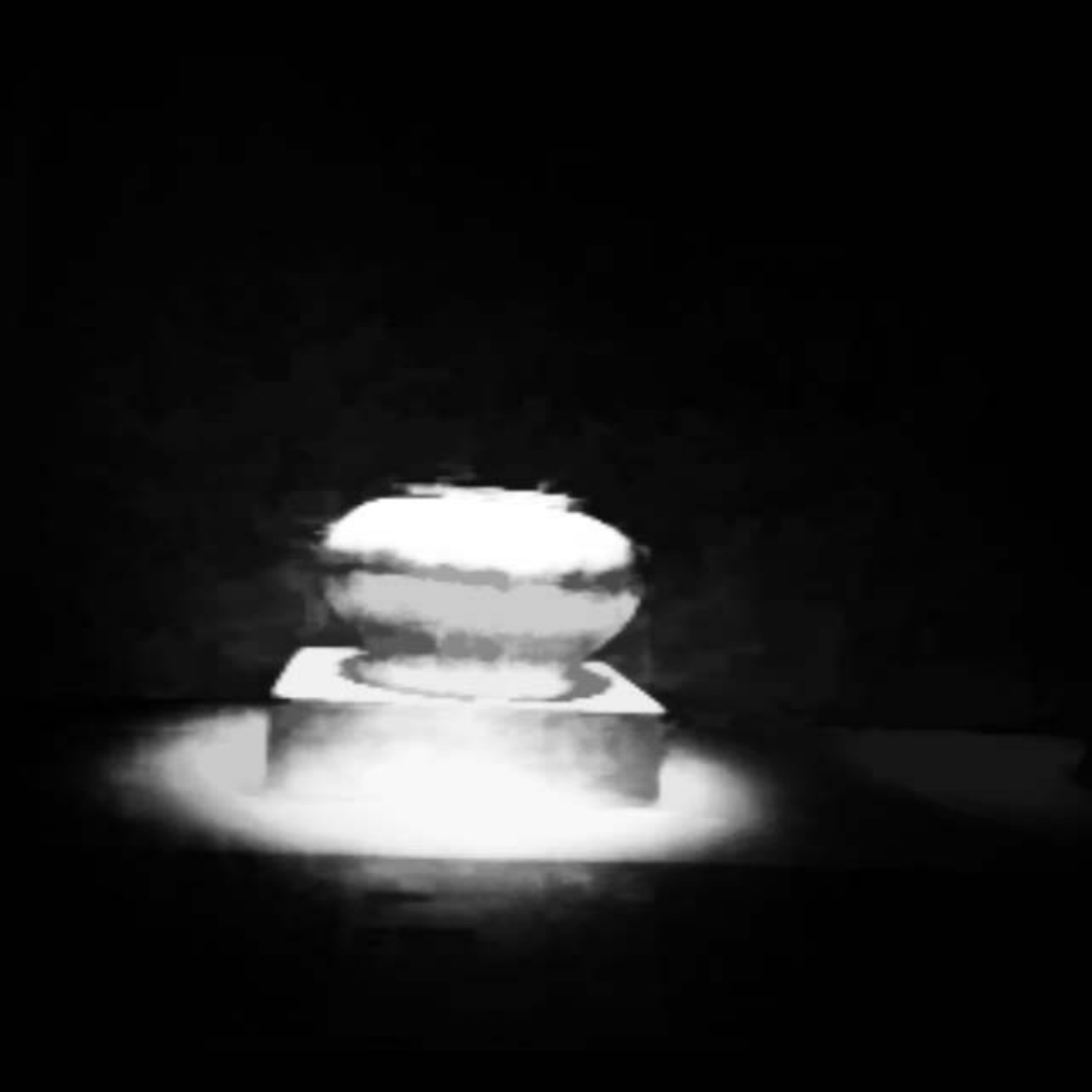}} \hspacefigure
\subfloat{ \includegraphics[width=\widthtwelve]{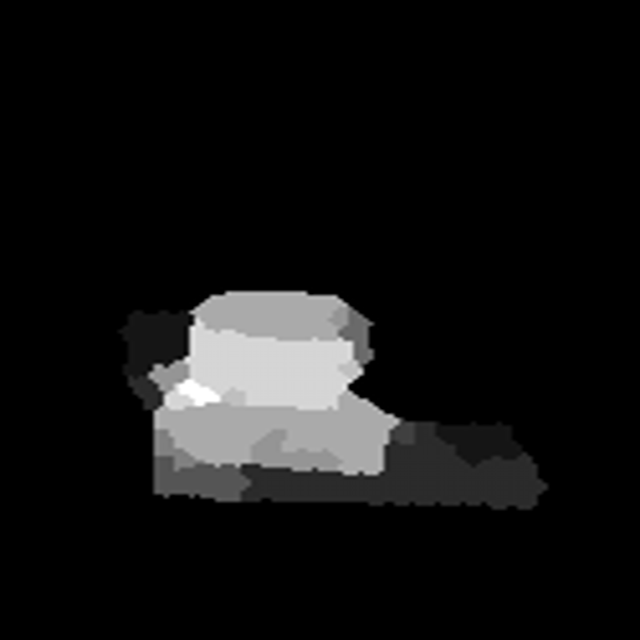}} \hspacefigure
\subfloat{ \includegraphics[width=\widthtwelve]{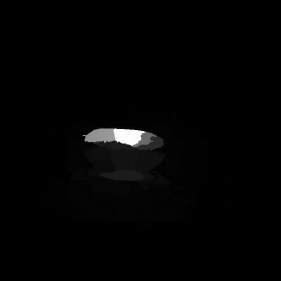}} \hspacefigure
\subfloat{ \includegraphics[width=\widthtwelve]{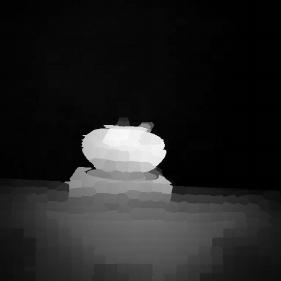}} \hspacefigure
\subfloat{ \includegraphics[width=\widthtwelve]{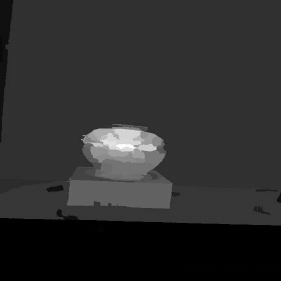}} \hspacefigure
\subfloat{ \includegraphics[width=\widthtwelve]{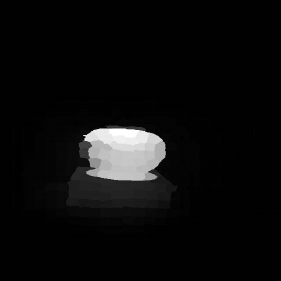}} \hspacefigure\\
 \vspace{-1.5mm}
\subfloat{ \includegraphics[width=\widthtwelve]{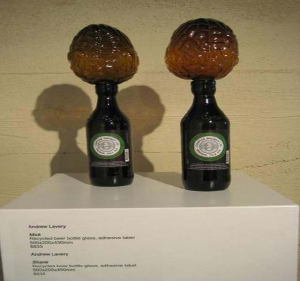}} \hspacefigure
\subfloat{ \includegraphics[width=\widthtwelve]{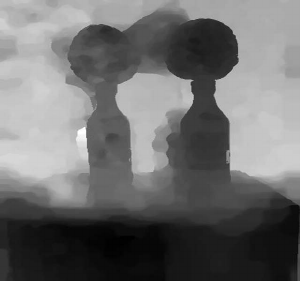}}\hspacefigure
\subfloat{ \includegraphics[width=\widthtwelve]{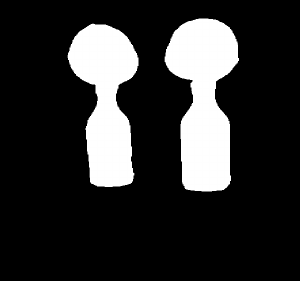}}\hspacefigure
\subfloat{ \includegraphics[width=\widthtwelve]{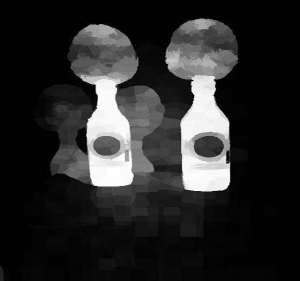}} \hspacefigure
\subfloat{ \includegraphics[width=\widthtwelve]{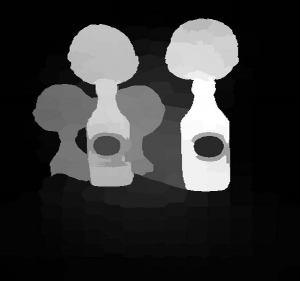}} \hspacefigure
\subfloat{ \includegraphics[width=\widthtwelve]{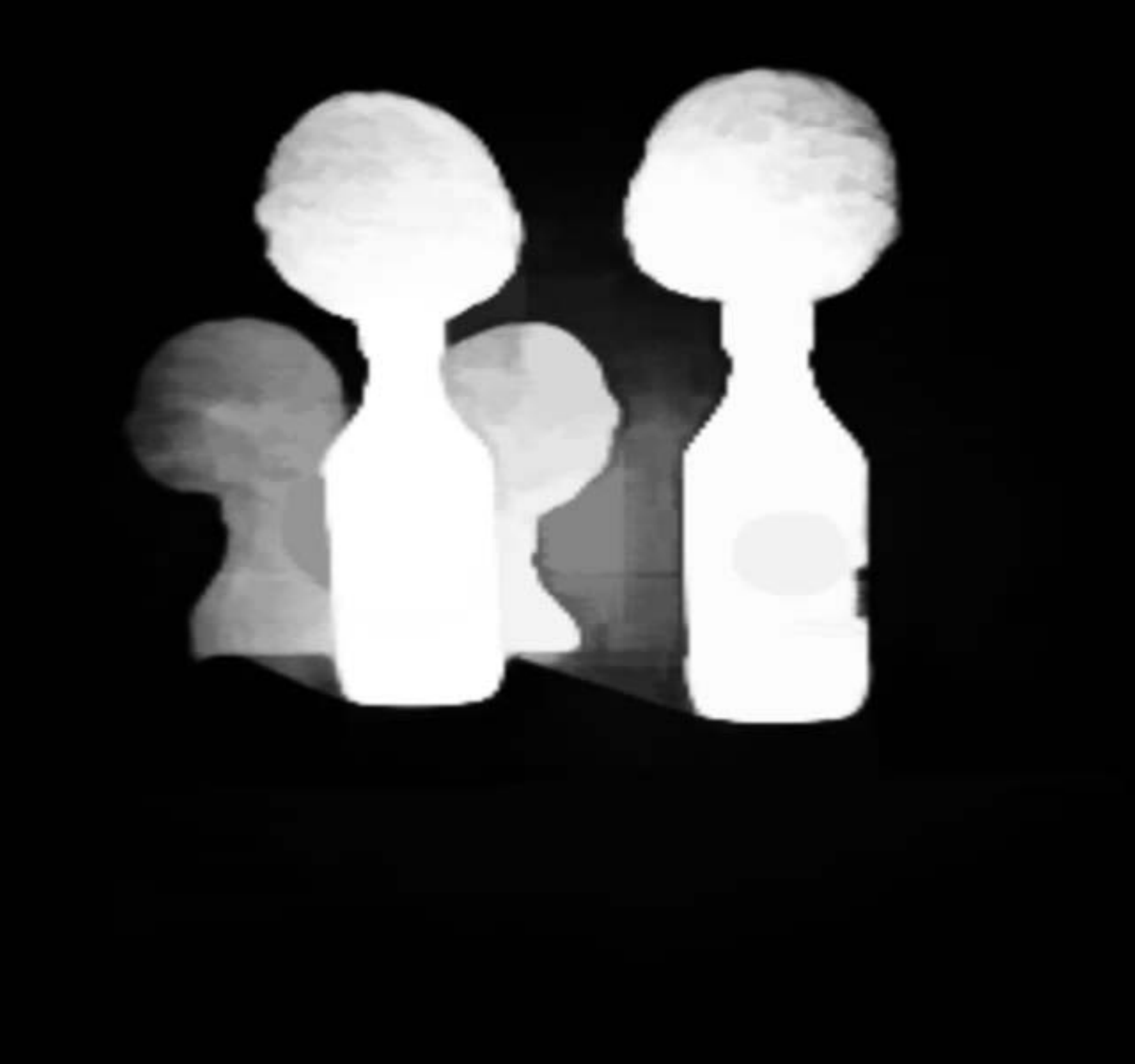}} \hspacefigure
\subfloat{ \includegraphics[width=\widthtwelve]{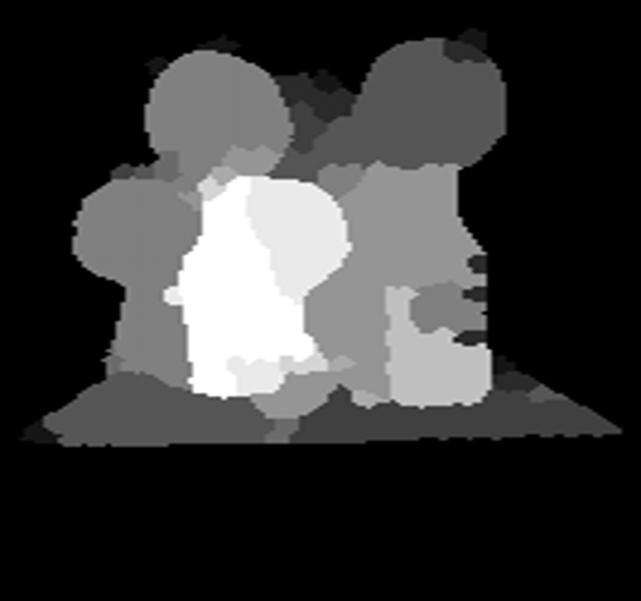}} \hspacefigure
\subfloat{ \includegraphics[width=\widthtwelve]{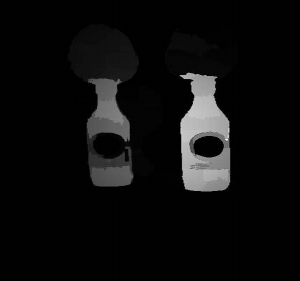}} \hspacefigure
\subfloat{ \includegraphics[width=\widthtwelve]{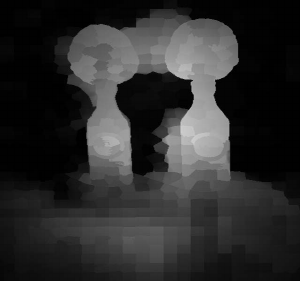}} \hspacefigure
\subfloat{ \includegraphics[width=\widthtwelve]{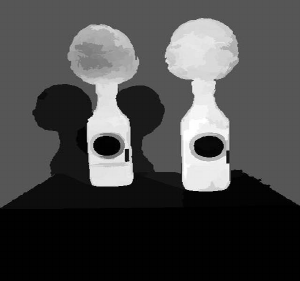}} \hspacefigure
\subfloat{ \includegraphics[width=\widthtwelve]{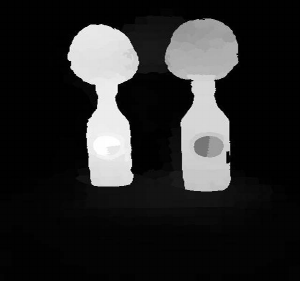}} \hspacefigure\\
\vspace{-1.5mm}
\subfloat{ \includegraphics[width=\widthtwelve]{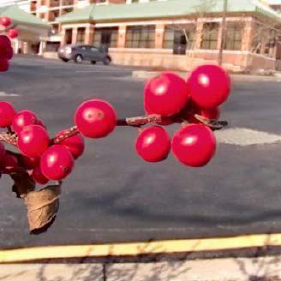}} \hspacefigure
\subfloat{ \includegraphics[width=\widthtwelve]{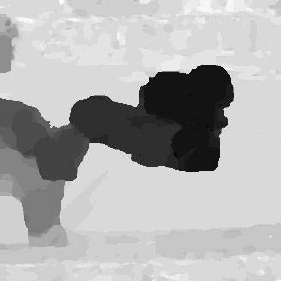}}\hspacefigure
\subfloat{ \includegraphics[width=\widthtwelve]{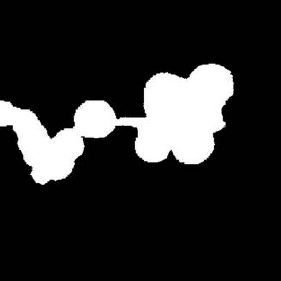}}\hspacefigure
\subfloat{ \includegraphics[width=\widthtwelve]{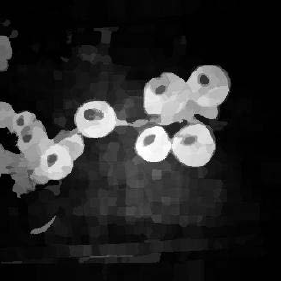}} \hspacefigure
\subfloat{ \includegraphics[width=\widthtwelve]{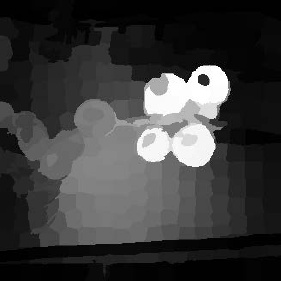}} \hspacefigure
\subfloat{ \includegraphics[width=\widthtwelve]{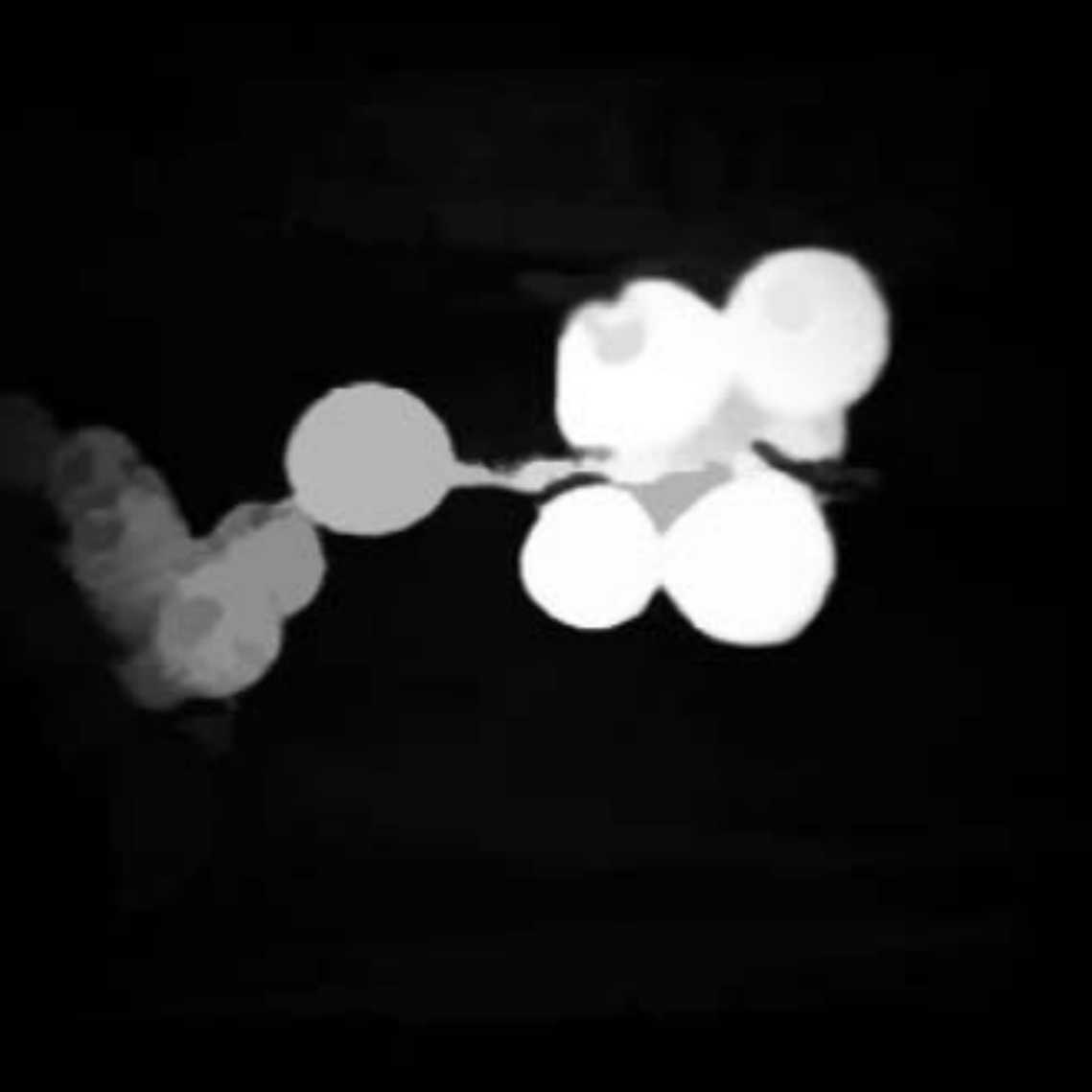}} \hspacefigure
\subfloat{ \includegraphics[width=\widthtwelve]{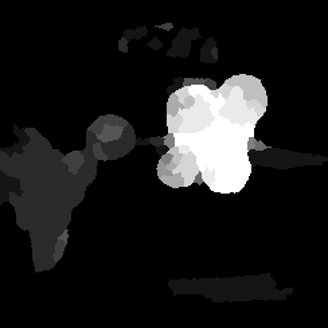}} \hspacefigure
\subfloat{ \includegraphics[width=\widthtwelve]{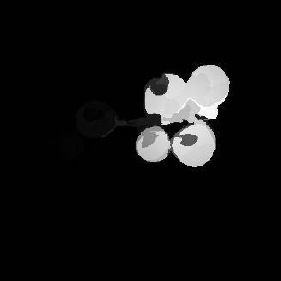}} \hspacefigure
\subfloat{ \includegraphics[width=\widthtwelve]{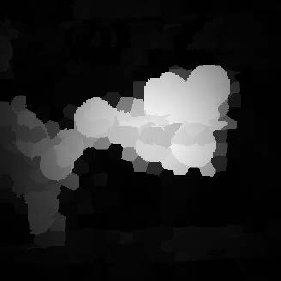}} \hspacefigure
\subfloat{ \includegraphics[width=\widthtwelve]{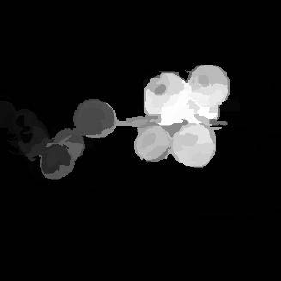}} \hspacefigure
\subfloat{ \includegraphics[width=\widthtwelve]{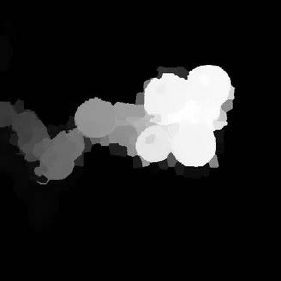}} \hspacefigure\\
\vspace{-1.5mm}
\subfloat{  \includegraphics[width=\widthtwelve]{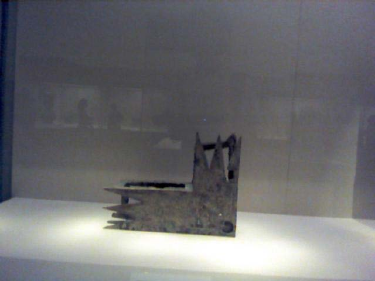}} \hspacefigure
\subfloat{ \includegraphics[width=\widthtwelve]{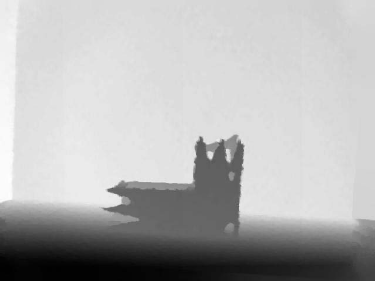}}\hspacefigure
\subfloat{ \includegraphics[width=\widthtwelve]{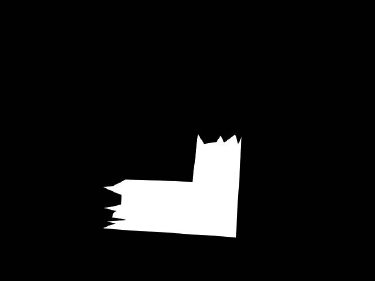}}\hspacefigure
\subfloat{ \includegraphics[width=\widthtwelve]{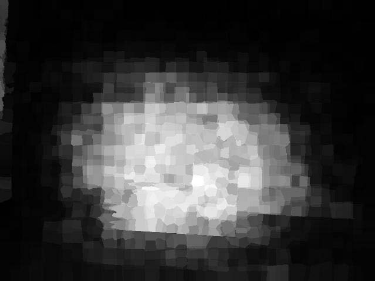}} \hspacefigure
\subfloat{ \includegraphics[width=\widthtwelve]{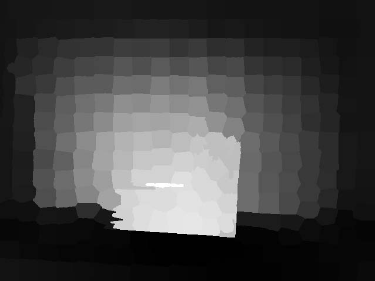}} \hspacefigure
\subfloat{ \includegraphics[width=\widthtwelve]{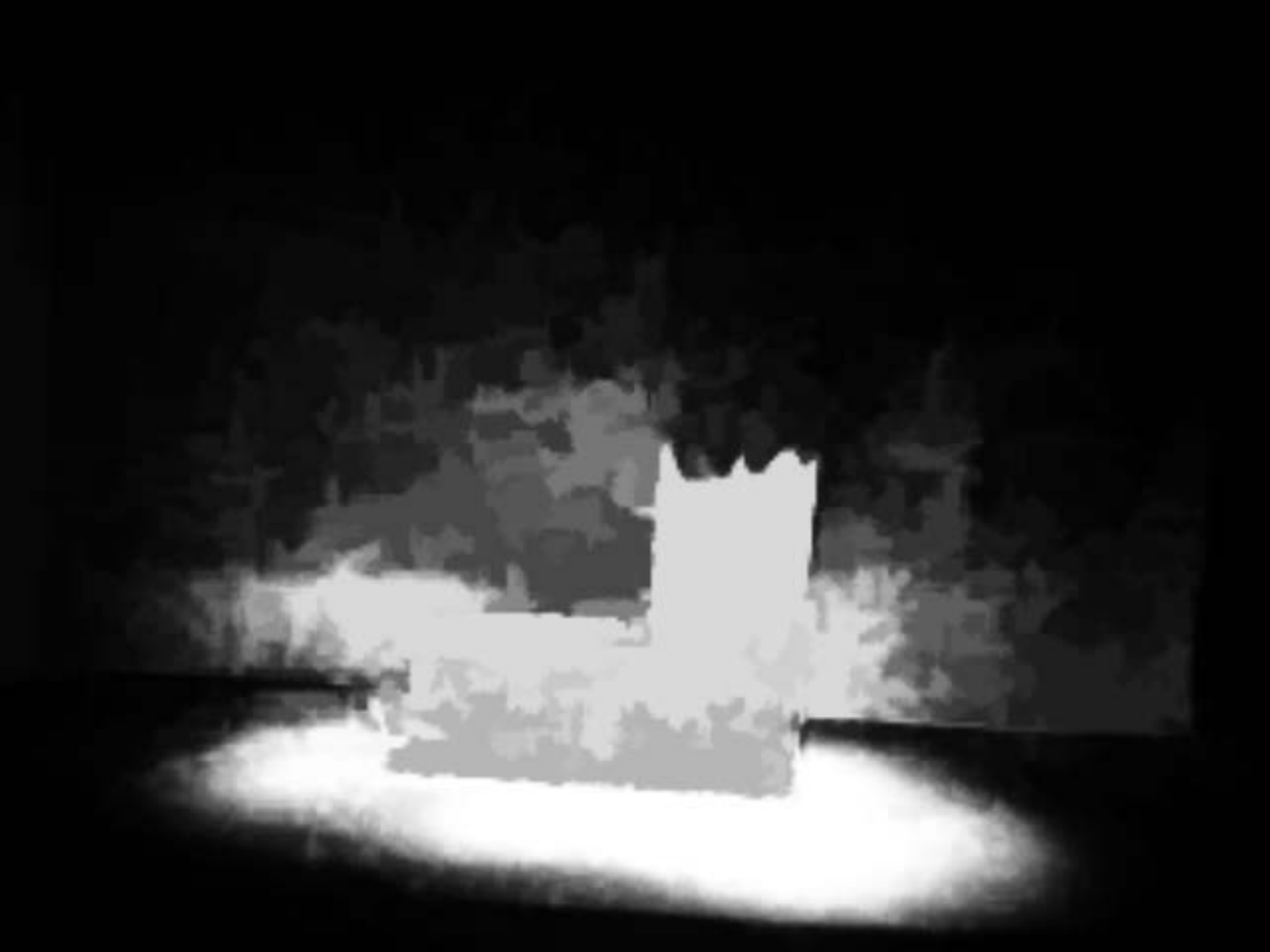}} \hspacefigure
\subfloat{ \includegraphics[width=\widthtwelve]{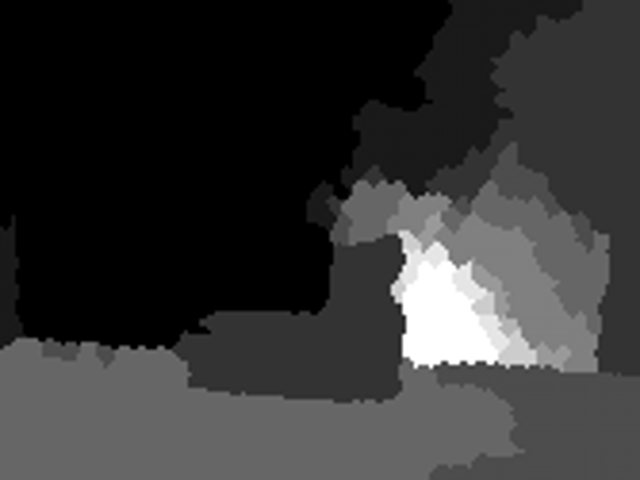}} \hspacefigure
\subfloat{ \includegraphics[width=\widthtwelve]{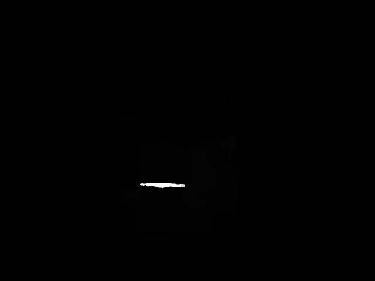}} \hspacefigure
\subfloat{ \includegraphics[width=\widthtwelve]{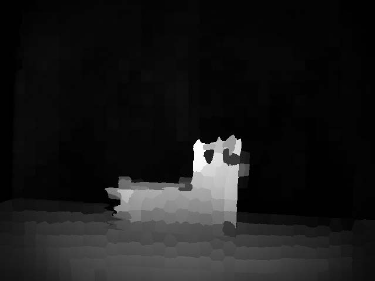}} \hspacefigure
\subfloat{ \includegraphics[width=\widthtwelve]{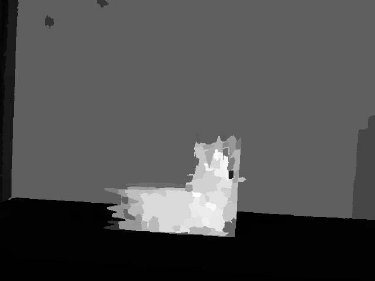}} \hspacefigure
\subfloat{ \includegraphics[width=\widthtwelve]{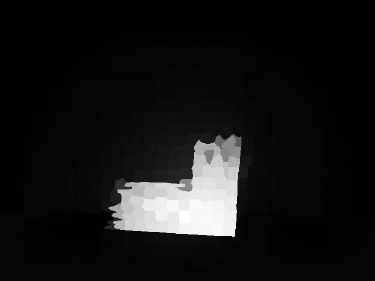}} \hspacefigure\\
\vspace{-1.5mm}
\subfloat{ \includegraphics[width=\widthtwelve,height=0.07\linewidth]{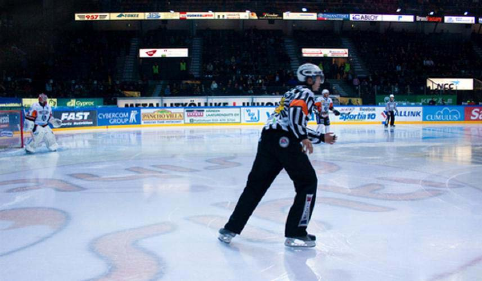}} \hspacefigure
\subfloat{ \includegraphics[width=\widthtwelve,height=0.07\linewidth]{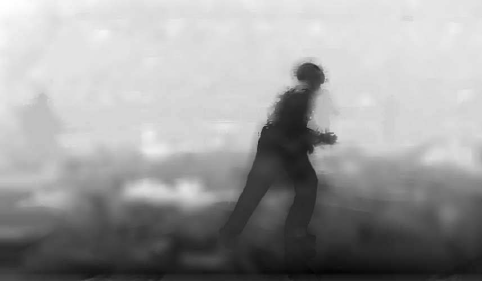}}\hspacefigure
\subfloat{ \includegraphics[width=\widthtwelve,height=0.07\linewidth]{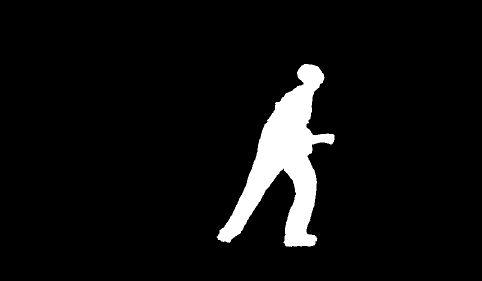}}\hspacefigure
\subfloat{ \includegraphics[width=\widthtwelve,height=0.07\linewidth]{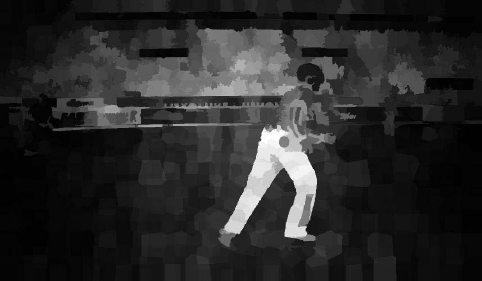}} \hspacefigure
\subfloat{ \includegraphics[width=\widthtwelve,height=0.07\linewidth]{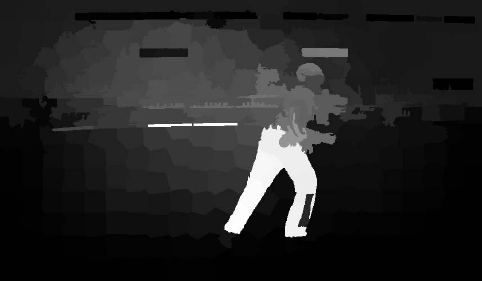}} \hspacefigure
\subfloat{ \includegraphics[width=\widthtwelve,height=0.07\linewidth]{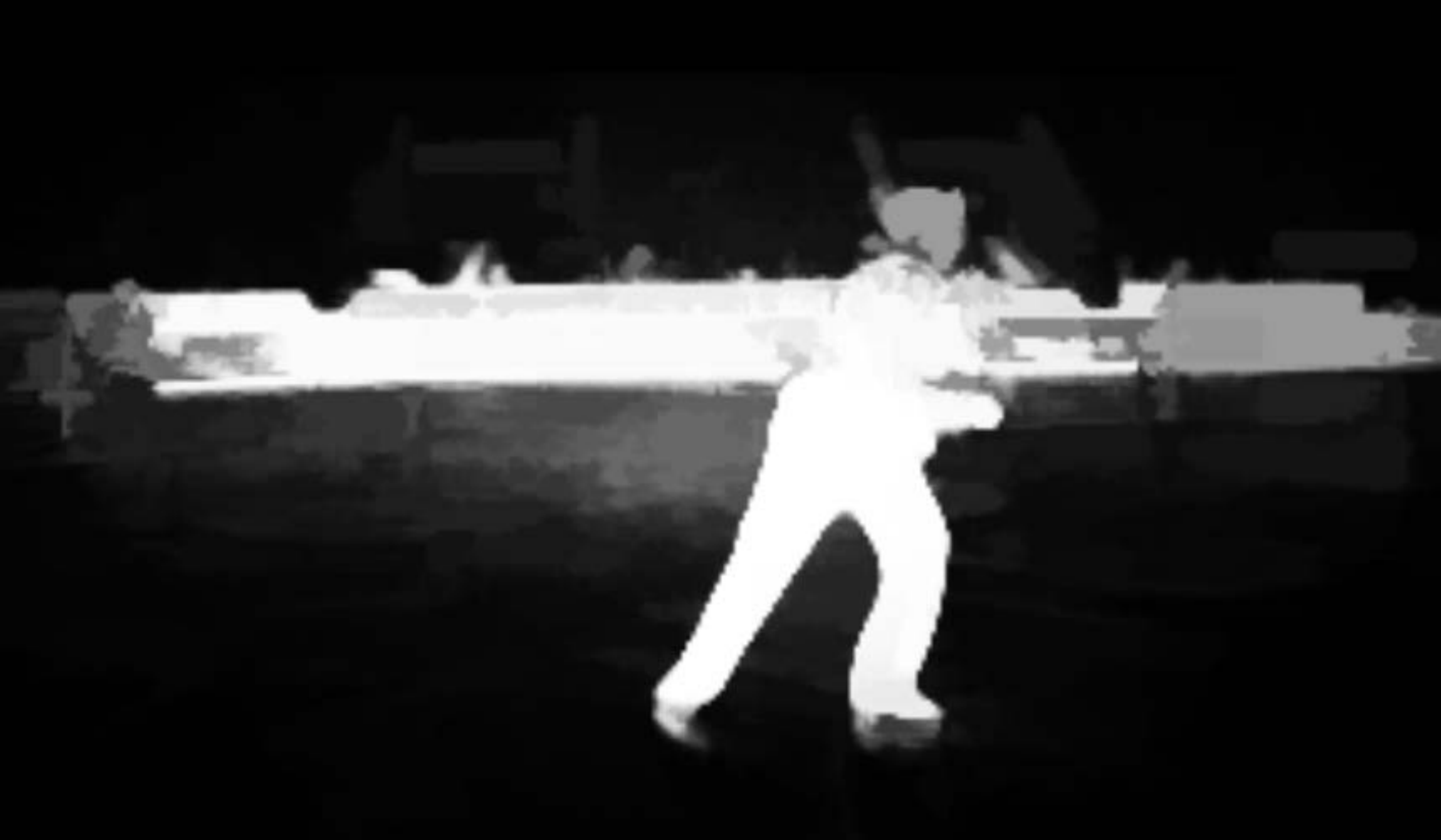}} \hspacefigure
\subfloat{ \includegraphics[width=\widthtwelve,height=0.07\linewidth]{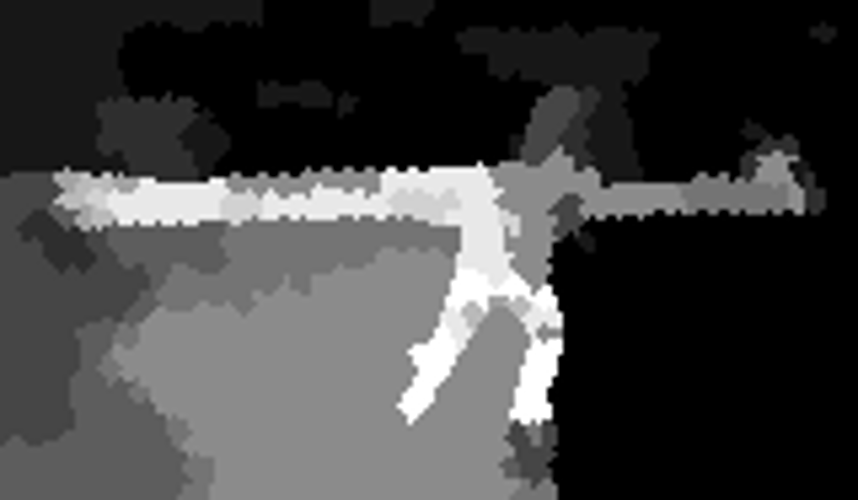}} \hspacefigure
\subfloat{ \includegraphics[width=\widthtwelve,height=0.07\linewidth]{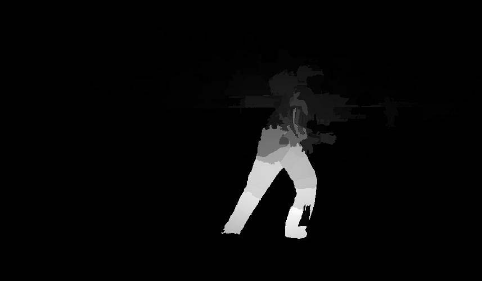}} \hspacefigure
\subfloat{ \includegraphics[width=\widthtwelve,height=0.07\linewidth]{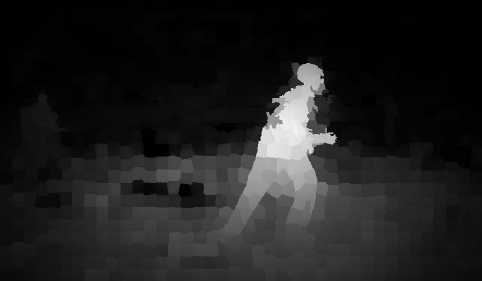}} \hspacefigure
\subfloat{ \includegraphics[width=\widthtwelve,height=0.07\linewidth]{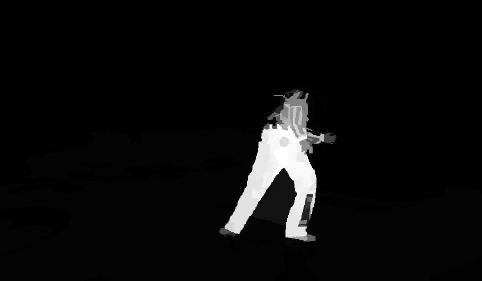}} \hspacefigure
\subfloat{ \includegraphics[width=\widthtwelve,height=0.07\linewidth]{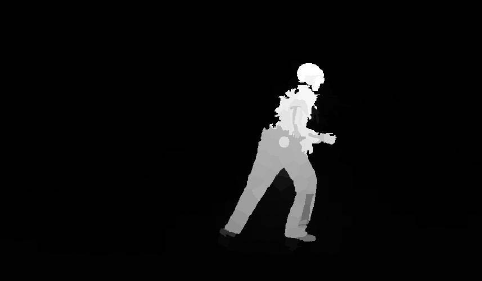}} \hspacefigure\\
\vspace{-1.5mm}
\subfloat[{RGB}]{  \includegraphics[width=\widthtwelve]{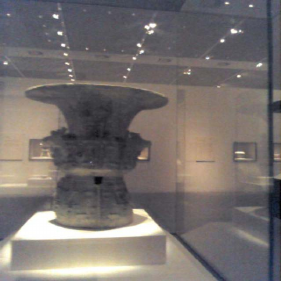}} \hspacefigure
\subfloat[{Depth}]{ \includegraphics[width=\widthtwelve]{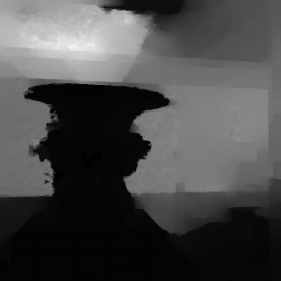}}\hspacefigure
\subfloat[{GT}]{ \includegraphics[width=\widthtwelve]{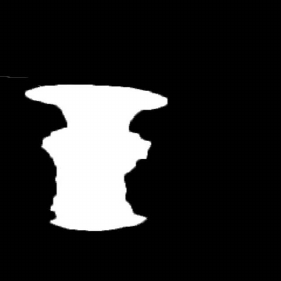}}\hspacefigure
\subfloat[{S-CNN}]{ \includegraphics[width=\widthtwelve]{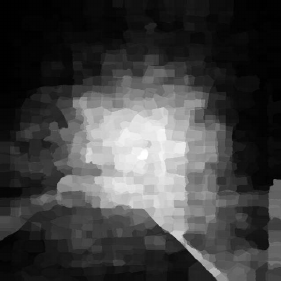}} \hspacefigure
\subfloat[{BSCA} ]{ \includegraphics[width=\widthtwelve]{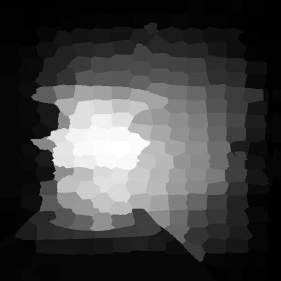}} \hspacefigure
\subfloat[{MB+} ]{ \includegraphics[width=\widthtwelve]{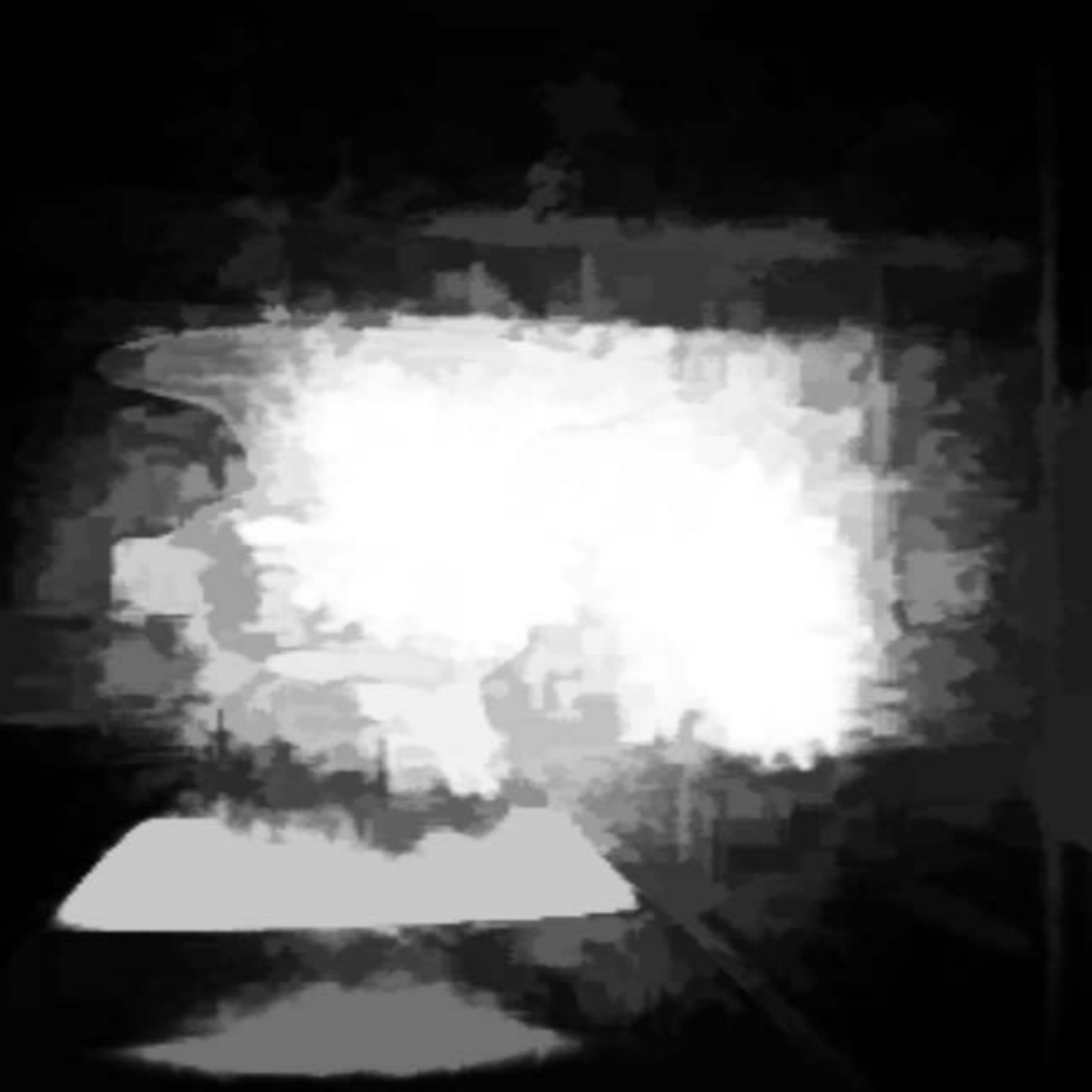}} \hspacefigure
\subfloat[{LEGS}]{ \includegraphics[width=\widthtwelve]{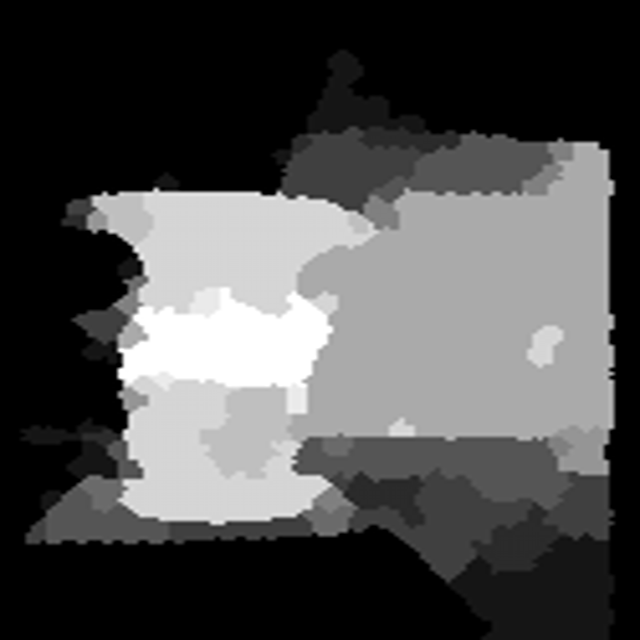}} \hspacefigure
\subfloat[{LMH}]{ \includegraphics[width=\widthtwelve]{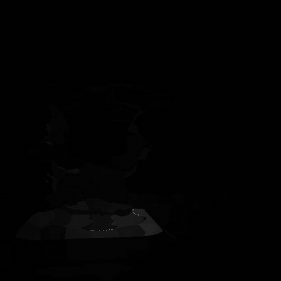}} \hspacefigure
\subfloat[{ACSD}]{ \includegraphics[width=\widthtwelve]{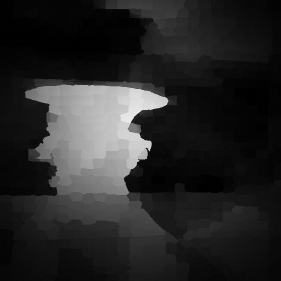}} \hspacefigure
\subfloat[{GP}]{ \includegraphics[width=\widthtwelve]{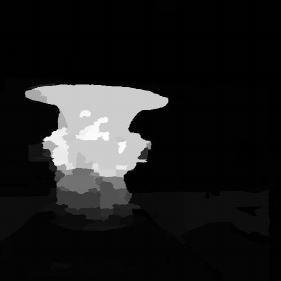}} \hspacefigure
\subfloat[{Ours}]{ \includegraphics[width=\widthtwelve]{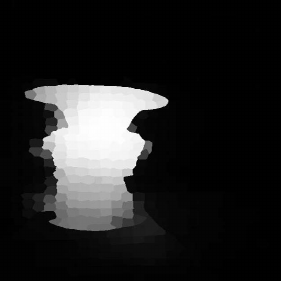}} \hspacefigure\\
  \caption{Visual comparisons of the proposed deep fusion framework with four RGB saliency methods and three RGBD saliency methods. The saliency maps of S-CNN \cite{he2015supercnn}, BSCA \cite{qin2015saliency}, MB+ \cite{zhang2015MBD}, and LEGS \cite{wang2015deep} are obtained from RGB image while the saliency maps of LMH \cite{peng2014rgbd}, ACSD \cite{ju2014depth}, GP \cite{ren2015exploiting} are from RGBD image.}
\label{fig:saliency2}
\end{figure*}

\section{Experimental evaluations}
In this section, we evaluate the proposed method on three datasets, NLPR RGBD salient dataset \cite{peng2014rgbd}, NJUDS2000 stereo datast \cite{ju2014depth}, and LFSD dataset \cite{Li_2014_CVPR}.

\textbf{NLPR dataset \cite{peng2014rgbd}.}
The NLPR RGBD salient dataset \cite{peng2014rgbd} contains 1000 images captured by Microsoft Kinect in different indoor and outdoor scenarios. We split this dataset into two part randomly: 750 for training and 250 for testing.

\textbf{NJUDS2000 dataset \cite{ju2014depth}.}
The NJUDS2000 dataset contains 2000 stereo images, as well as the corresponding depth maps and manually labeled groundtruth. The depth maps are generated using an optical flow method.
We also split this dataset into two part randomly: 1000 for training and 1000 for testing.

\textbf{LFSD dataset \cite{Li_2014_CVPR}.}
The LFSD dataset \cite{Li_2014_CVPR} contains 100 images with depth information and manually labeled groundtruth. The depth information are captured with Lytro light field camera. All the images in this dataset are for testing.

\textbf{Evaluation metrics.} We compute the precision-recall (PR) curve, mean of average precision and recall, and F-measure score to evaluate the performance of different saliency detection methods. The PR curve indicates the mean precision and recall of the saliency map at different thresholds. The F-measure is defined as ${F_\beta } = \frac{{(1 + {\beta ^2}) \times precision \times recall}}{{{\beta ^2} \times precision + recall}}$, where ${{\beta ^2}}$ is set to 0.3.

\subsection{Implementation details}
We use the randomly sampled 750 training images of NLPR dataset \cite{peng2014rgbd} and the randomly sampled 1000 training images of NJUDS2000 dataset \cite{ju2014depth} to train our deep learning framework. These randomly selected training dateset covers more than 1000 kinds of common objects under different circumstances. The remaining NLPR, NJUDS2000, and LFSD datesets are used to verify the generalization of the proposed method.

The proposed method is implemented using Matlab. We set the momentum in our network to 0.9 and the weight decay to be 0.0005. The learning rate of our network is gradually decreased from 1 to 0.001. Due to the ``data-hungry'' nature of CNN, the existing training data is insufficient for training, in addition to the dropout procedure, we also employed data augmentation to enrich our training dataset. Similar to \cite{krizhevsky2012imagenet}, we adopted two different image augmentation operations, the first one consists of image translations and horizontal flipping and the other is to alter the intensities of the RGB channels. These data augmentations greatly enlarge our training dataset and make it possible for us to train the proposed CNN without overfitting. It took around $5\sim7$ days for our training to converge.

\begin{table*}
\centering
\caption{The F-measure scores of different approaches on three datasets.}
\label{table:belta}
\begin{tabular}{|c||c|c|c|cIc|c|c|c|}
\hline
{Dataset} & S-CNN  & BSCA  &  MB+ & LEGS & LMH & ACSD & GP & Ours \\
\hline\hline
NLPR test set & 0.5141 & 0.5634 & 0.6049 & 0.6335 & 0.6519 & 0.5448 & 0.7184 & \textbf{0.7823}\\
NJUD test set & 0.6096 & 0.6133 & 0.6156 & 0.6791 & 0.6381 & 0.6952 &0.7246 & \textbf{0.7874} \\
LFSD dataset & 0.6982 & 	0.7311&	0.7029 & 0.7384&	0.7041&	0.7567&	0.7877&	\textbf{0.8439}\\
\hline
\end{tabular}
\end{table*}
\begin{figure*}
\vspace{-3mm}
\centering
\includegraphics[width=0.31\linewidth]{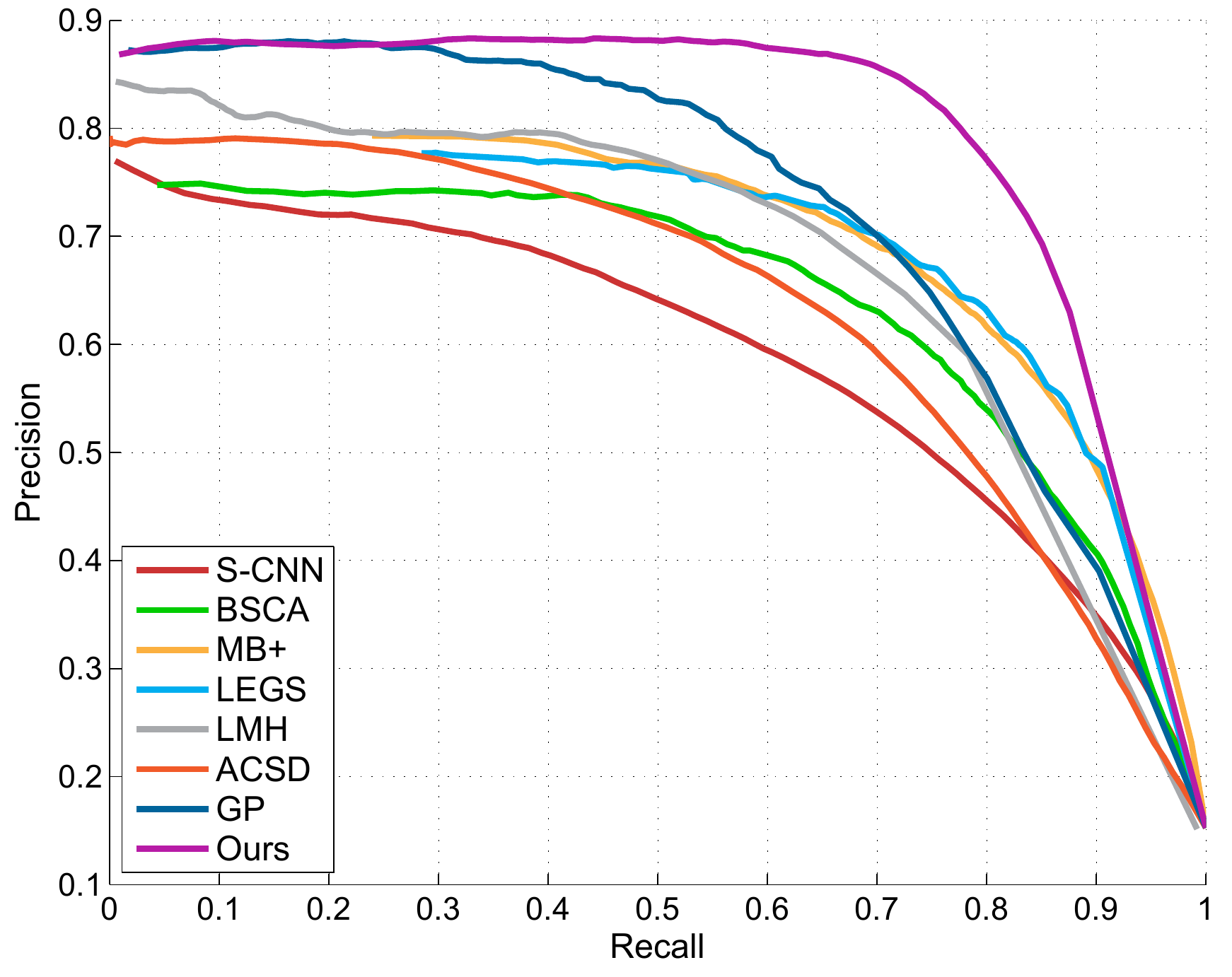}
\includegraphics[width=0.31\linewidth]{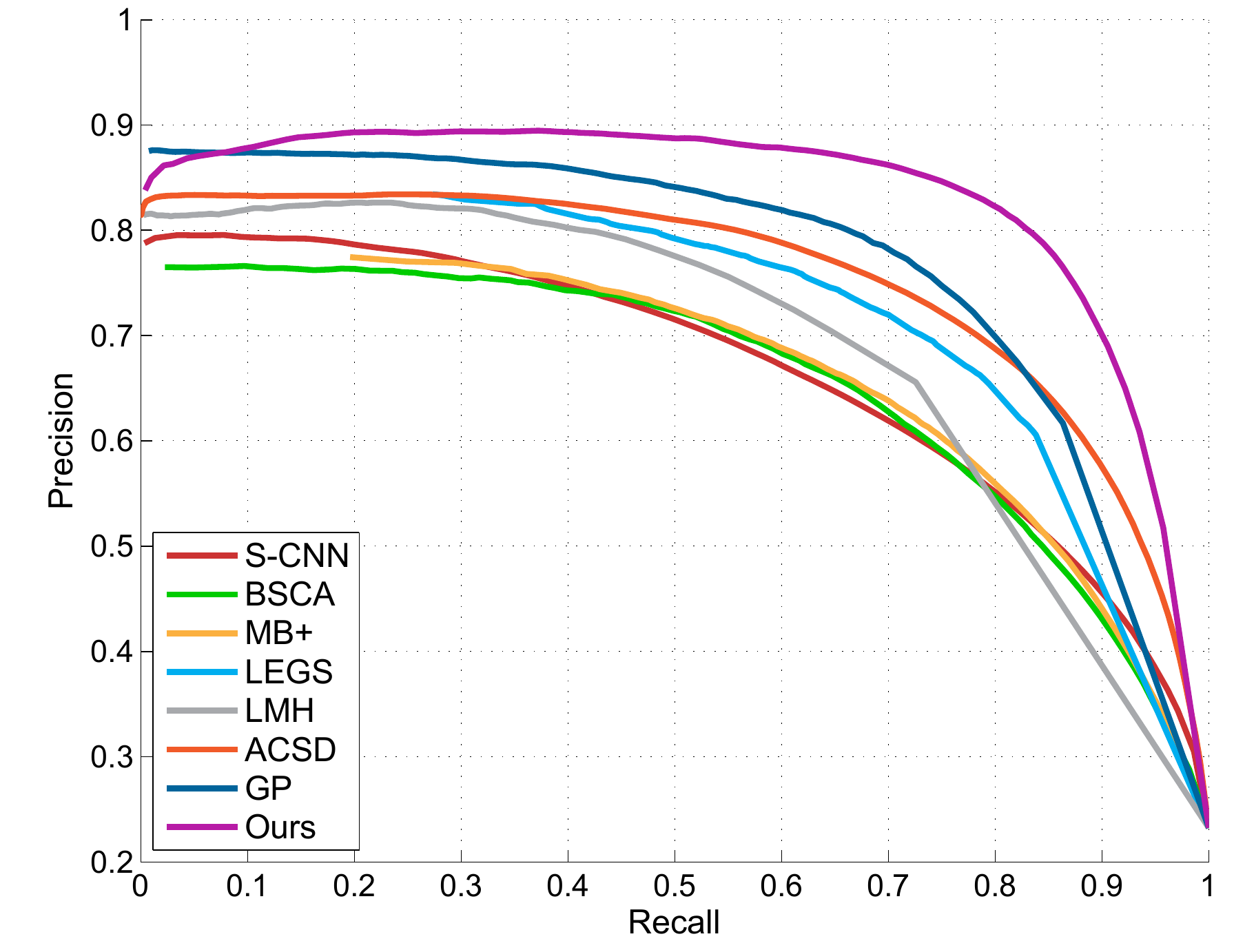}
\includegraphics[width=0.31\linewidth]{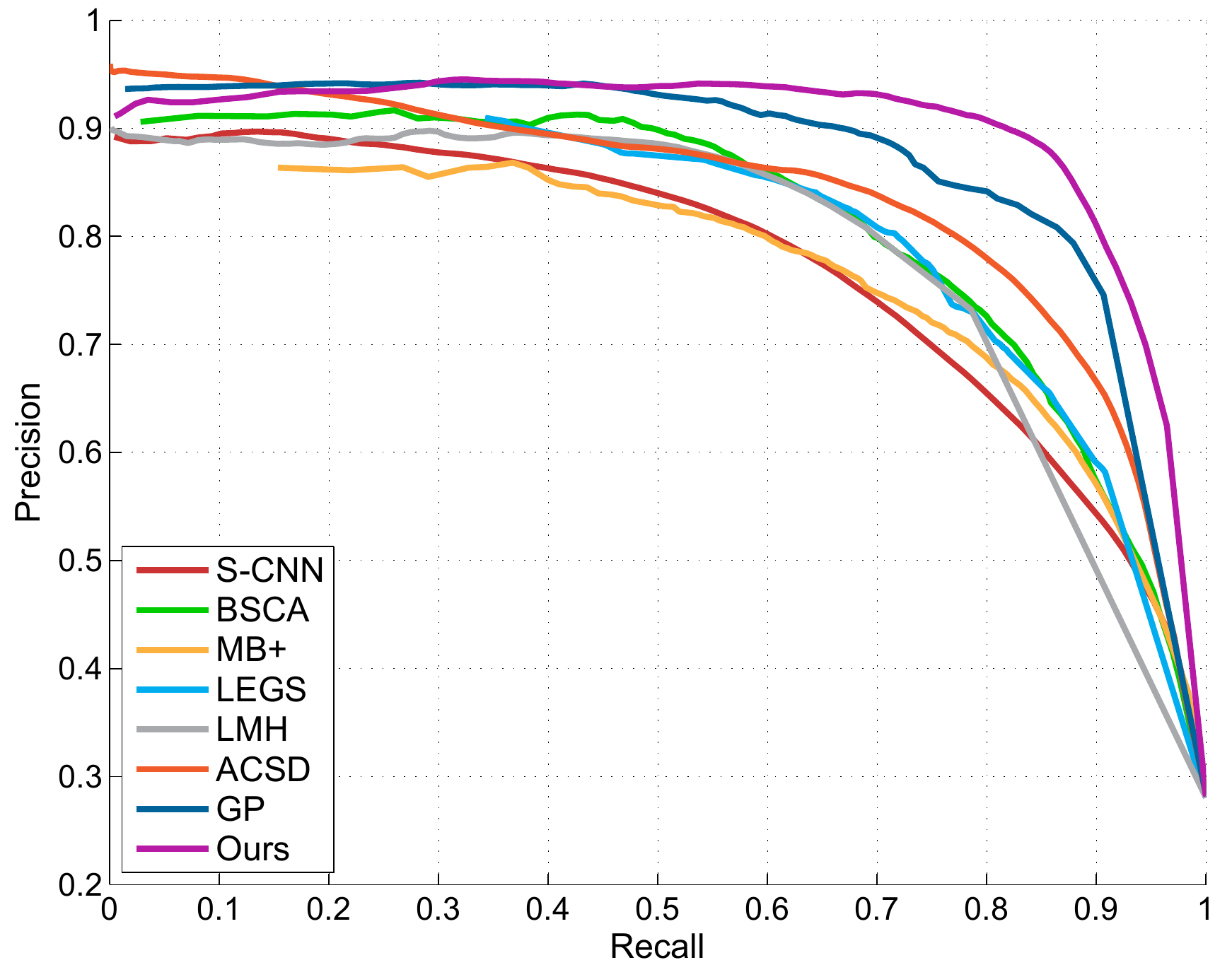}
\includegraphics[width=0.31\linewidth]{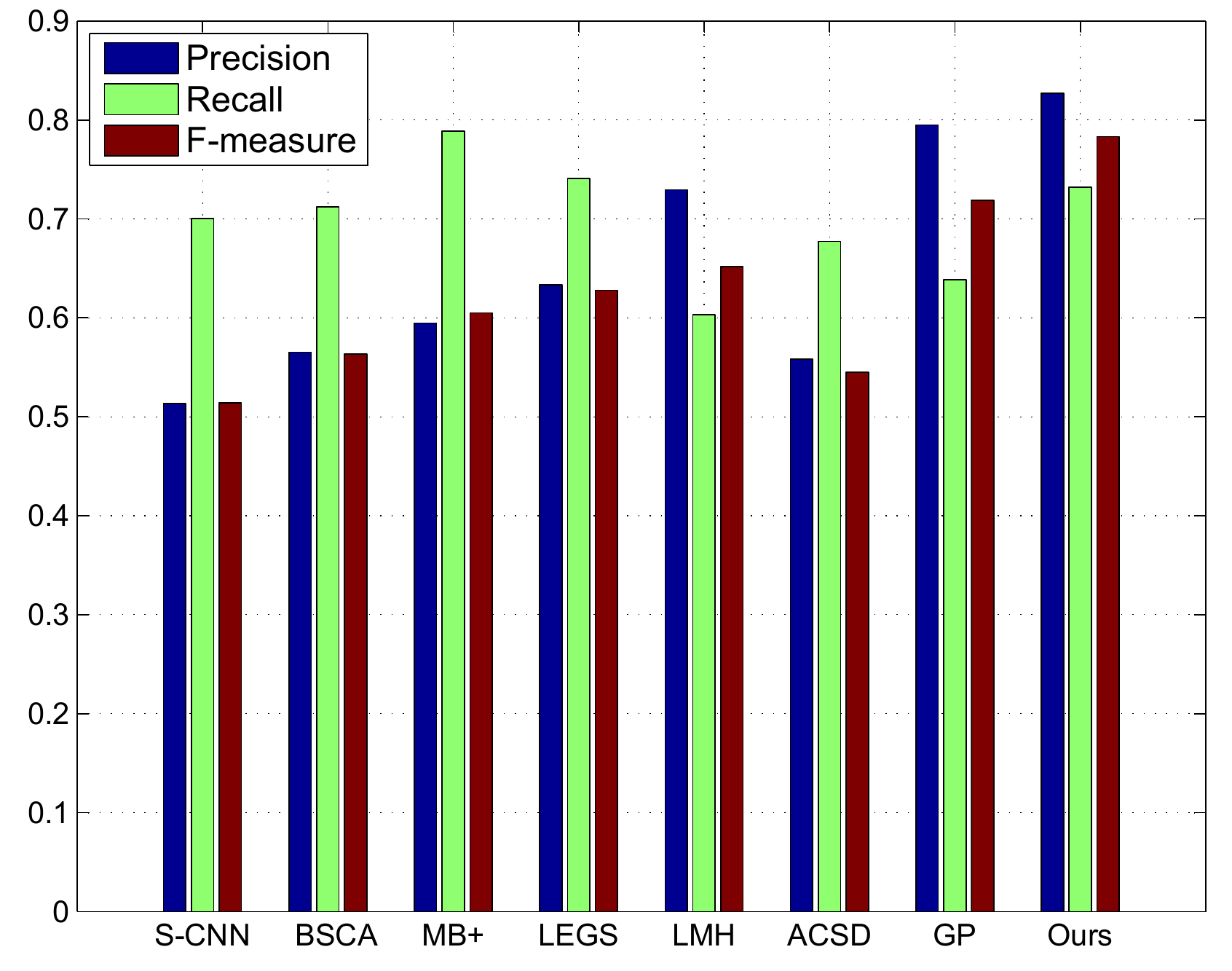}
\includegraphics[width=0.31\linewidth]{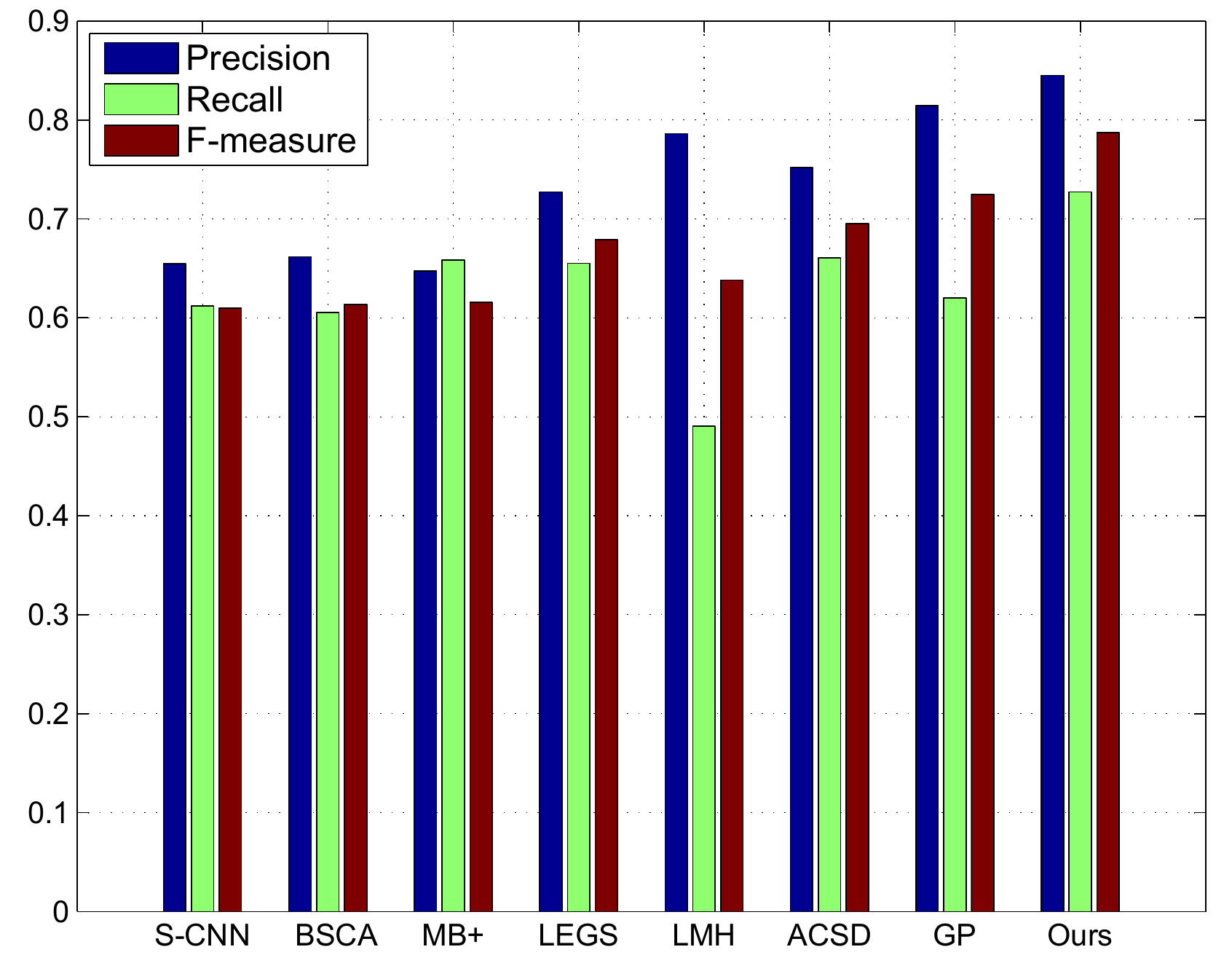}
\includegraphics[width=0.31\linewidth]{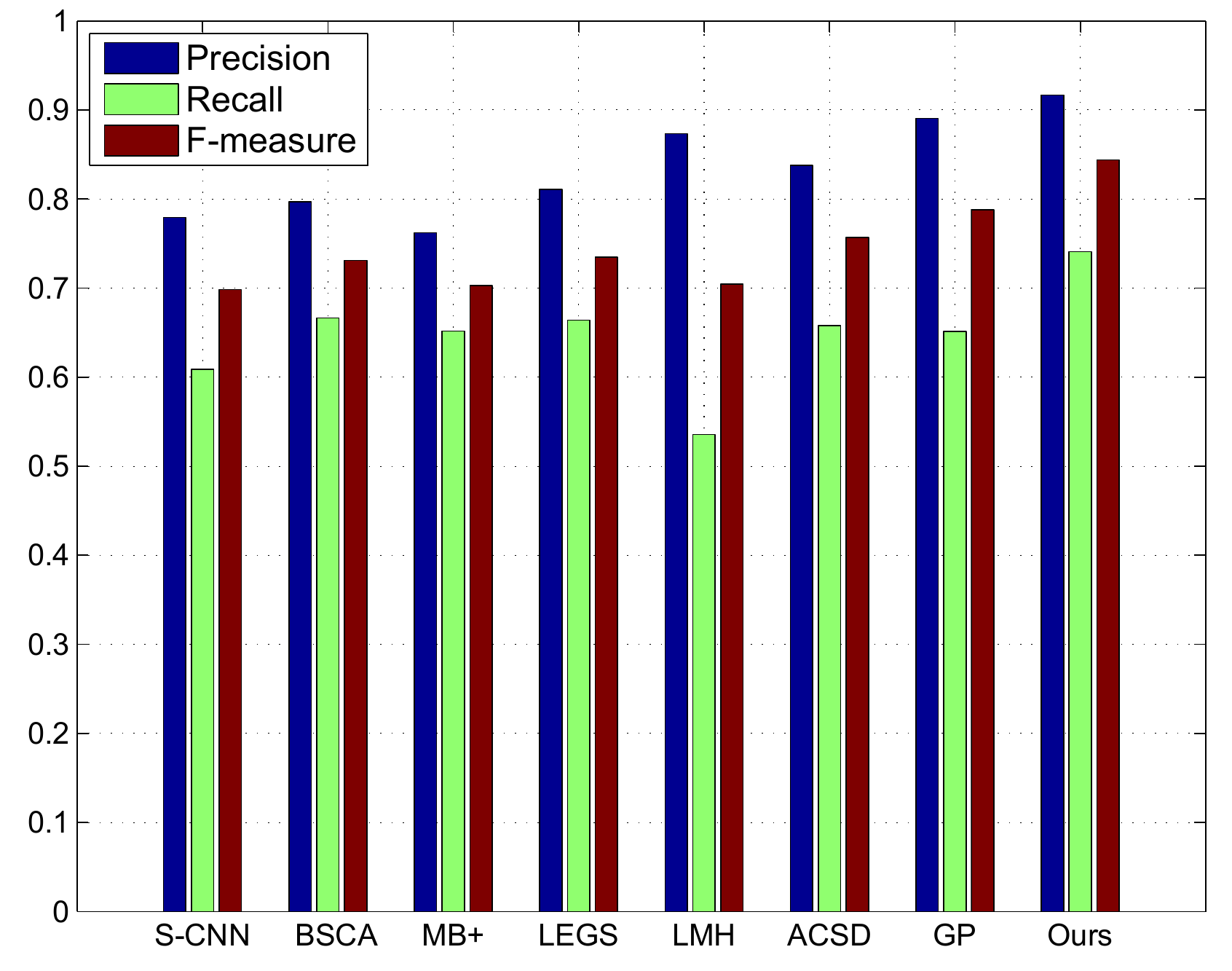}
\caption{PR curves and F-measure curves of different methods on three datasets. Left: quantitative comparisons on the 250 test images of NLPR dataset \cite{peng2014rgbd}. Middle: quantitative comparisons on the 1000 test images of NJUDS2000 dataset \cite{ju2014depth}. Right: quantitative comparisons on the LFSD dataset \cite{Li_2014_CVPR}.}
\label{fig:saliency_qut}
\end{figure*}

\subsection{Performance Comparison}
In this section, we compare our method with four state-of-the-art methods designed for RGB image (
S-CNN \cite{he2015supercnn},  BSCA \cite{qin2015saliency}, MB+ \cite{zhang2015MBD}, and LEGS \cite{wang2015deep}), and three RGBD saliency methods designed specially for RGBD image (LMH \cite{peng2014rgbd}, ACSD \cite{ju2014depth}, and GP \cite{ren2015exploiting}).

The results of these different methods are either provided by authors or achieved using the publicly available source codes. The qualitative comparisons of different methods on different scenes are shown in Figure \ref{fig:saliency2}. As can be seen in the first and fifth rows of Figure \ref{fig:saliency2}, the salient object has a high color contrast with the background, as thus RGB saliency methods are able to detect salient object correctly. However, when the salient object shares similar color with the background, e.g., sixth, seventh, and eighth rows in Figure \ref{fig:saliency2}, it is difficult for existing RGB models to extract saliency. With the help of depth information, salient object can be easily detected by the proposed RGBD method. Figure \ref{fig:saliency2} also shows that the proposed method consistently outperforms all the other RGBD saliency methods (LMH \cite{peng2014rgbd}, ACSD \cite{ju2014depth}, and GP \cite{ren2015exploiting}).

The quantitative comparisons on NLPR, NJUDS2000, and LFSD dataset are shown in Figure \ref{fig:saliency_qut} and Table \ref{table:belta}.
Figure \ref{fig:saliency_qut} and Table \ref{table:belta} show that the proposed method performs favorably against the existing algorithms with higher
precision, recall values and F-measure scores on all the three datasets. For the NLPR dataset, it is challenging as most of the salient object share similar color to the background. As a consequence, RGB saliency methods perform relative worse than RGBD saliency methods in terms of precision. By providing accurate depth map (NLPR dataset), LMH \cite{peng2014rgbd} and GP \cite{ren2015exploiting} methods perform well in both precision and recall. However, they performs not well when tested on the NJUDS2000 dataset and LFSD dataset. This is because these two datasets provide only the rough depth information (calculated from stereo images or using Light field camera), LMH \cite{peng2014rgbd} and GP \cite{ren2015exploiting} can only detect a small fraction of the salient objects (high precision but with low recall). ACSD \cite{ju2014depth} works worse when the salient object lies in the same plane with the background, e.g., the third row in the Figure \ref{fig:saliency2}, and the bad quantitative results on the NLPR dataset. Both qualitative and quantitative results show that the proposed method performs better in terms of accuracy and robustness than the compared methods with RGBD input images.

\begin{table*}
\centering
\caption{The comparisons of F-measure scores for different saliency map merging approaches with or without LP on NLPR test dataset \cite{peng2014rgbd}.}
\label{table:anlysis_NLPR}
\begin{tabular}{|c|c|c|c|cIc|c|c|cIc|cIc|}
\hline
\multirow{2}{*}{LP$?$} & \multicolumn{4}{cI}{\bf{Fundamental fusion}} & \multicolumn{4}{cI}{\bf{Sophisticated fusion}} & \multicolumn{2}{cI}{{Heuristic fusion}}& \multirow{2}{*}{Ours}\\
\cline{2-11}
& LF & CRF & MCA & CNN-F & LF & CRF & MCA & CNN-F &LMH &GP& \\
\hline\hline
{No} &  0.393 & 0.2991 & 0.3713 & 0.4667 & 0.7020 & 0.698 & 0.7017 & 0.6921 &0.6519 & \bf{0.718} &{\color{blue}{0.7315}} \\
{Yes} & \bf{0.536} & \bf{0.398} & \bf{0.486} & \bf{0.597} & \bf{0.711} & \bf{0.739} &\bf{0.7623} &\bf{0.737} &\bf{0.665}&0.7111&{\color{red}\bf{0.7823}}\\
\hline
\end{tabular}
\end{table*}

\begin{table*}
\centering
\caption{The comparisons of F-measure scores for different saliency map merging approaches with or without Laplacian propagation (LP) on NJUD test dataset \cite{ju2014depth}.}
\label{table:anlysis}
\begin{tabular}{|c|c|c|c|cIc|c|c|cIc|cIc|}
\hline
\multirow{2}{*}{LP$?$} & \multicolumn{4}{cI}{\bf{Fundamental fusion}} & \multicolumn{4}{cI}{\bf{Sophisticated fusion}} & \multicolumn{2}{cI}{{Heuristic fusion}}& \multirow{2}{*}{Ours}\\
\cline{2-11}
& LF & CRF & MCA & CNN-F & LF & CRF & MCA & CNN-F &LMH &GP& \\
\hline\hline
{No} &  0.437 & 0.450 & 0.458 & 0.644 & 0.675 & 0.671 &0.7376 & 0.7319& {0.6381} &\bf{0.7246} &{\color{blue}{0.7447}} \\
{Yes} & \bf{0.605} & \bf{0.609} & \bf{0.632} & \bf{0.731} & \bf{0.698} & \bf{0.741} &\bf{0.742} &\bf{0.7423} &\bf{0.6810}&0.7179&{\color{red}\bf{0.7874}}\\
\hline
\end{tabular}
\end{table*}

\begin{table*}
\centering
\caption{The comparisons of F-measure scores for different saliency map merging approaches with or without LP on LFSD dataset \cite{Li_2014_CVPR}.}
\label{table:anlysis_LFSD}
\begin{tabular}{|c|c|c|c|cIc|c|c|cIc|cIc|}
\hline
\multirow{2}{*}{LP$?$} & \multicolumn{4}{cI}{\bf{Fundamental fusion}} & \multicolumn{4}{cI}{\bf{Sophisticated fusion}} & \multicolumn{2}{cI}{{Heuristic fusion}}& \multirow{2}{*}{Ours}\\
\cline{2-11}
& LF & CRF & MCA & CNN-F & LF & CRF & MCA & CNN-F &LMH &GP& \\
\hline\hline
{No} &  0.461 & 0.436 & 0.558 & 0.672 & 0.723 & 0.771 & 0.8071 & 0.706 &0.704 & \bf{0.7877} &{\color{blue}{0.8157}} \\
{Yes} & \bf{0.616} & \bf{0.693} & \bf{0.654} & \bf{0.757} & \bf{0.762} & \bf{0.792} & \bf{0.802} &\bf{0.800} &\bf{0.718} &{0.7830}&{\color{red}\bf{0.8439}}\\
\hline
\end{tabular}
\end{table*}

\noindent{\textbf{Saliency maps vs. features.}}
In here we conduct a series of experiments to analyze the flexibility of the proposed framework and the effectiveness of Laplacian propagation.

Apart from previous heuristic saliency map merging algorithm \cite{peng2014rgbd,ren2015exploiting}, we further compare our method with four other
saliency map integration methods on three test dataset \cite{peng2014rgbd,ju2014depth,Li_2014_CVPR} to show the flexibility of fusing different cues in feature level.  These four integration methods are directly linear fusion (LF), fusing in CRF \cite{liu2011learning}, the latest Multi-layer Cellular Automata (MCA) integration \cite{qin2015saliency}, and a CNN based fusion (denoted as CNN-F).
To investigate the importance of saliency map quality, we test these saliency map merging methods on two set of inputs. The first set is from seven saliency maps computed by widely used features (similar to those seven saliency feature vectors computed in section \ref{title_3_1}), and the second set is from more representative sophisticated saliency maps (obtained using three state-of-the-art RGBD saliency detection methods \cite{peng2014rgbd,ju2014depth,ren2015exploiting}).

The original CRF fusion framework in \cite{liu2011learning} is utilized for merging three color based saliency maps. In our implementation, we retrain this CRF framework for merging the seven adopted fundamental saliency maps and three sophisticated saliency maps respectively \footnote{ We adopt the implementation in \url{http://www.cs.unc.edu/~vicente/code.html} for training and testing. This CRF is trained on the NLPR training dataset \cite{peng2014rgbd} and the NJUDS2000 training dataset \cite{ju2014depth}}.

For CNN-F, we utilize the same CNN architecture as shown in Fig. \ref{fig:saliency_cue} to perform the CNN based saliency map fusion, i.e., the same convolutional layers and fully connected layers except the input layer.
More specifically, we formulate the saliency map merging as a binary logistic regression problem, which takes several saliency map patches as input (size $52\times52\times7$ for fundamental saliency map merging and $52\times52\times3$ for sophisticated saliency map merging), and output the probabilities of the pixel being salient and non-salient. CNN-F is trained in patch-wise manner. We collect training samples by cropping patches of size $52\times52$ from each saliency map using sliding window. 
We label a patch as salient if the central pixel is salient or 75\% pixels in this patch are salient, otherwise it is labeled as non-salient. This CNN-F is trained on the cropped patches of the NLPR training set \cite{peng2014rgbd} and NJUDS2000 training set \cite{ju2014depth}.

\begin{figure*}
\centering
\captionsetup[subfigure]{labelformat=empty}

\vspace{-1.5mm}
\subfloat{ \includegraphics[width=\widthtwelve]{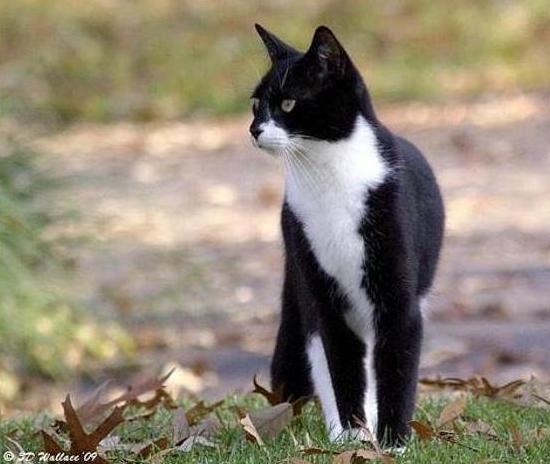}} \hspacefigure
\subfloat{ \includegraphics[width=\widthtwelve]{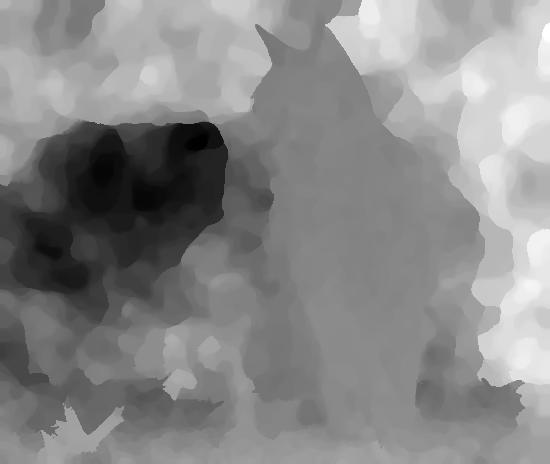}}\hspacefigure
\subfloat{ \includegraphics[width=\widthtwelve]{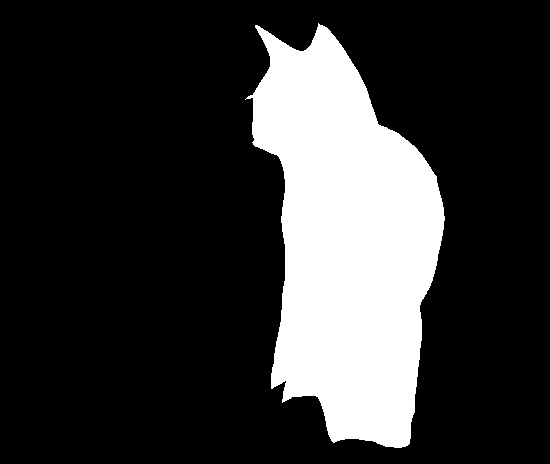}}\hspacefigure
\subfloat{ \includegraphics[width=\widthtwelve]{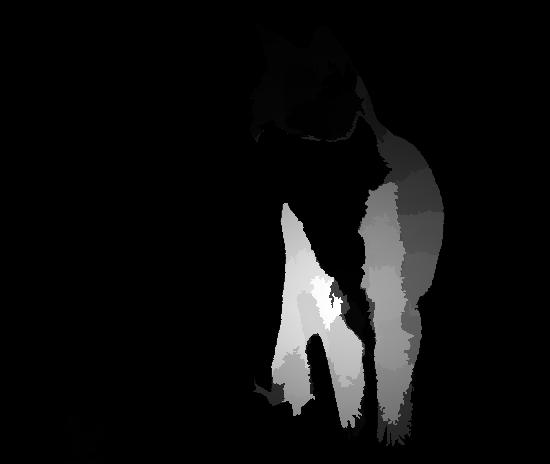}} \hspacefigure
\subfloat{ \includegraphics[width=\widthtwelve]{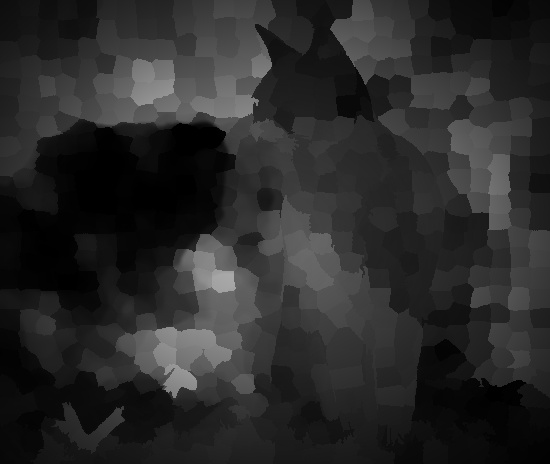}} \hspacefigure
\subfloat{ \includegraphics[width=\widthtwelve]{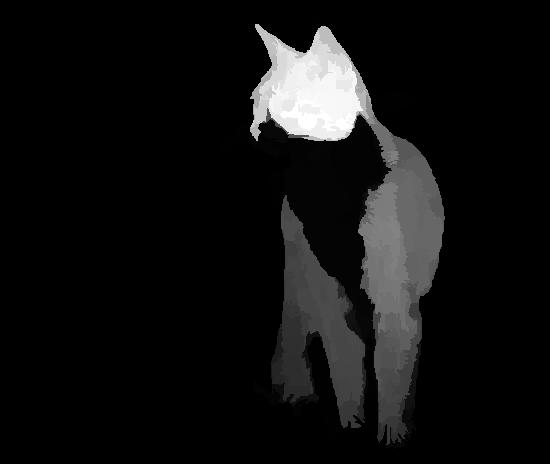}} \hspacefigure
\subfloat{ \includegraphics[width=\widthtwelve]{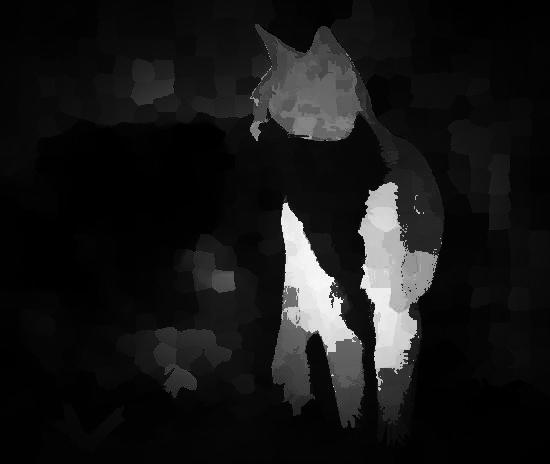}} \hspacefigure
\subfloat{ \includegraphics[width=\widthtwelve]{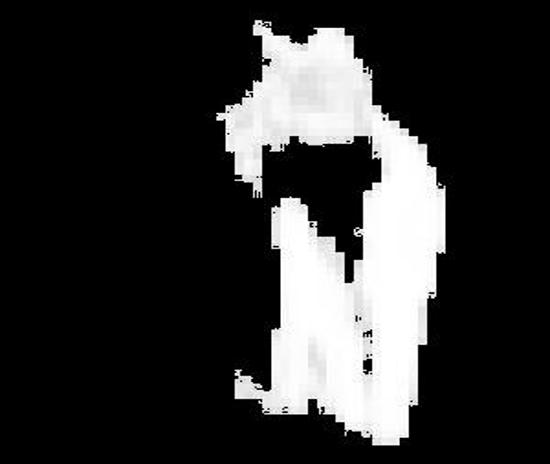}} \hspacefigure
\subfloat{ \includegraphics[width=\widthtwelve]{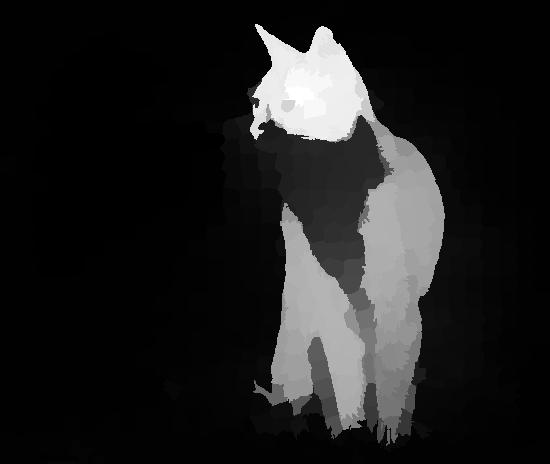}} \hspacefigure
\subfloat{ \includegraphics[width=\widthtwelve]{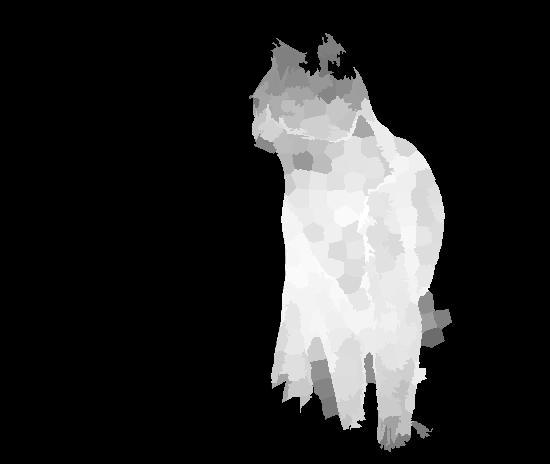}} \hspacefigure
\subfloat{ \includegraphics[width=\widthtwelve]{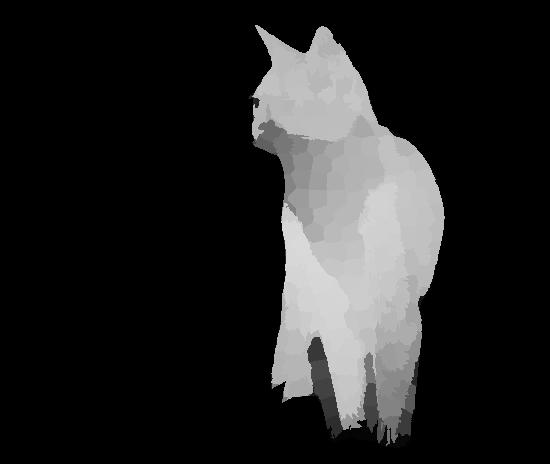}} \hspacefigure \\
\vspace{-1.5mm}
\subfloat{ \includegraphics[width=\widthtwelve,height=0.091\linewidth]{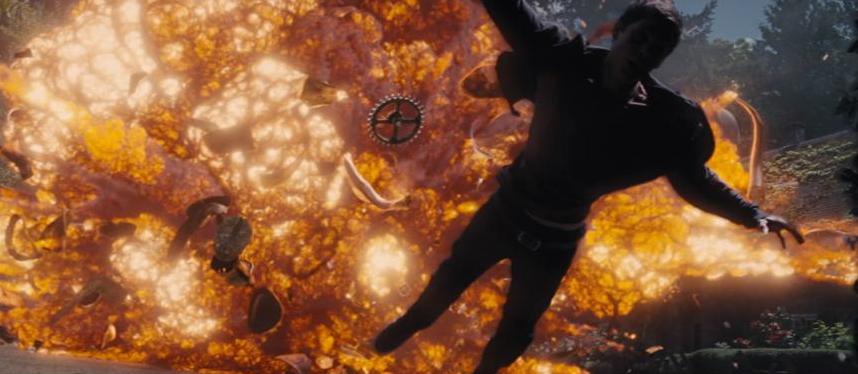}} \hspacefigure
\subfloat{ \includegraphics[width=\widthtwelve,height=0.091\linewidth]{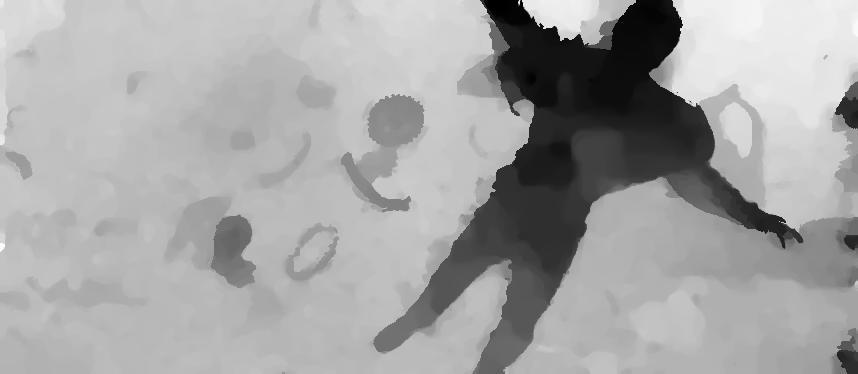}}\hspacefigure
\subfloat{ \includegraphics[width=\widthtwelve,height=0.091\linewidth]{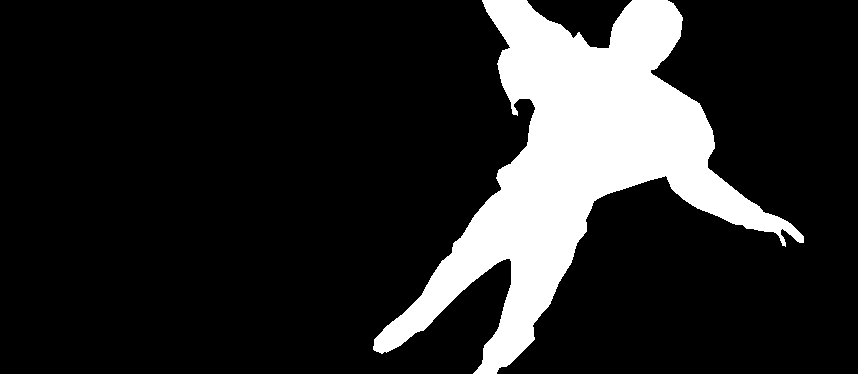}}\hspacefigure
\subfloat{ \includegraphics[width=\widthtwelve,height=0.091\linewidth]{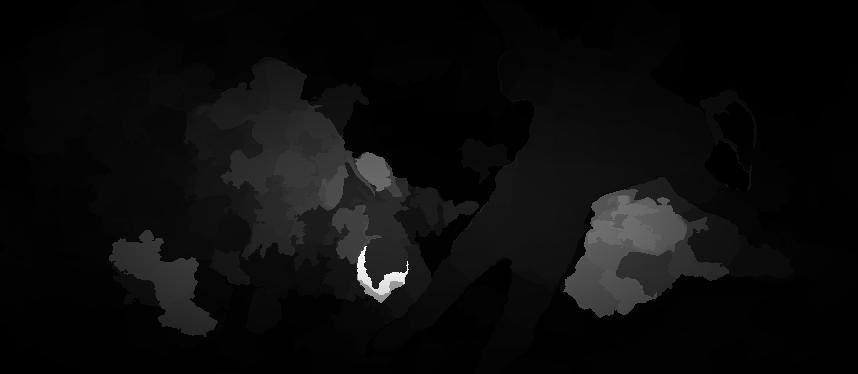}} \hspacefigure
\subfloat{ \includegraphics[width=\widthtwelve,height=0.091\linewidth]{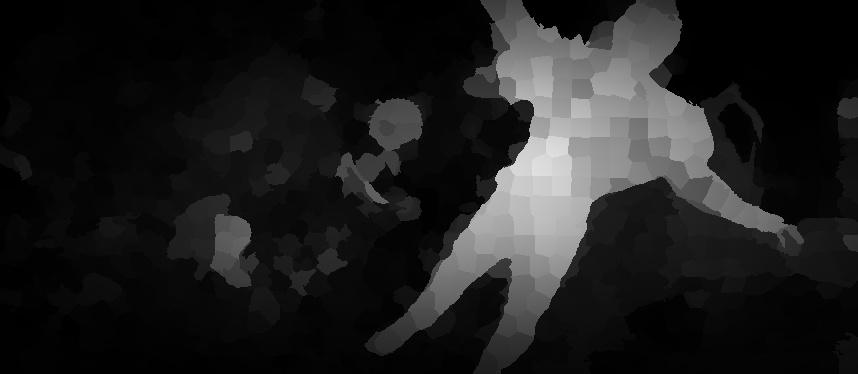}} \hspacefigure
\subfloat{ \includegraphics[width=\widthtwelve,height=0.091\linewidth]{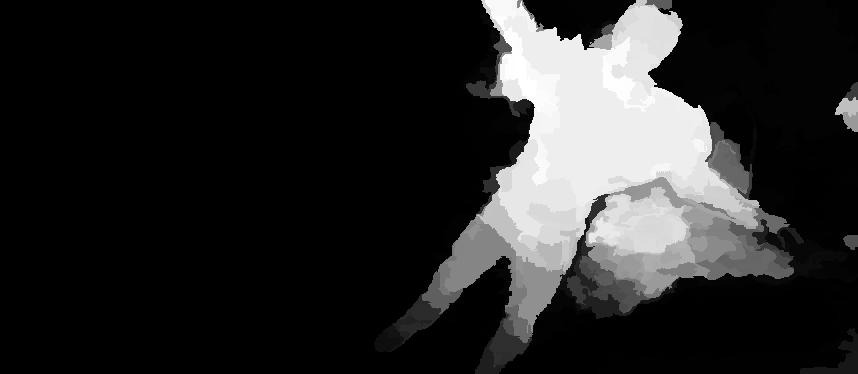}} \hspacefigure
\subfloat{ \includegraphics[width=\widthtwelve,height=0.091\linewidth]{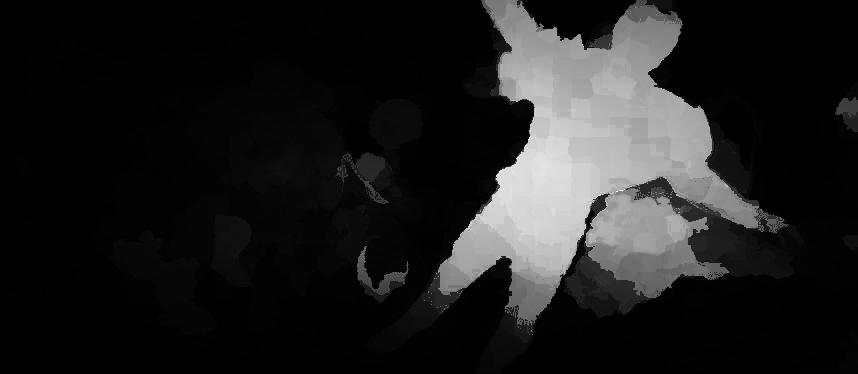}} \hspacefigure
\subfloat{ \includegraphics[width=\widthtwelve,height=0.091\linewidth]{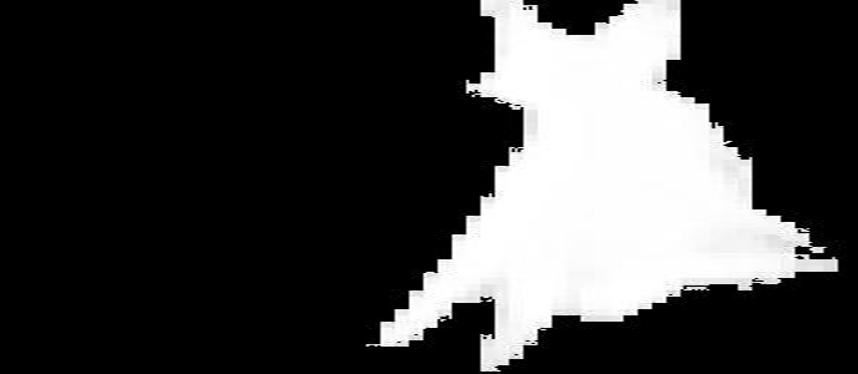}} \hspacefigure
\subfloat{ \includegraphics[width=\widthtwelve,height=0.091\linewidth]{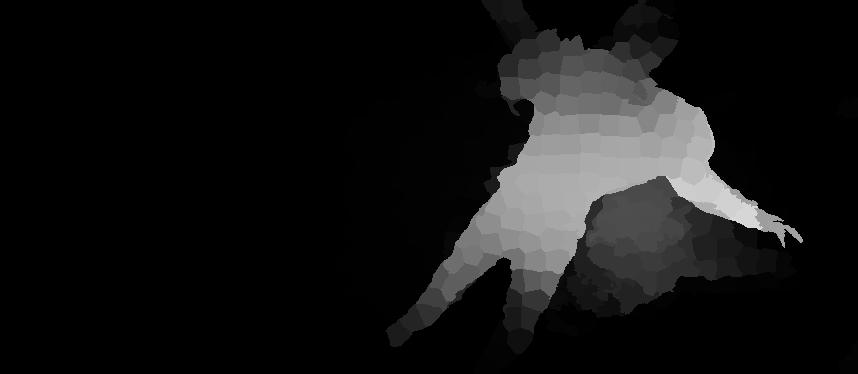}} \hspacefigure
\subfloat{ \includegraphics[width=\widthtwelve,height=0.091\linewidth]{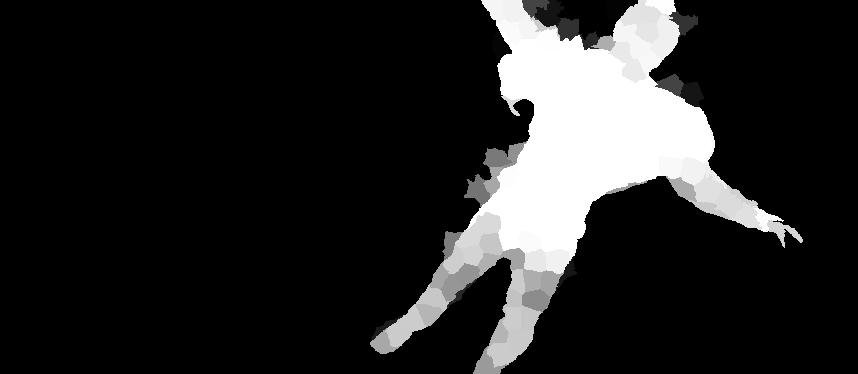}} \hspacefigure
\subfloat{ \includegraphics[width=\widthtwelve,height=0.091\linewidth]{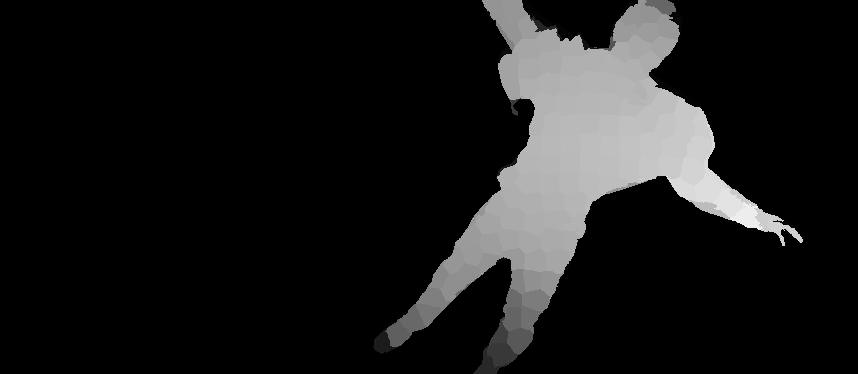}} \hspacefigure \\
\vspace{-1.5mm}
\subfloat{ \includegraphics[width=\widthtwelve,height=0.095\linewidth]{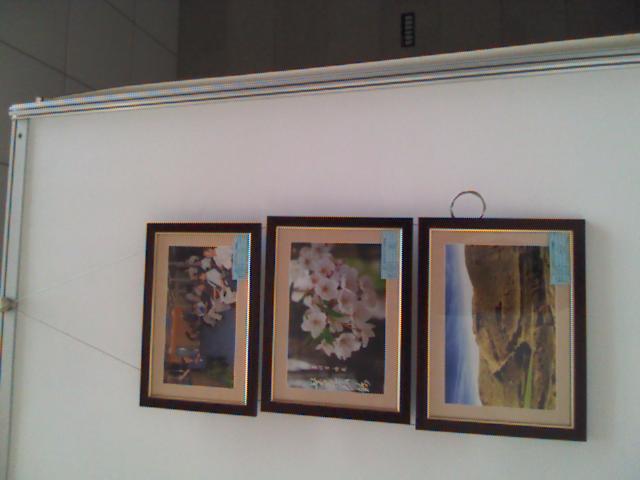}} \hspacefigure
\subfloat{ \includegraphics[width=\widthtwelve,height=0.095\linewidth]{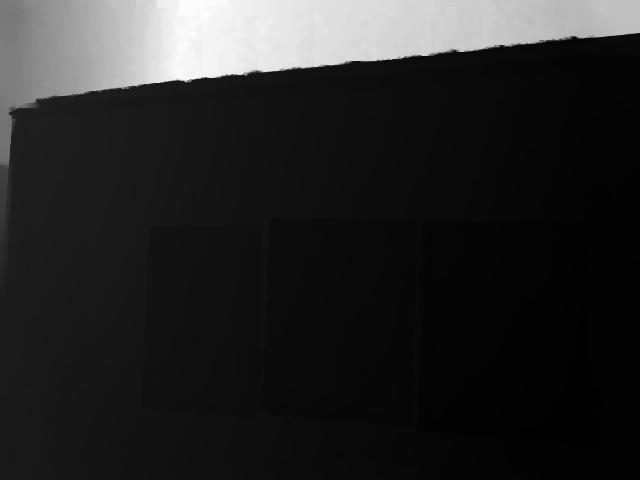}}\hspacefigure
\subfloat{ \includegraphics[width=\widthtwelve,height=0.095\linewidth]{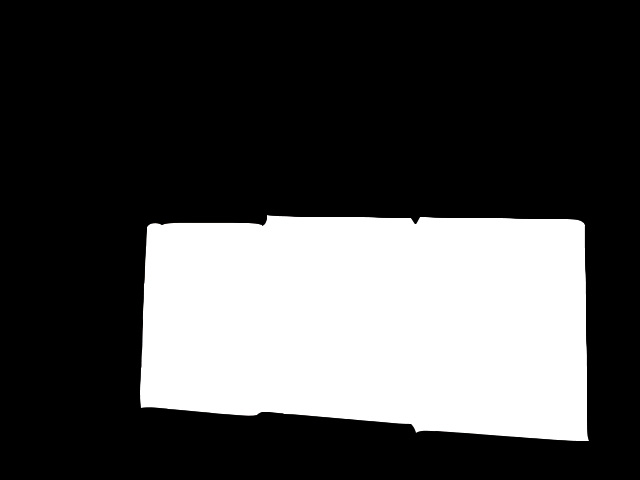}}\hspacefigure
\subfloat{ \includegraphics[width=\widthtwelve,height=0.095\linewidth]{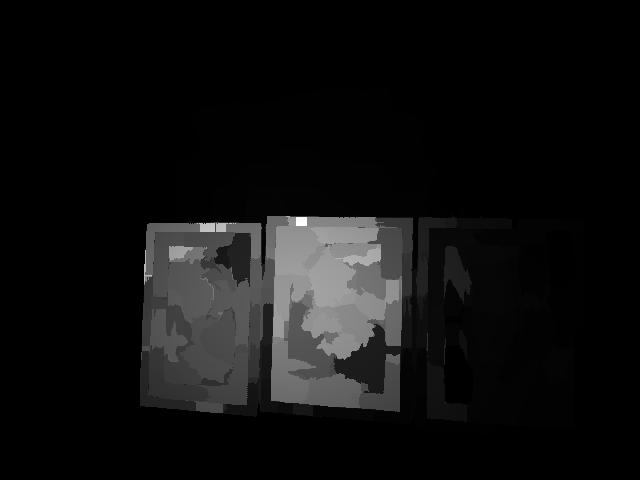}} \hspacefigure
\subfloat{ \includegraphics[width=\widthtwelve,height=0.095\linewidth]{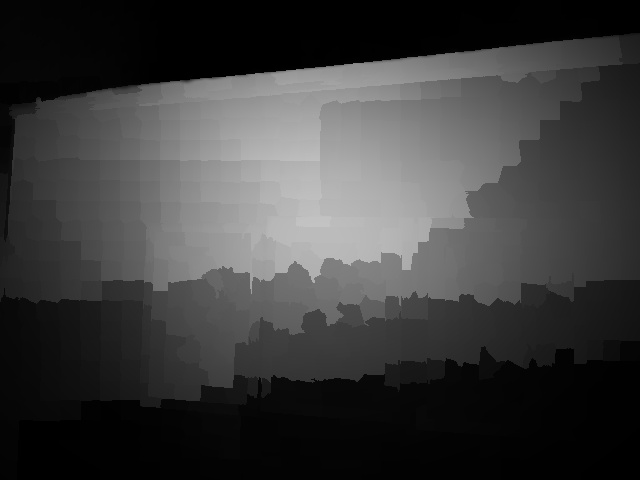}} \hspacefigure
\subfloat{ \includegraphics[width=\widthtwelve,height=0.095\linewidth]{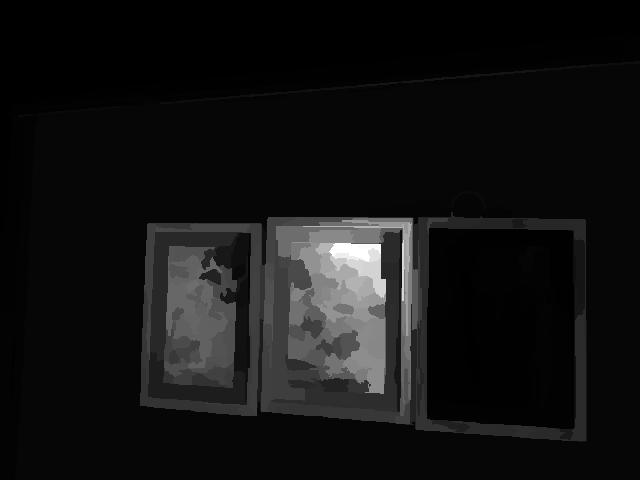}} \hspacefigure
\subfloat{ \includegraphics[width=\widthtwelve,height=0.095\linewidth]{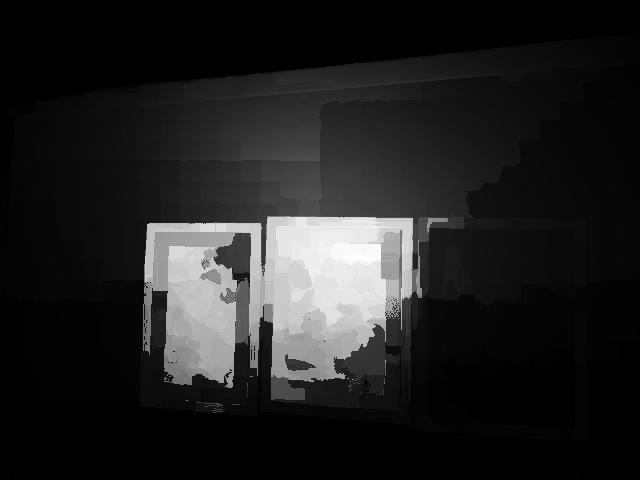}} \hspacefigure
\subfloat{ \includegraphics[width=\widthtwelve,height=0.095\linewidth]{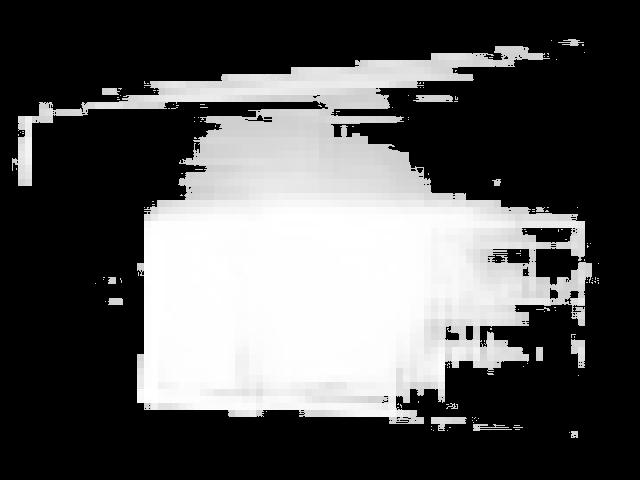}} \hspacefigure
\subfloat{ \includegraphics[width=\widthtwelve,height=0.095\linewidth]{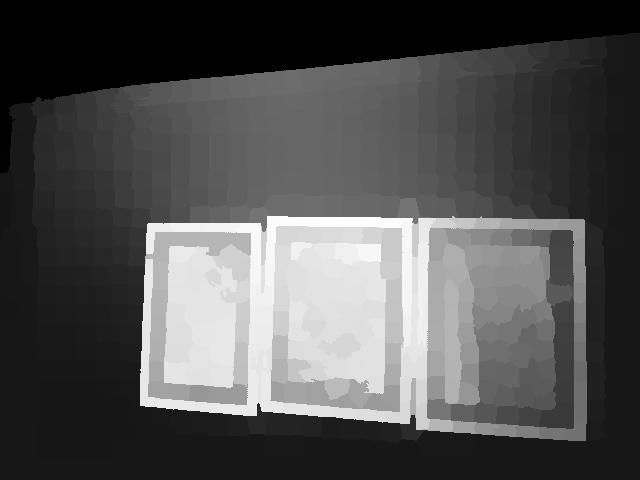}} \hspacefigure
\subfloat{ \includegraphics[width=\widthtwelve,height=0.095\linewidth]{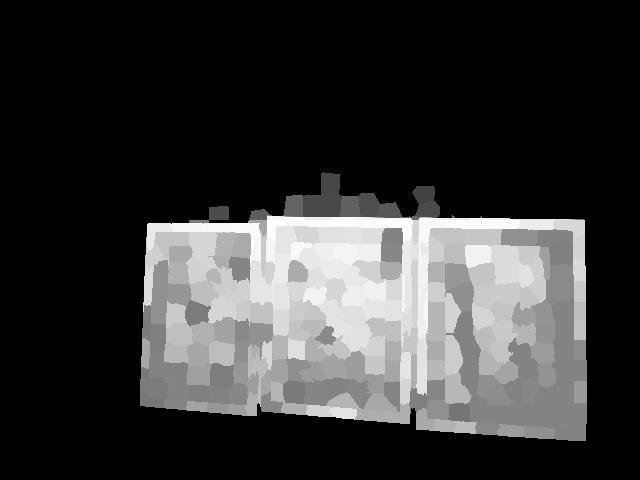}} \hspacefigure
\subfloat{ \includegraphics[width=\widthtwelve,height=0.095\linewidth]{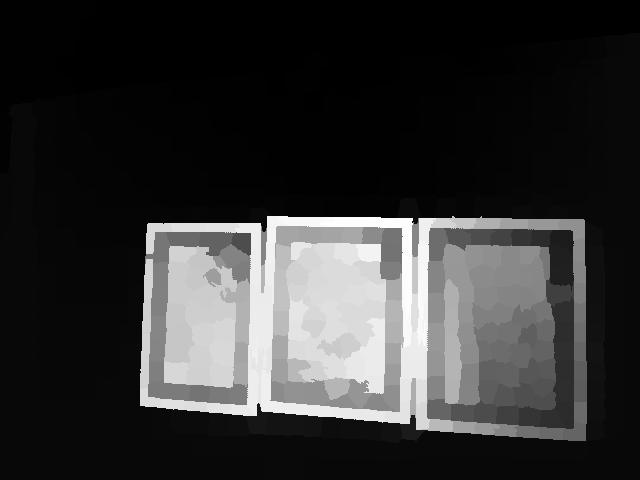}} \hspacefigure \\
\vspace{-1.5mm}
\subfloat{ \includegraphics[width=\widthtwelve,height=0.09\linewidth]{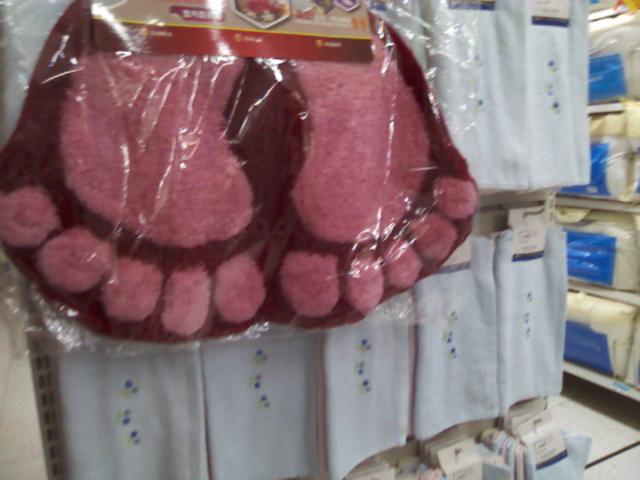}} \hspacefigure
\subfloat{ \includegraphics[width=\widthtwelve,height=0.09\linewidth]{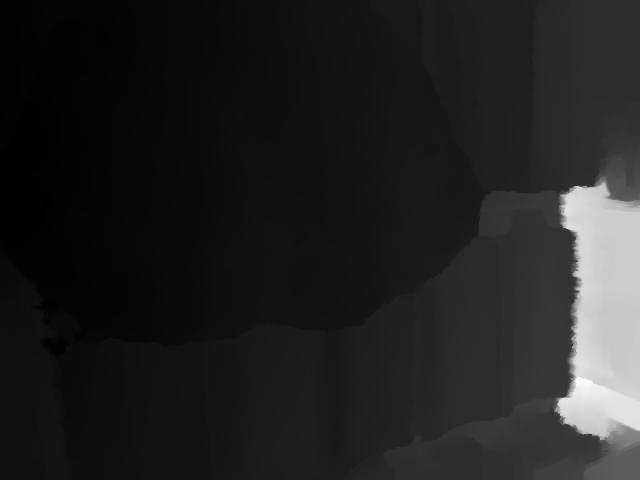}}\hspacefigure
\subfloat{ \includegraphics[width=\widthtwelve,height=0.09\linewidth]{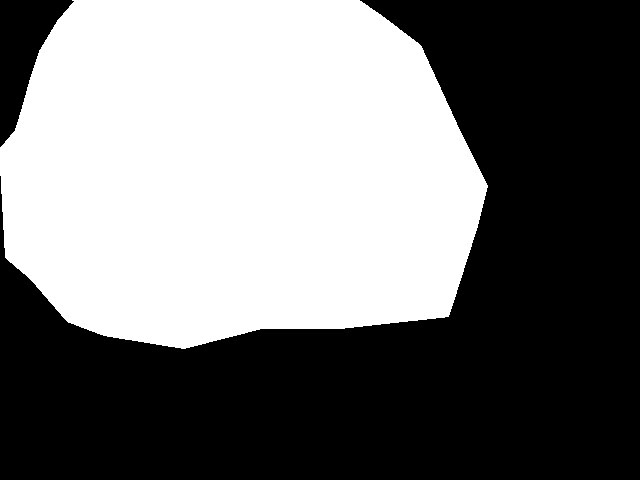}}\hspacefigure
\subfloat{ \includegraphics[width=\widthtwelve,height=0.09\linewidth]{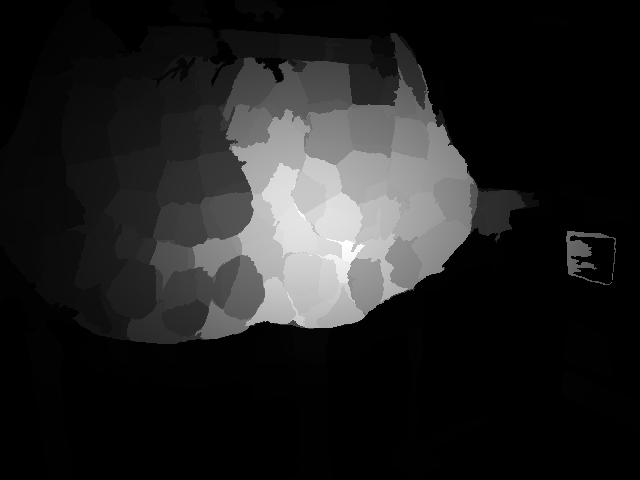}} \hspacefigure
\subfloat{ \includegraphics[width=\widthtwelve,height=0.09\linewidth]{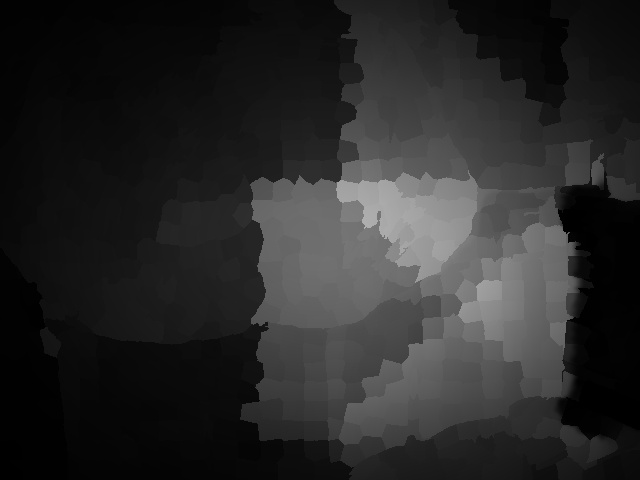}} \hspacefigure
\subfloat{ \includegraphics[width=\widthtwelve,height=0.09\linewidth]{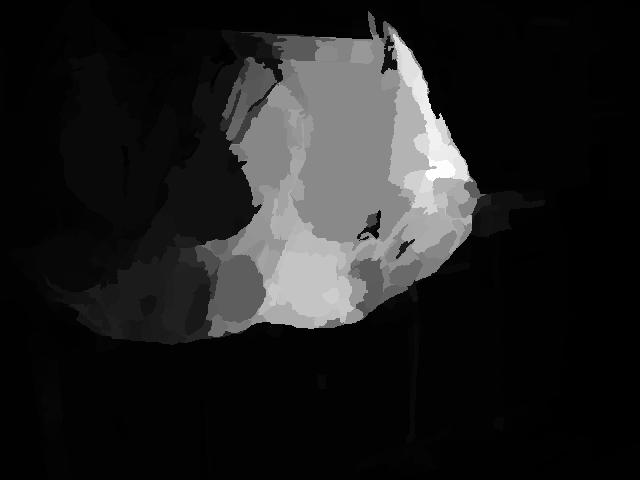}} \hspacefigure
\subfloat{ \includegraphics[width=\widthtwelve,height=0.09\linewidth]{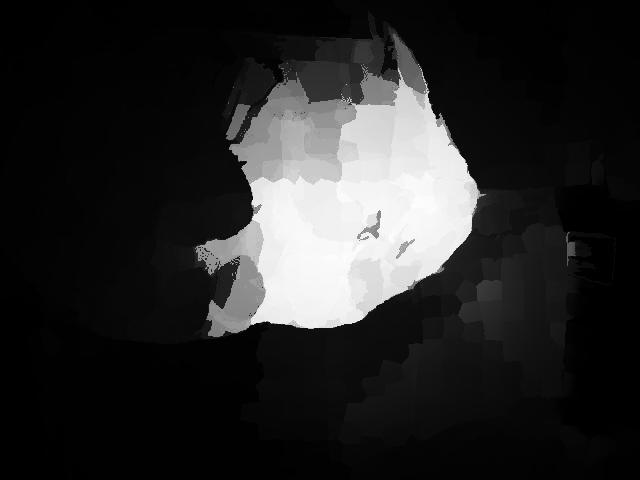}} \hspacefigure
\subfloat{ \includegraphics[width=\widthtwelve,height=0.09\linewidth]{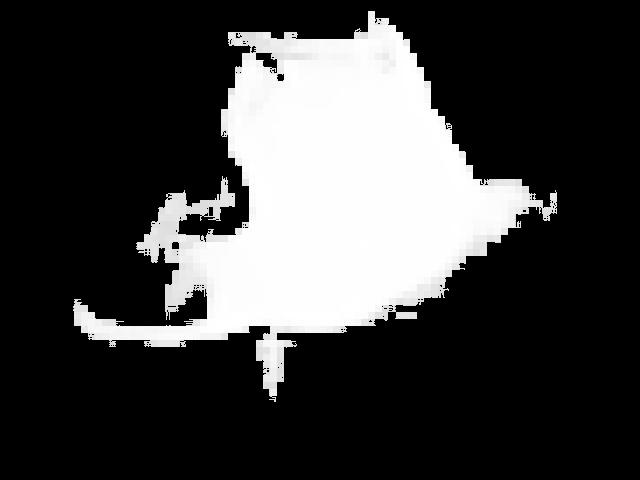}} \hspacefigure
\subfloat{ \includegraphics[width=\widthtwelve,height=0.09\linewidth]{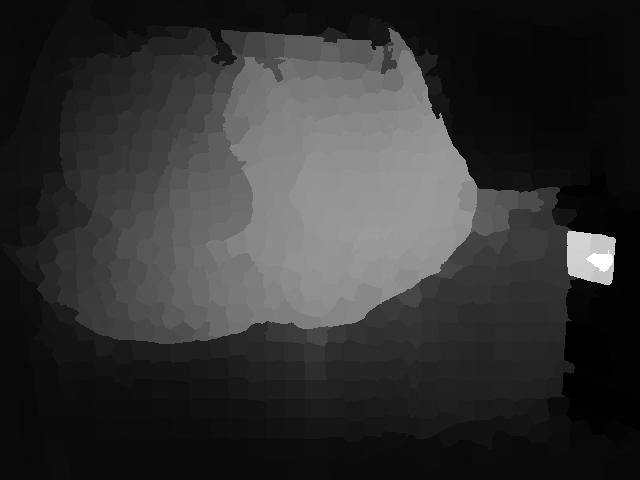}} \hspacefigure
\subfloat{ \includegraphics[width=\widthtwelve,height=0.09\linewidth]{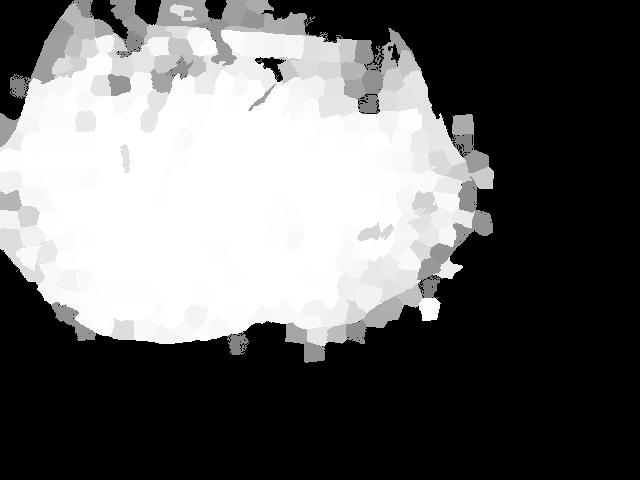}} \hspacefigure
\subfloat{ \includegraphics[width=\widthtwelve,height=0.09\linewidth]{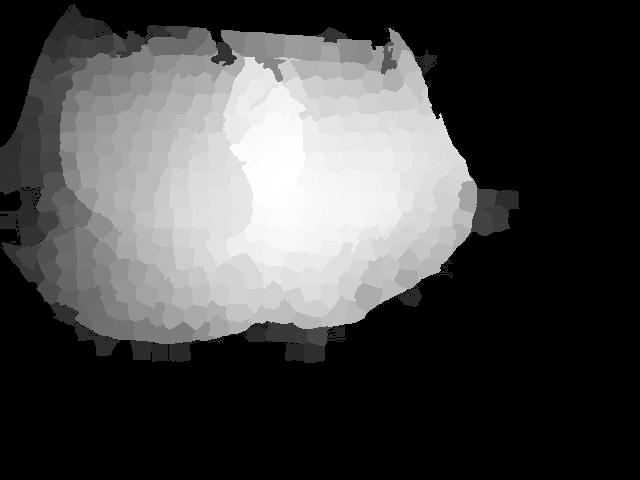}} \hspacefigure \\
\vspace{-1.5mm}
\subfloat{ \includegraphics[width=\widthtwelve,height=0.08\linewidth]{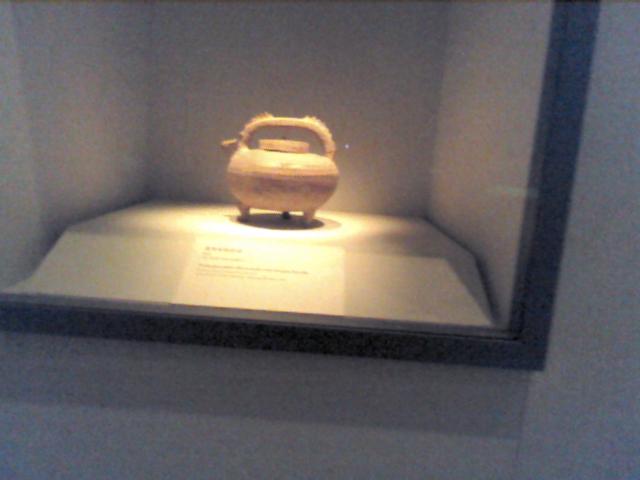}} \hspacefigure
\subfloat{ \includegraphics[width=\widthtwelve,height=0.08\linewidth]{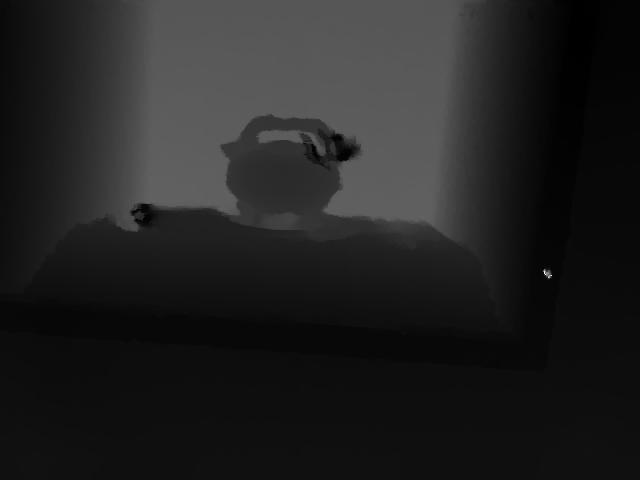}}\hspacefigure
\subfloat{ \includegraphics[width=\widthtwelve,height=0.08\linewidth]{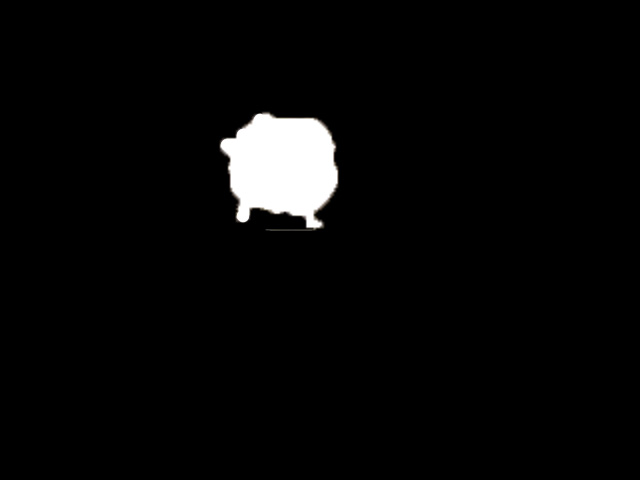}}\hspacefigure
\subfloat{ \includegraphics[width=\widthtwelve,height=0.08\linewidth]{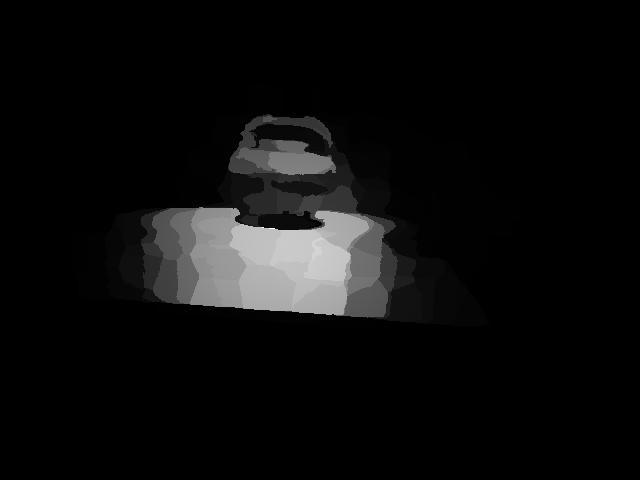}} \hspacefigure
\subfloat{ \includegraphics[width=\widthtwelve,height=0.08\linewidth]{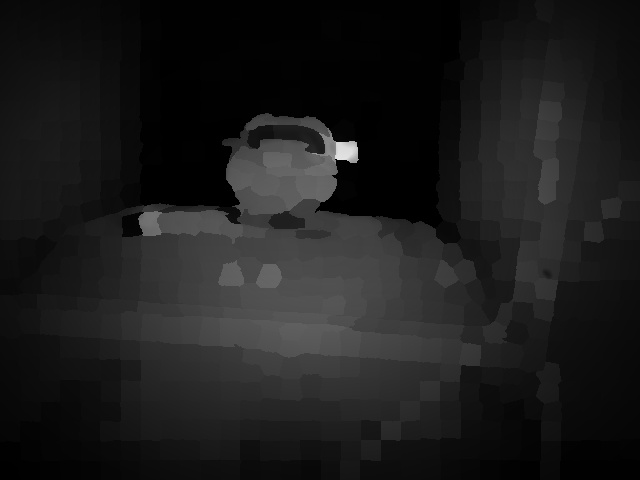}} \hspacefigure
\subfloat{ \includegraphics[width=\widthtwelve,height=0.08\linewidth]{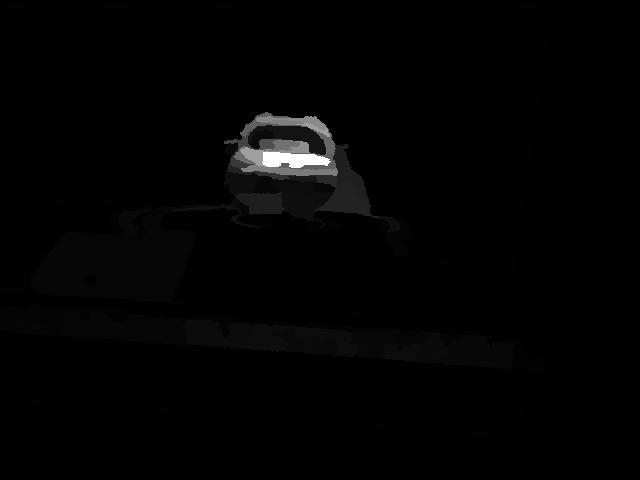}} \hspacefigure
\subfloat{ \includegraphics[width=\widthtwelve,height=0.08\linewidth]{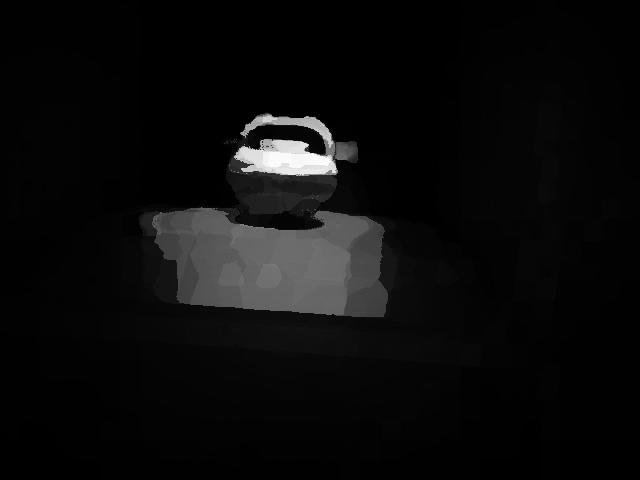}} \hspacefigure
\subfloat{ \includegraphics[width=\widthtwelve,height=0.08\linewidth]{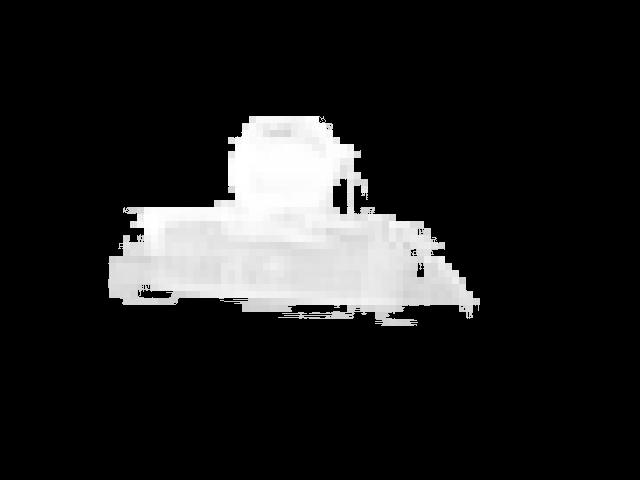}} \hspacefigure
\subfloat{ \includegraphics[width=\widthtwelve,height=0.08\linewidth]{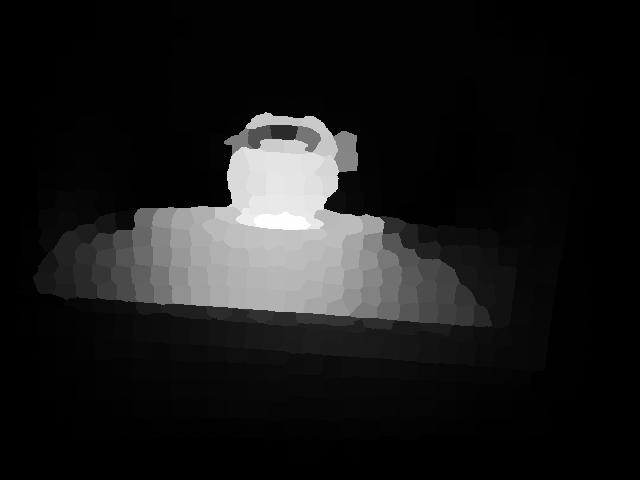}} \hspacefigure
\subfloat{ \includegraphics[width=\widthtwelve,height=0.08\linewidth]{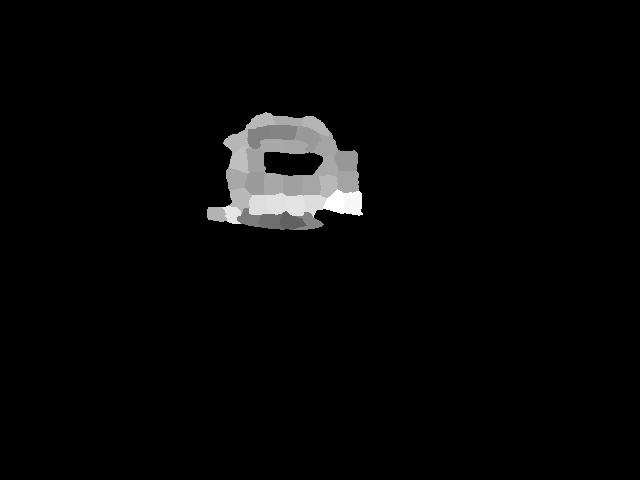}} \hspacefigure
\subfloat{ \includegraphics[width=\widthtwelve,height=0.08\linewidth]{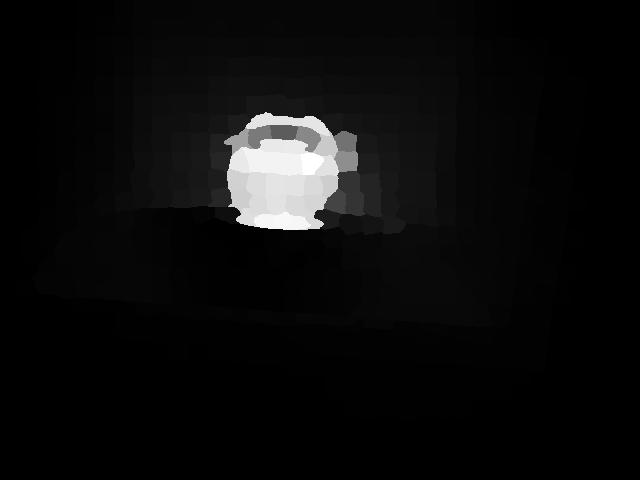}} \hspacefigure \\
\vspace{-1.5mm}
\subfloat[{RGB}]{ \includegraphics[width=\widthtwelve]{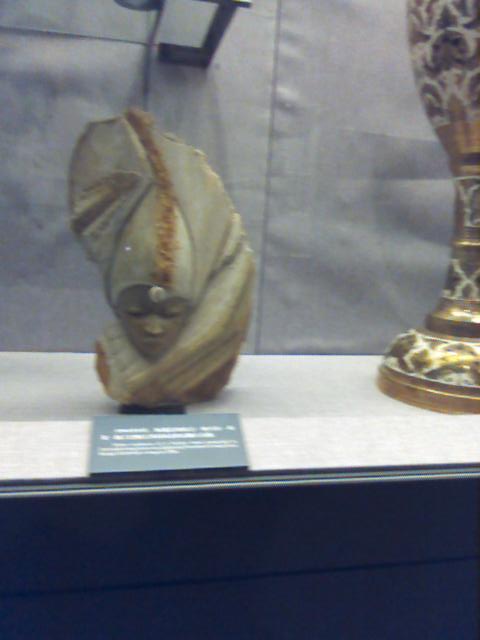}} \hspacefigure
\subfloat[{Depth}]{ \includegraphics[width=\widthtwelve]{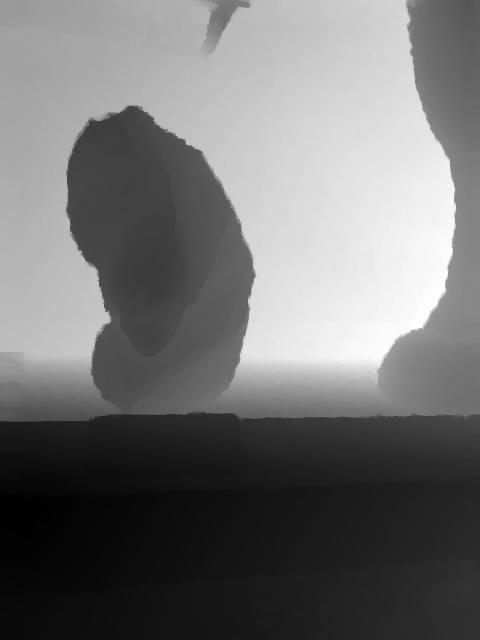}}\hspacefigure
\subfloat[{GT}]{ \includegraphics[width=\widthtwelve]{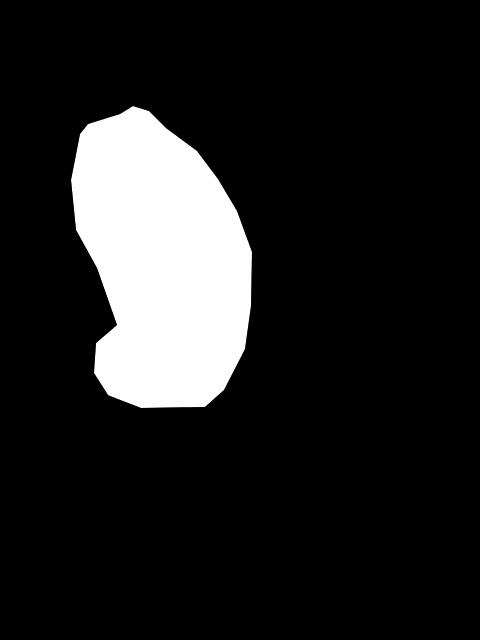}}\hspacefigure
\subfloat[{LMH}]{ \includegraphics[width=\widthtwelve]{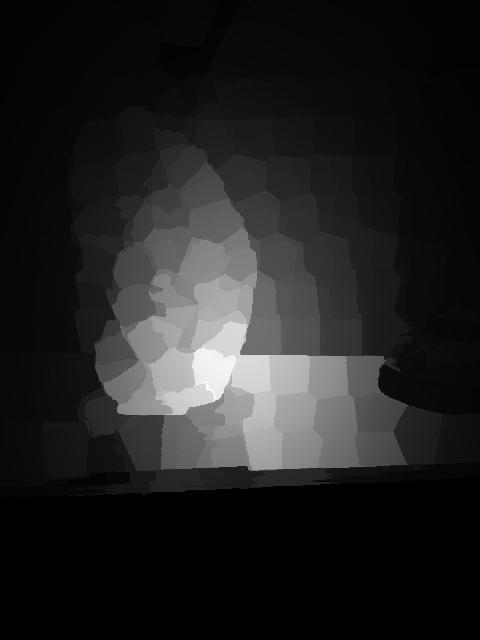}} \hspacefigure
\subfloat[{ACSD}]{ \includegraphics[width=\widthtwelve]{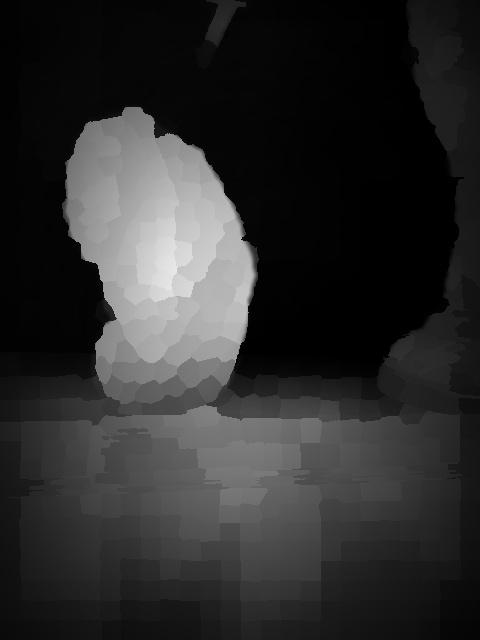}} \hspacefigure
\subfloat[{GP}]{ \includegraphics[width=\widthtwelve]{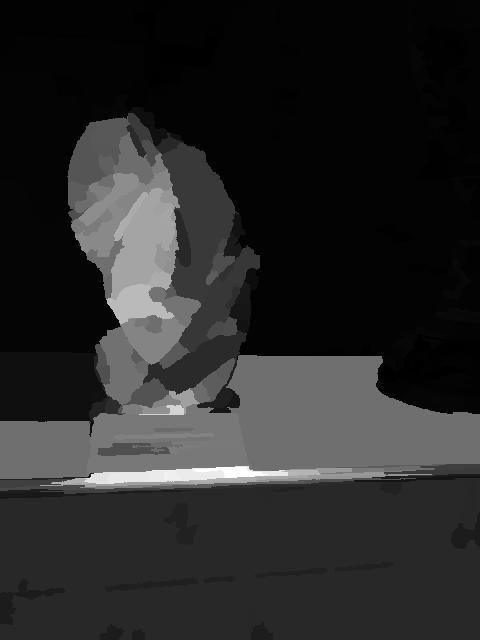}} \hspacefigure
\subfloat[{MCA}]{ \includegraphics[width=\widthtwelve]{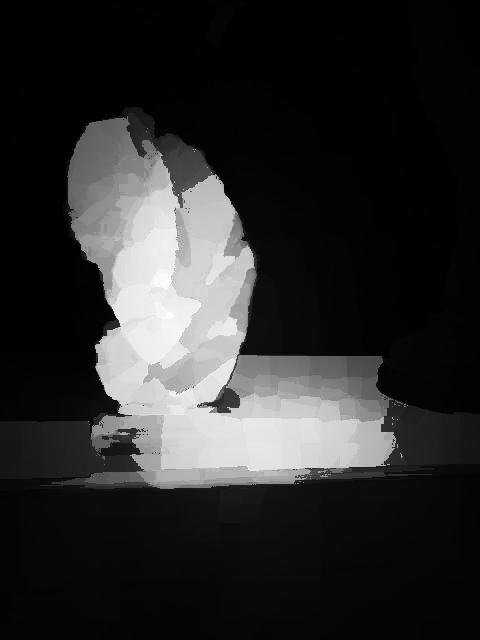}} \hspacefigure
\subfloat[{CNN\_F Init}]{ \includegraphics[width=\widthtwelve]{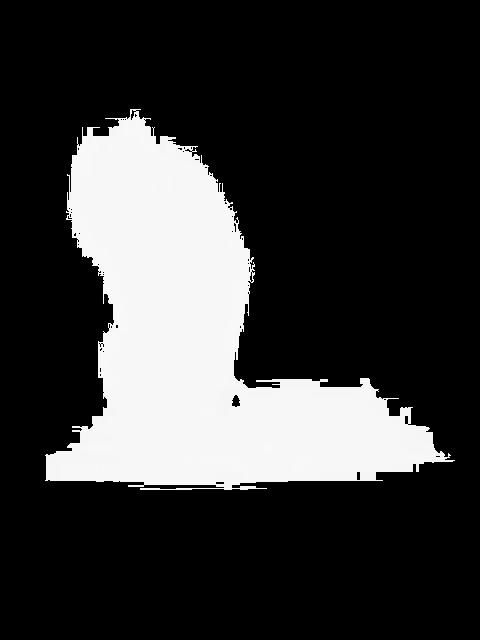}} \hspacefigure
\subfloat[{CNN\_F+LP}]{ \includegraphics[width=\widthtwelve]{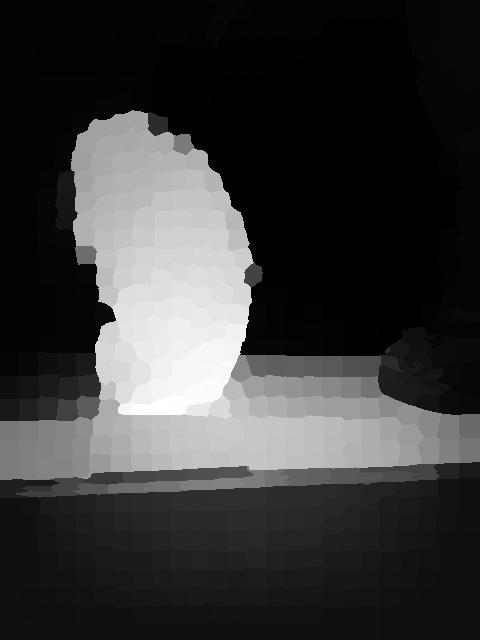}} \hspacefigure
\subfloat[{{Our Init}}]{ \includegraphics[width=\widthtwelve]{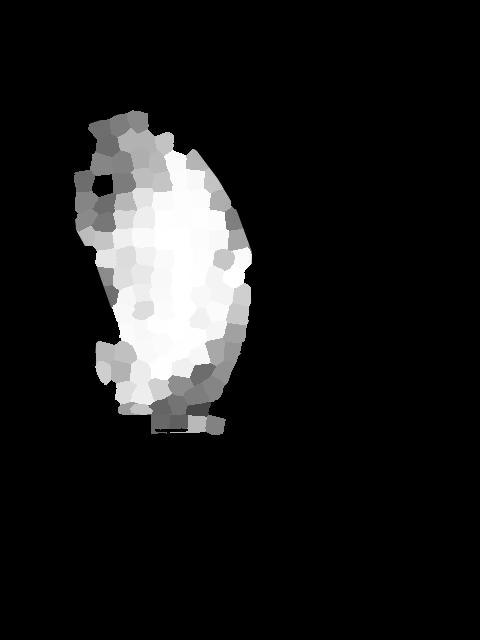}} \hspacefigure
\subfloat[{{Ours+LP}}]{ \includegraphics[width=\widthtwelve]{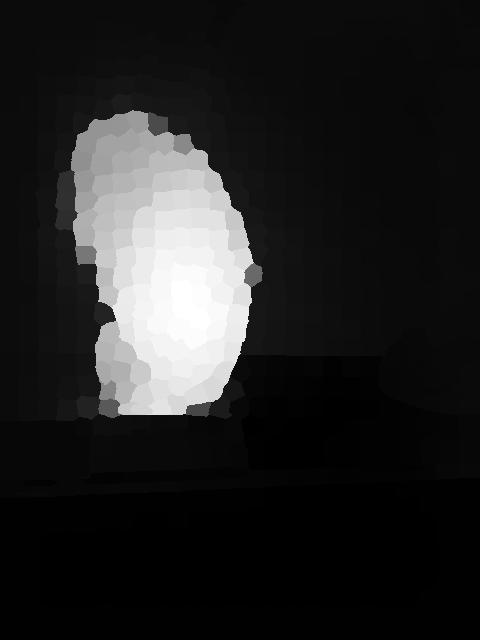}} \hspacefigure \\
\caption{ More examples to show the problem of saliency map merging methods. MCA and CNN\_F are the results of sophisticated fusion (fusing LMH \cite{peng2014rgbd}, ACSD \cite{ju2014depth}, and GP \cite{ren2015exploiting}). ``CNN\_F init'' and ``ours init'' are the initial results of CNN\_F and proposed hyper-feature without Lapalacian propagation respectively.}
\label{fig:sal_map}
\end{figure*}

The relevant comparison results of our proposed methods with these saliency map merging methods are shown in Figure \ref{fig:sal_map},
Table \ref{table:anlysis_NLPR}, Table \ref{table:anlysis}, and Table \ref{table:anlysis_LFSD}.  ``Fundamental fusion'' represents the results of four merging methods performed on seven fundamental saliency maps. ``Heuristic fusion'' gives the results of two state-of-the-art heuristic saliency map merging methods \cite{peng2014rgbd,ren2015exploiting}, while ``Sophisticated fusion'' gives the results of four merging methods performed on three sophisticated saliency maps (calculated from the existing state-of-the-art RGBD saliency detection methods LMH
\cite{peng2014rgbd}, ACSD \cite{ju2014depth}, and GP \cite{ren2015exploiting}).

For ``Fundamental fusion'' in Table \ref{table:anlysis_LFSD}, all the existing saliency map merging methods (including deep learning framework) cannot achieve satisfactory  performance.
Even though feeding with the state-of-the-art sophisticated saliency maps, these saliency merging methods still perform worse than our saliency feature fusion without LP
framework (0.8071 vs 0.8157), which further validates the flexibility of our feature level fusion. Note that 0.8157 are obtained from our initial saliency feature fusion network, which performs only on the pixel level and without considering spatial consistency. Our model achieves superior performance even though the input features are very simple (similar to the features used in ``Fundamental fusion''). Compared to those methods using similar features (in ``Fundamental fusion''), we can observe that fusing features is much more flexible than fusing saliency map.

\begin{figure*}
\centering
\captionsetup[subfigure]{labelformat=empty}
\subfloat{ \includegraphics[width=\widthNne,height=0.0859\linewidth]{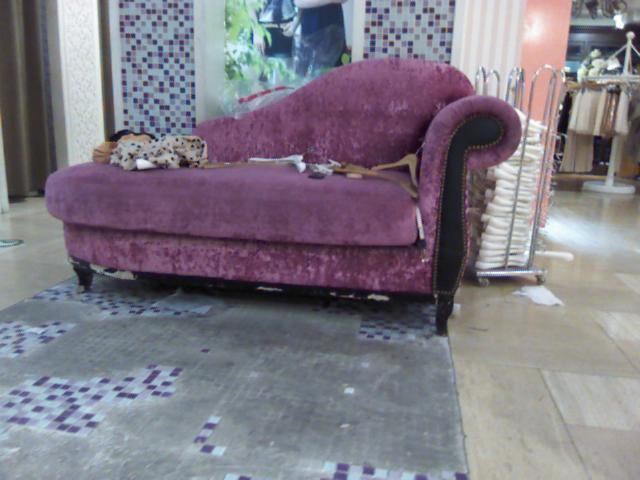}} \hspacefigure
\subfloat{ \includegraphics[width=\widthNne,height=0.0859\linewidth]{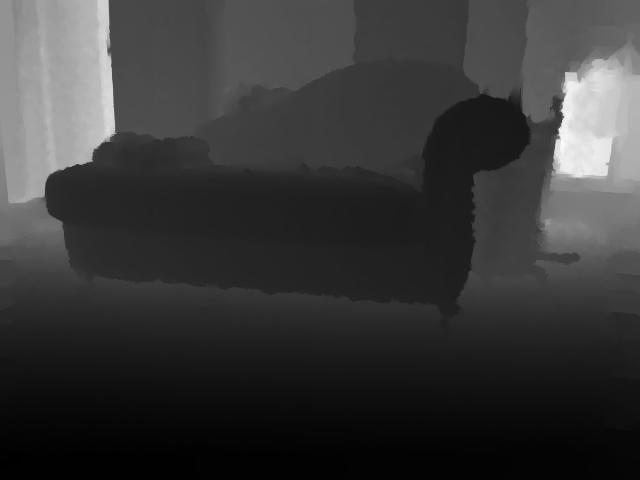}}\hspacefigure
\subfloat{ \includegraphics[width=\widthNne,height=0.0859\linewidth]{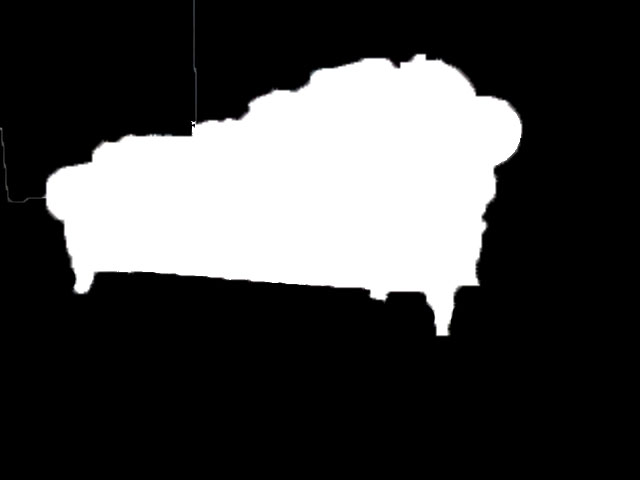}}\hspacefigure
\subfloat{ \includegraphics[width=\widthNne,height=0.0859\linewidth]{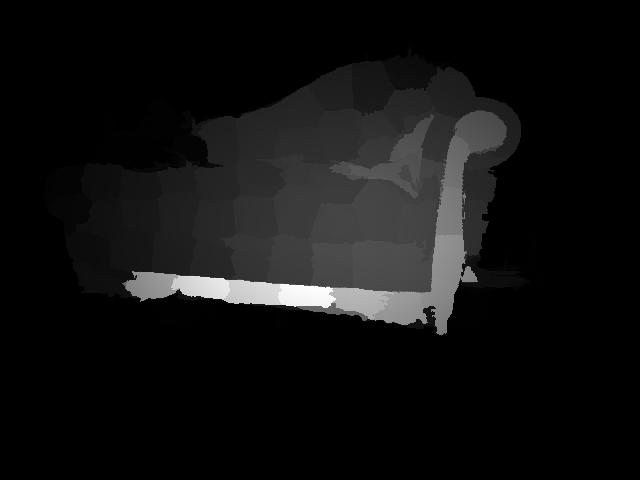}} \hspacefigure
\subfloat{ \includegraphics[width=\widthNne,height=0.0859\linewidth]{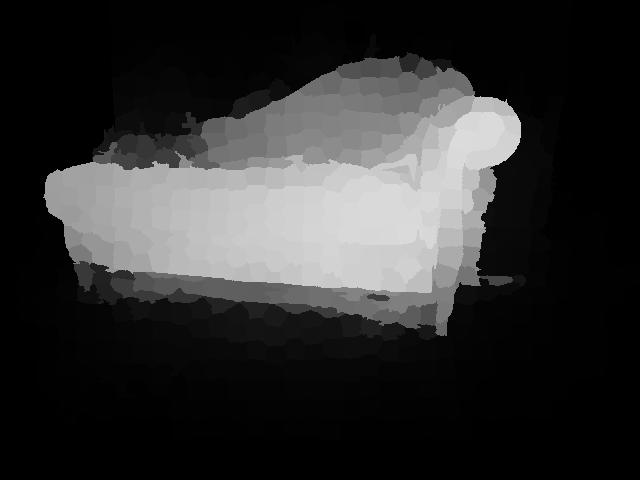}} \hspacefigure
\subfloat{ \includegraphics[width=\widthNne,height=0.0859\linewidth]{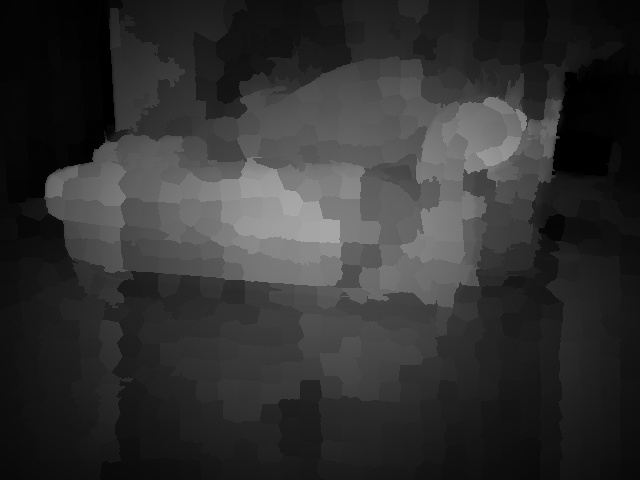}} \hspacefigure
\subfloat{ \includegraphics[width=\widthNne,height=0.0859\linewidth]{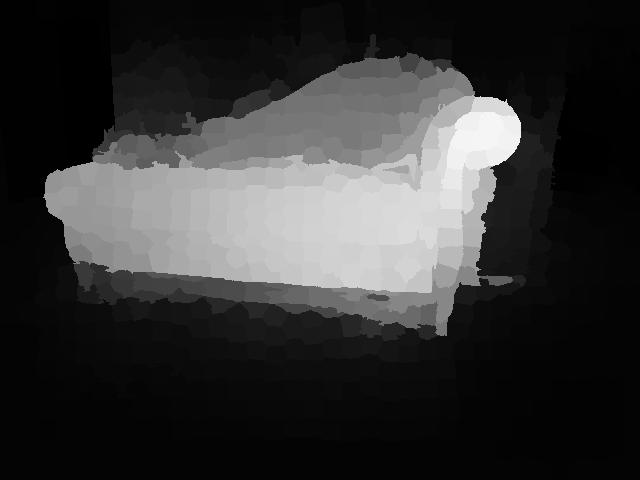}} \hspacefigure
\subfloat{ \includegraphics[width=\widthNne,height=0.0859\linewidth]{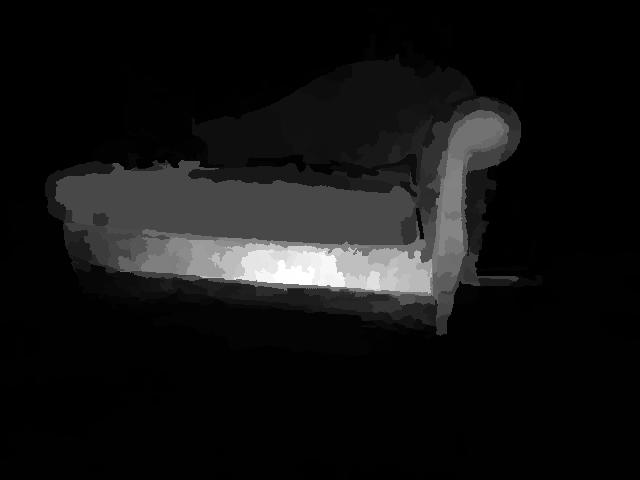}} \hspacefigure
\subfloat{ \includegraphics[width=\widthNne,height=0.0859\linewidth]{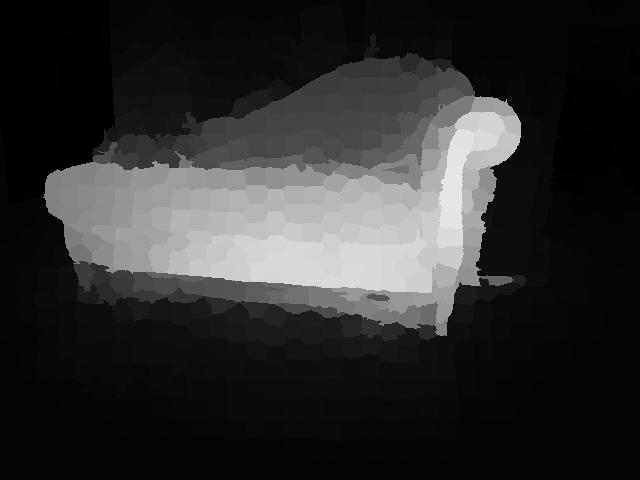}} \hspacefigure
\subfloat{ \includegraphics[width=\widthNne,height=0.0859\linewidth]{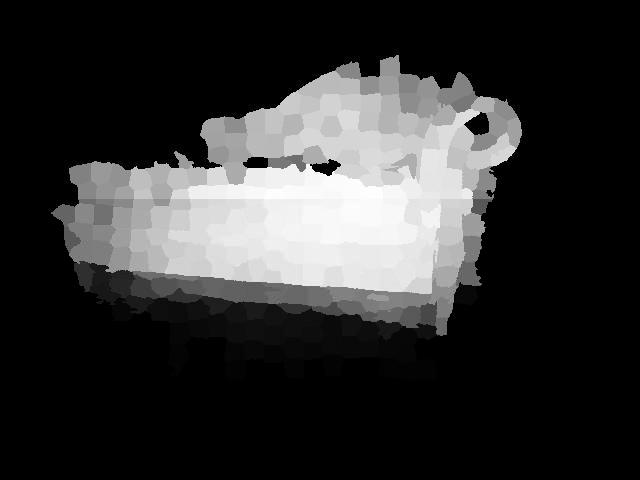}} \hspacefigure
\subfloat{ \includegraphics[width=\widthNne,height=0.0859\linewidth]{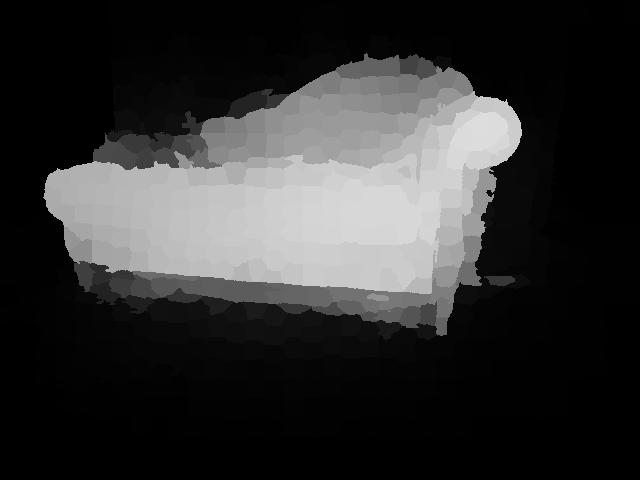}} \hspacefigure \\
\vspace{-1.5mm}
\subfloat{ \includegraphics[width=\widthNne]{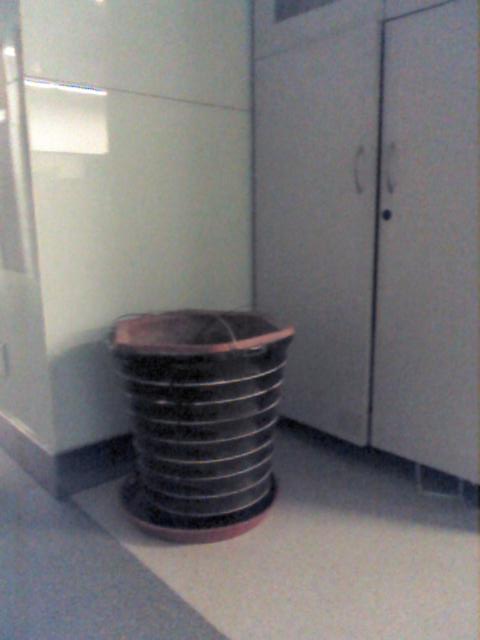}} \hspacefigure
\subfloat{ \includegraphics[width=\widthNne]{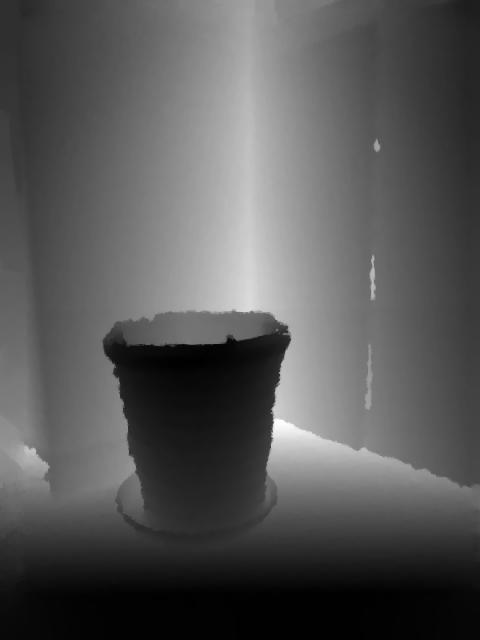}}\hspacefigure
\subfloat{ \includegraphics[width=\widthNne]{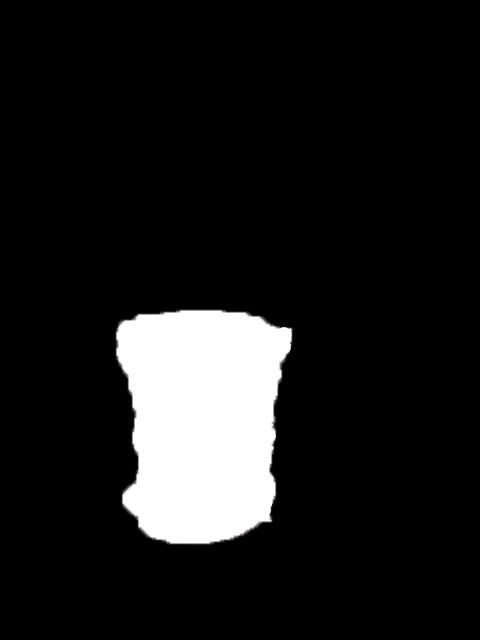}}\hspacefigure
\subfloat{ \includegraphics[width=\widthNne]{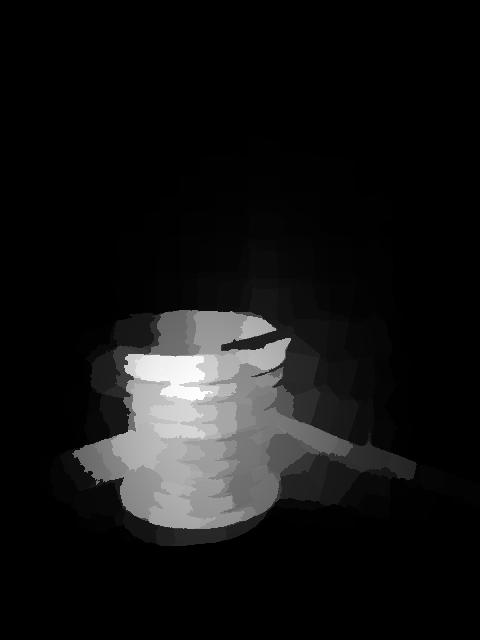}} \hspacefigure
\subfloat{ \includegraphics[width=\widthNne]{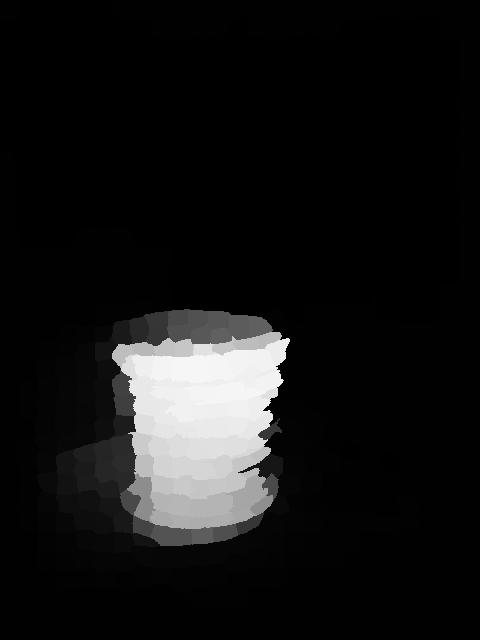}} \hspacefigure
\subfloat{ \includegraphics[width=\widthNne]{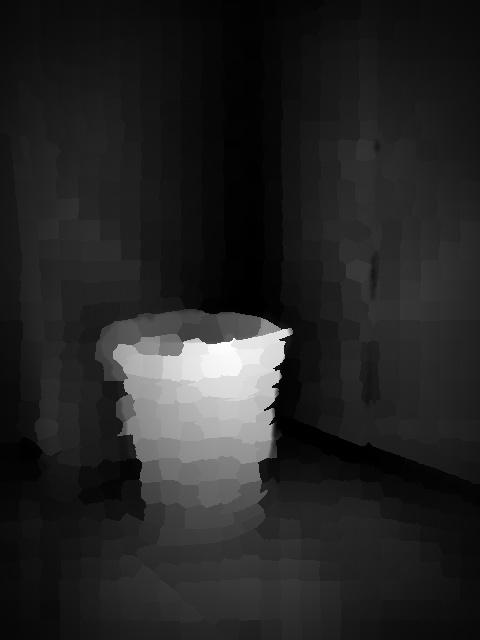}} \hspacefigure
\subfloat{ \includegraphics[width=\widthNne]{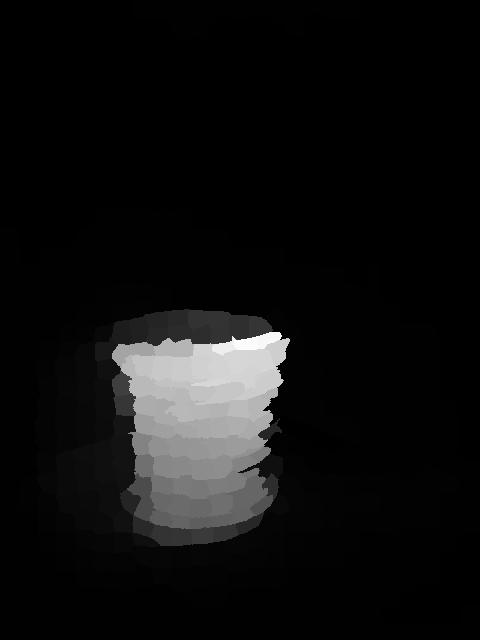}} \hspacefigure
\subfloat{ \includegraphics[width=\widthNne]{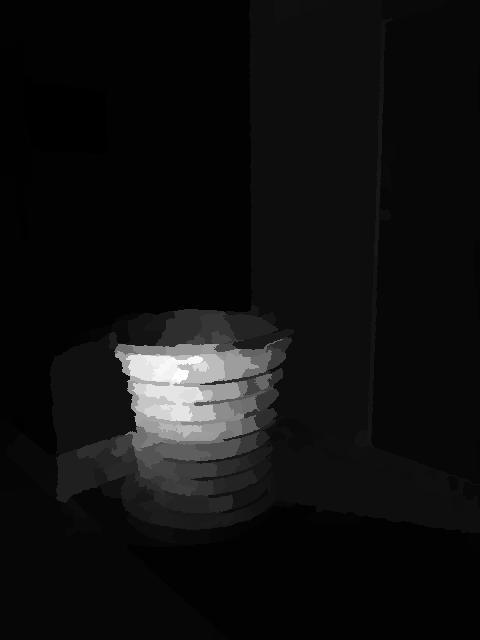}} \hspacefigure
\subfloat{ \includegraphics[width=\widthNne]{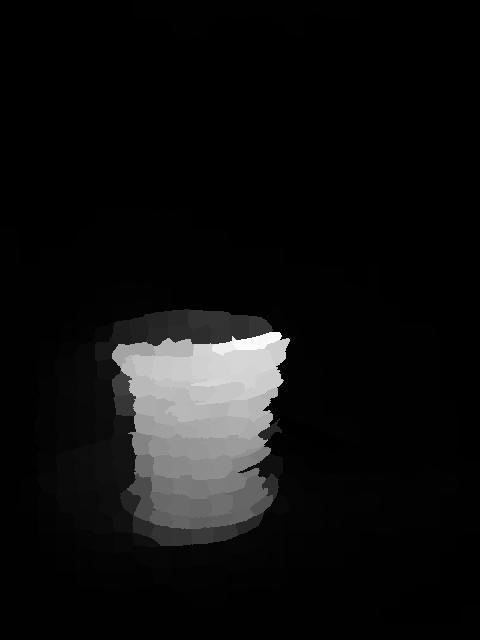}} \hspacefigure
\subfloat{ \includegraphics[width=\widthNne]{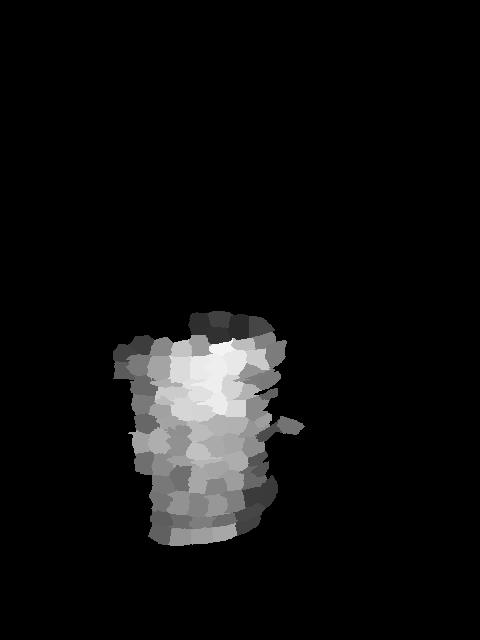}} \hspacefigure
\subfloat{ \includegraphics[width=\widthNne]{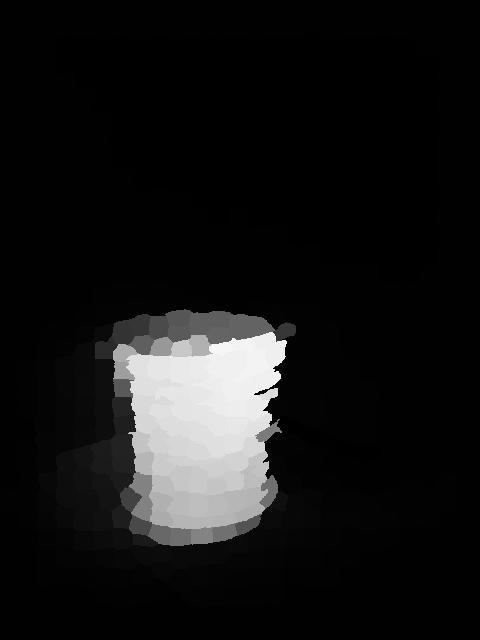}} \hspacefigure \\
\vspace{-1.5mm}
\subfloat{ \includegraphics[width=\widthNne]{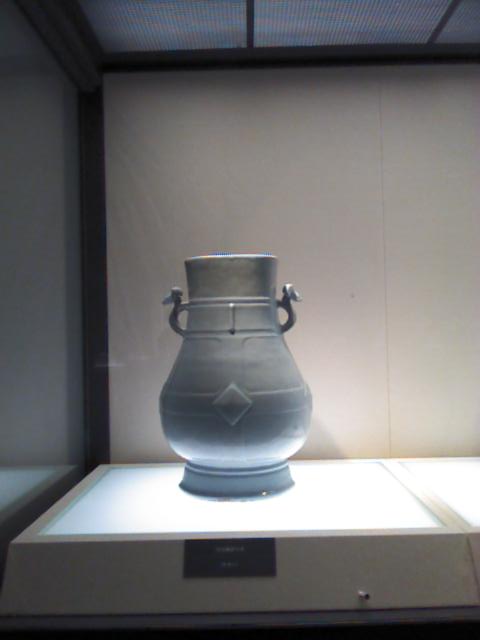}} \hspacefigure
\subfloat{ \includegraphics[width=\widthNne]{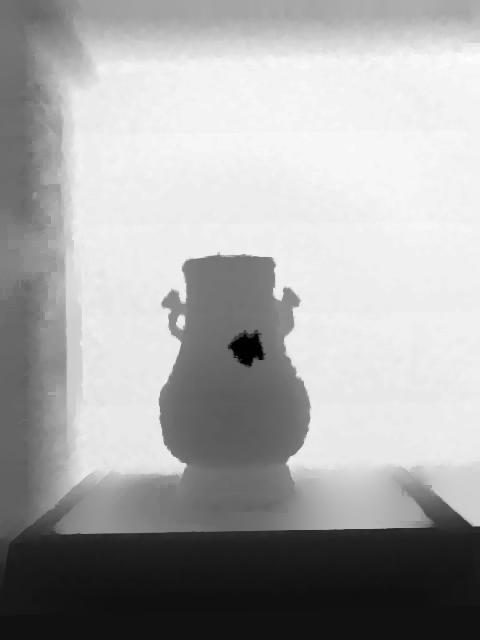}}\hspacefigure
\subfloat{ \includegraphics[width=\widthNne]{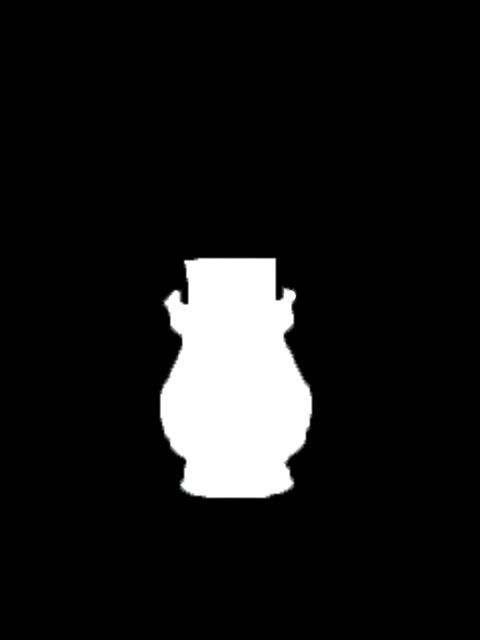}}\hspacefigure
\subfloat{ \includegraphics[width=\widthNne]{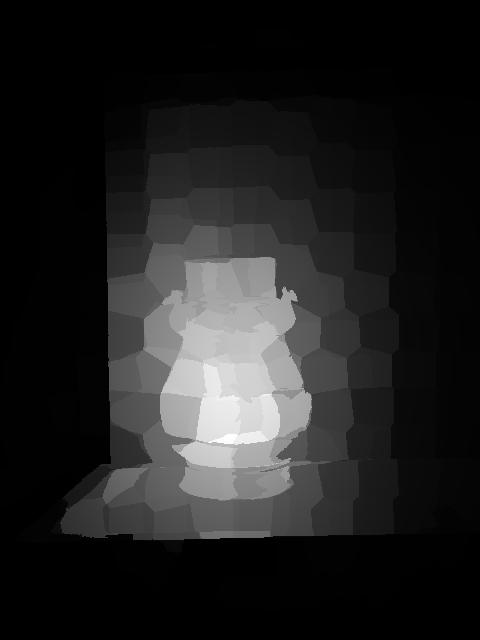}} \hspacefigure
\subfloat{ \includegraphics[width=\widthNne]{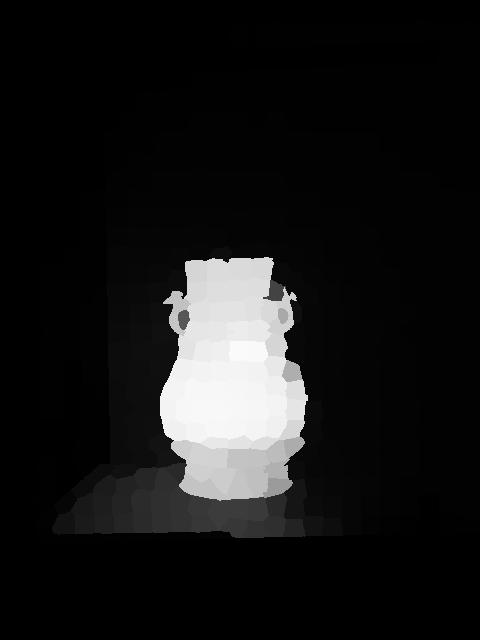}} \hspacefigure
\subfloat{ \includegraphics[width=\widthNne]{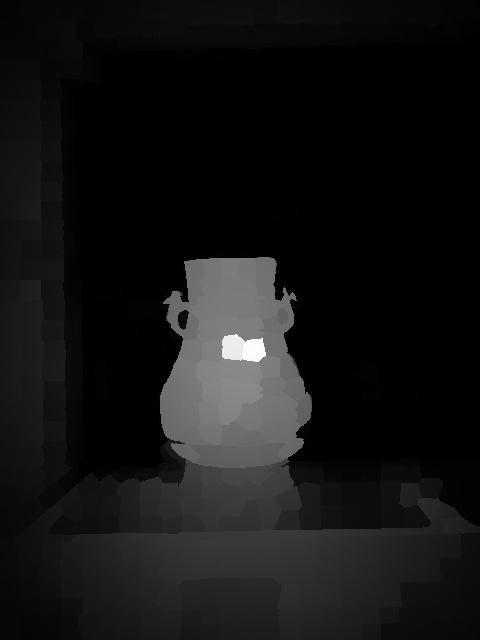}} \hspacefigure
\subfloat{ \includegraphics[width=\widthNne]{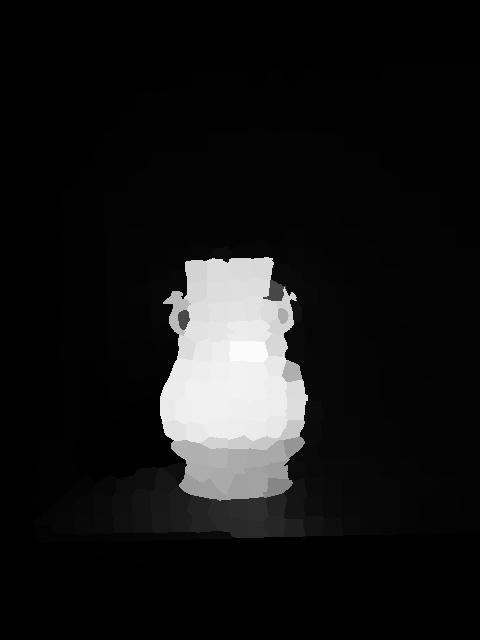}} \hspacefigure
\subfloat{ \includegraphics[width=\widthNne]{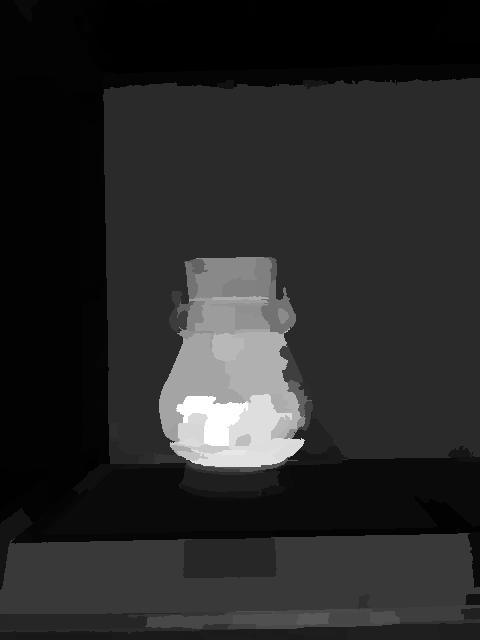}} \hspacefigure
\subfloat{ \includegraphics[width=\widthNne]{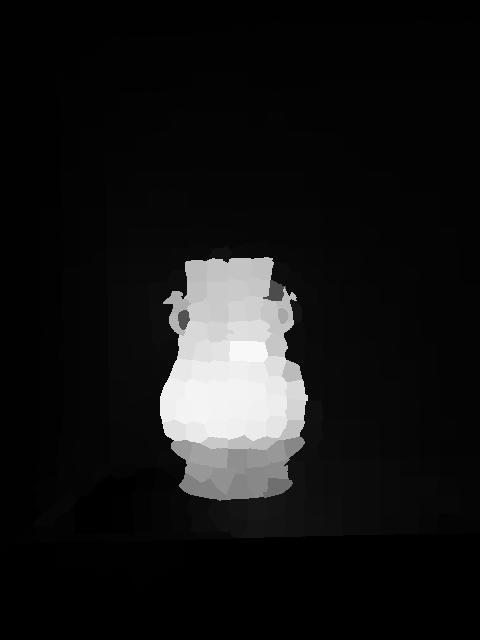}} \hspacefigure
\subfloat{ \includegraphics[width=\widthNne]{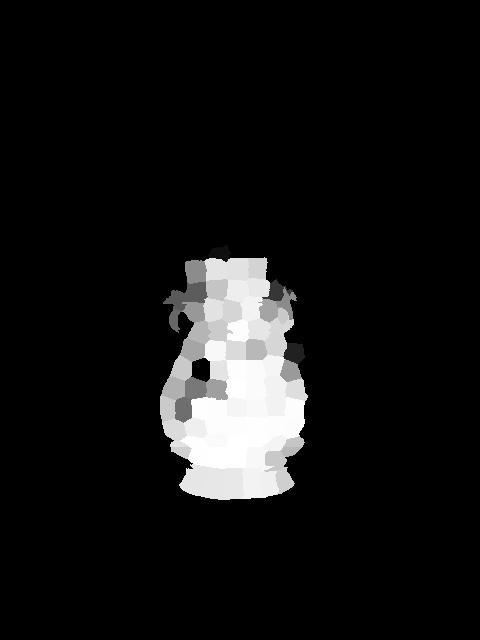}} \hspacefigure
\subfloat{ \includegraphics[width=\widthNne]{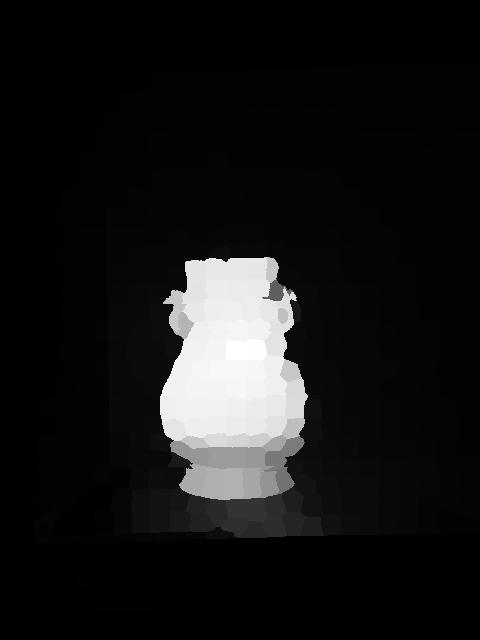}} \hspacefigure \\
\vspace{-1.5mm}
\subfloat{ \includegraphics[width=\widthNne]{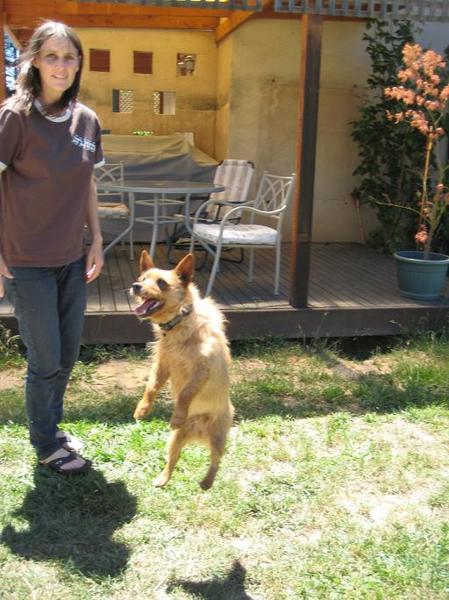}} \hspacefigure
\subfloat{ \includegraphics[width=\widthNne]{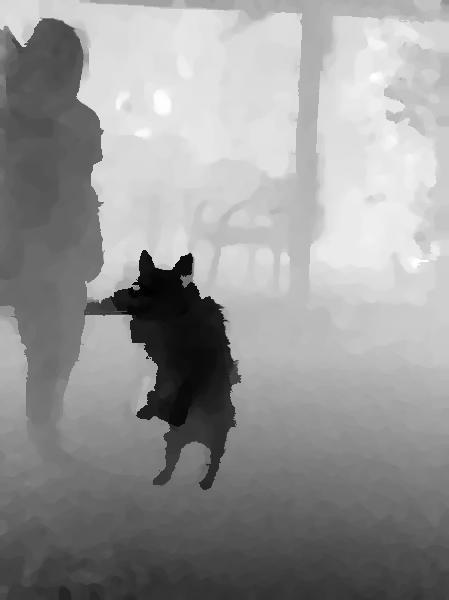}}\hspacefigure
\subfloat{ \includegraphics[width=\widthNne]{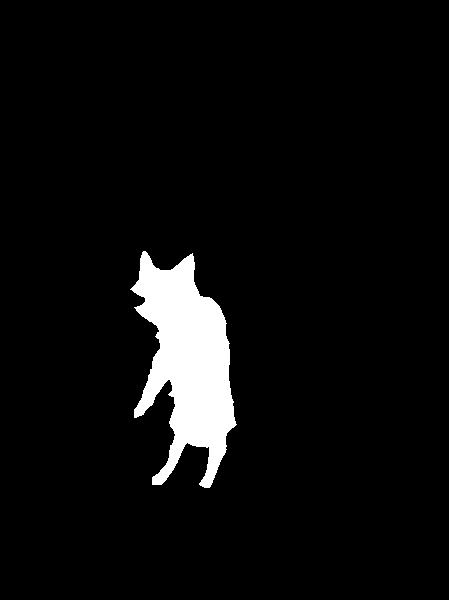}}\hspacefigure
\subfloat{ \includegraphics[width=\widthNne]{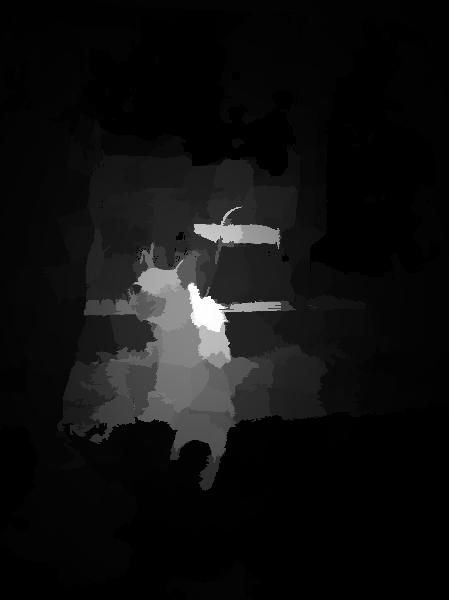}} \hspacefigure
\subfloat{ \includegraphics[width=\widthNne]{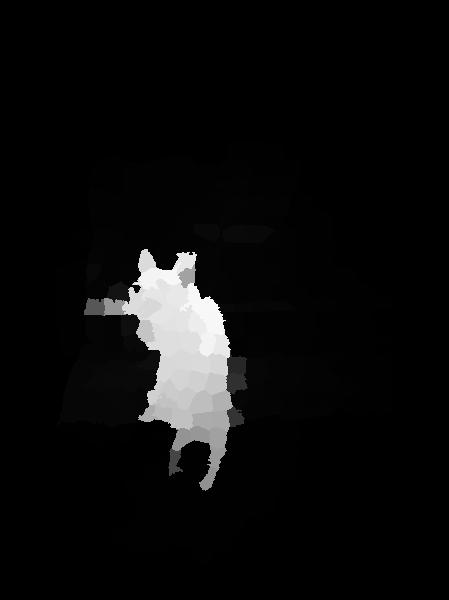}} \hspacefigure
\subfloat{ \includegraphics[width=\widthNne]{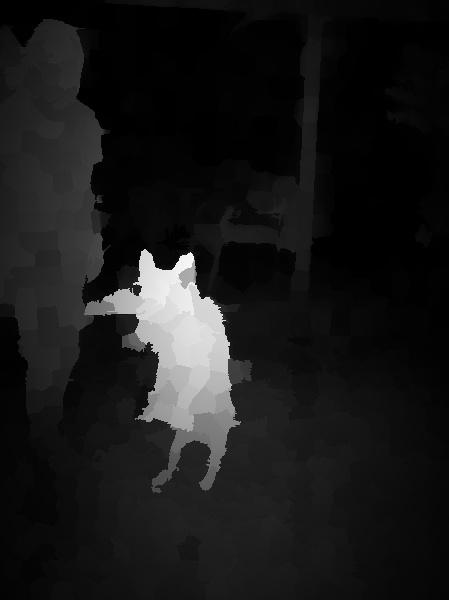}} \hspacefigure
\subfloat{ \includegraphics[width=\widthNne]{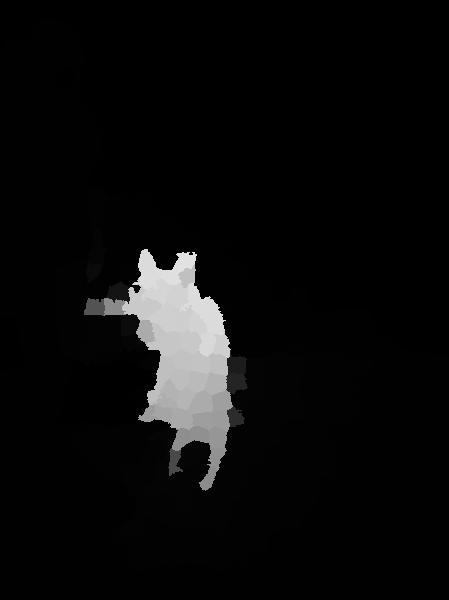}} \hspacefigure
\subfloat{ \includegraphics[width=\widthNne]{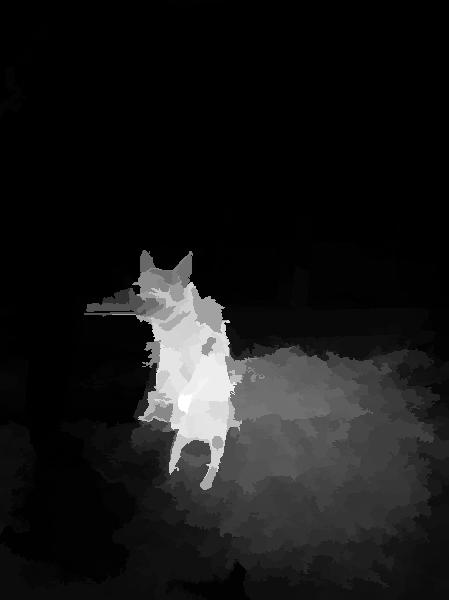}} \hspacefigure
\subfloat{ \includegraphics[width=\widthNne]{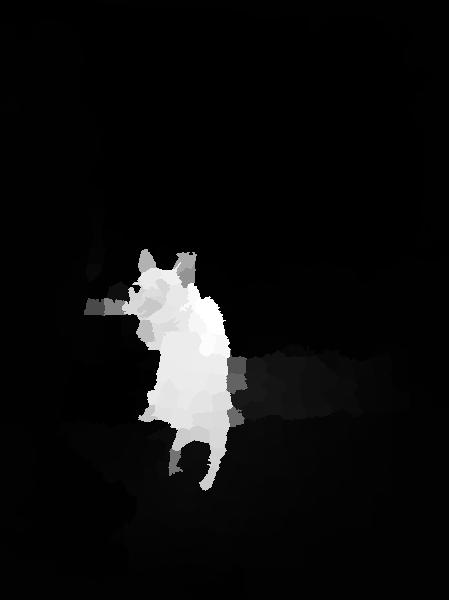}} \hspacefigure
\subfloat{ \includegraphics[width=\widthNne]{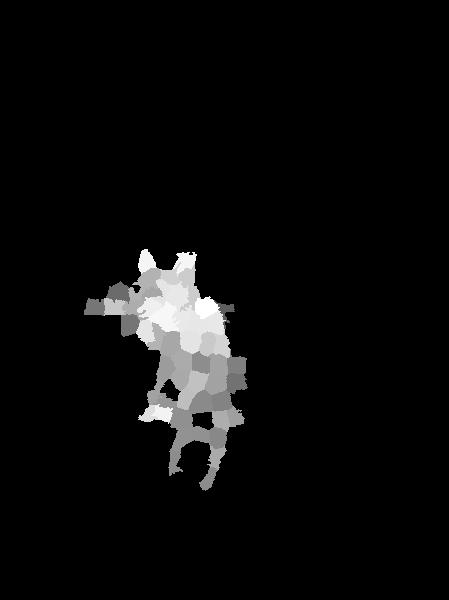}} \hspacefigure
\subfloat{ \includegraphics[width=\widthNne]{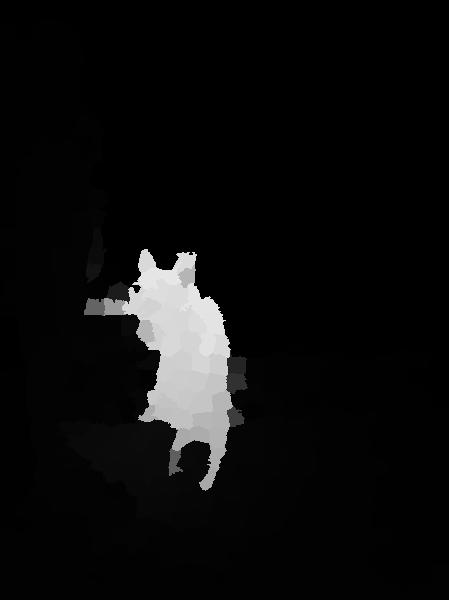}} \hspacefigure\\
\vspace{-1.5mm}
\subfloat{ \includegraphics[width=\widthNne]{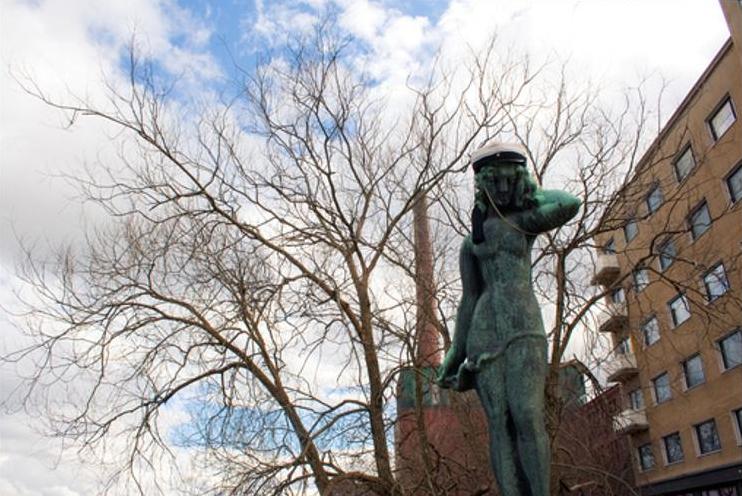}} \hspacefigure
\subfloat{ \includegraphics[width=\widthNne]{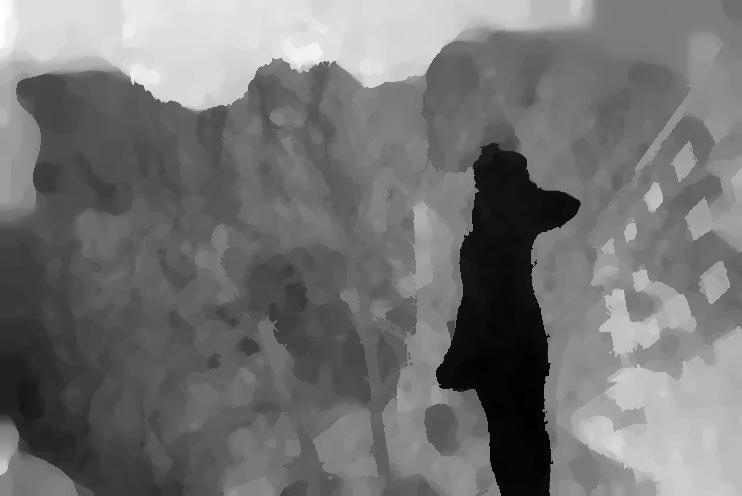}}\hspacefigure
\subfloat{ \includegraphics[width=\widthNne]{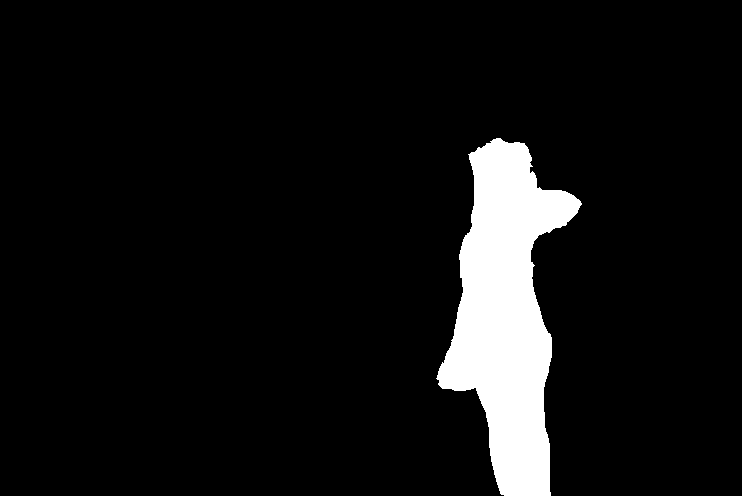}}\hspacefigure
\subfloat{ \includegraphics[width=\widthNne]{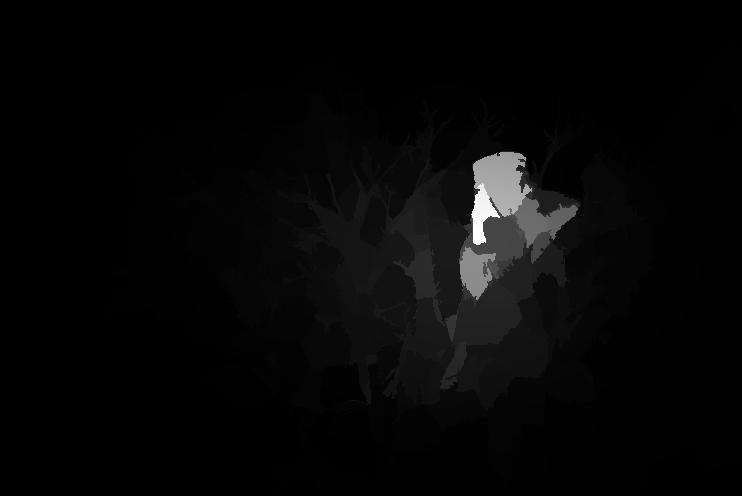}} \hspacefigure
\subfloat{ \includegraphics[width=\widthNne]{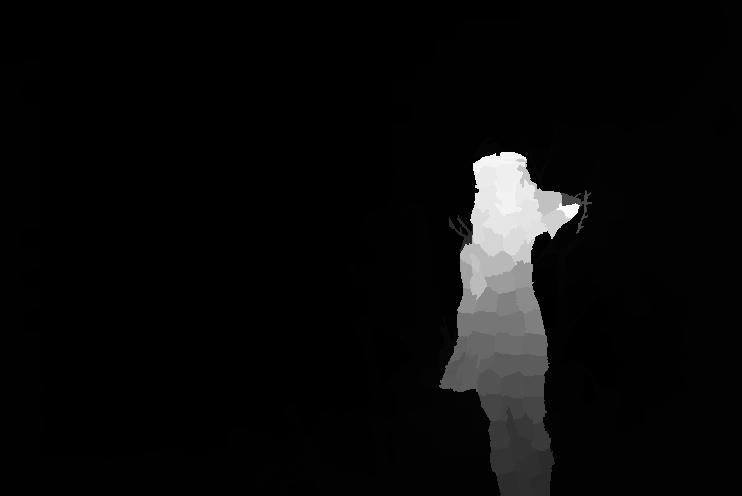}} \hspacefigure
\subfloat{ \includegraphics[width=\widthNne]{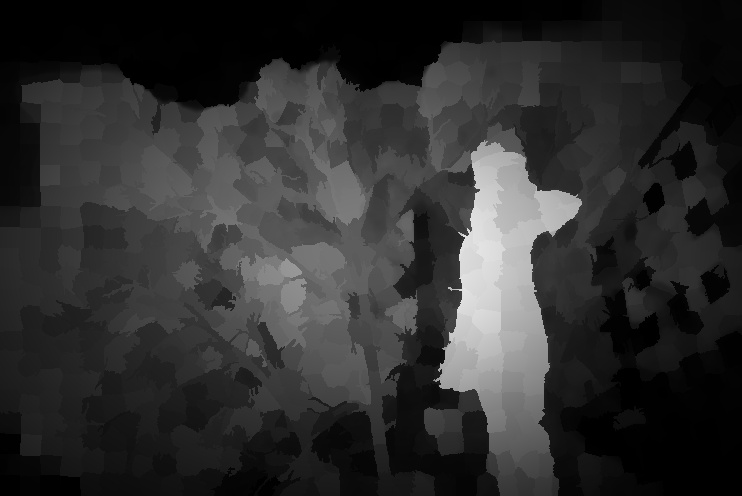}} \hspacefigure
\subfloat{ \includegraphics[width=\widthNne]{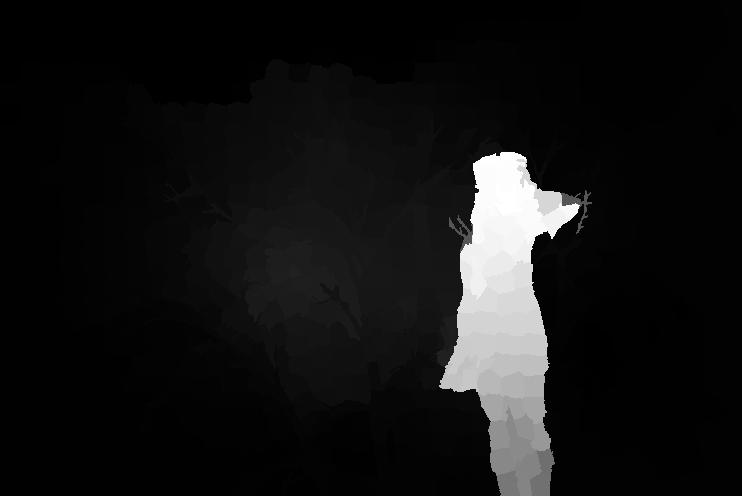}} \hspacefigure
\subfloat{ \includegraphics[width=\widthNne]{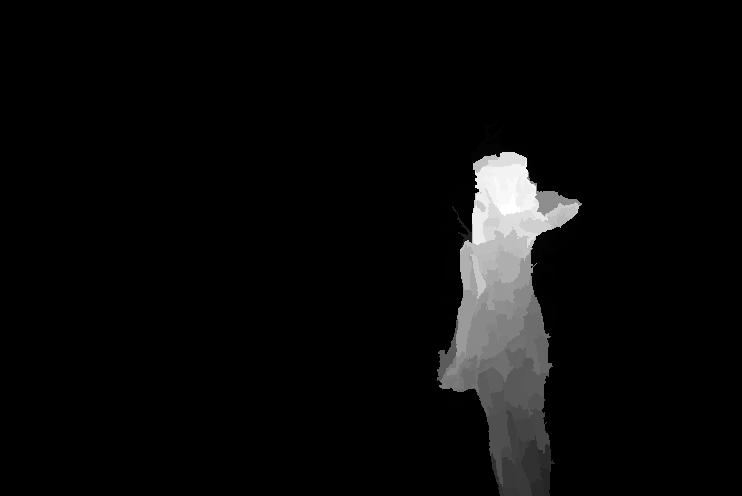}} \hspacefigure
\subfloat{ \includegraphics[width=\widthNne]{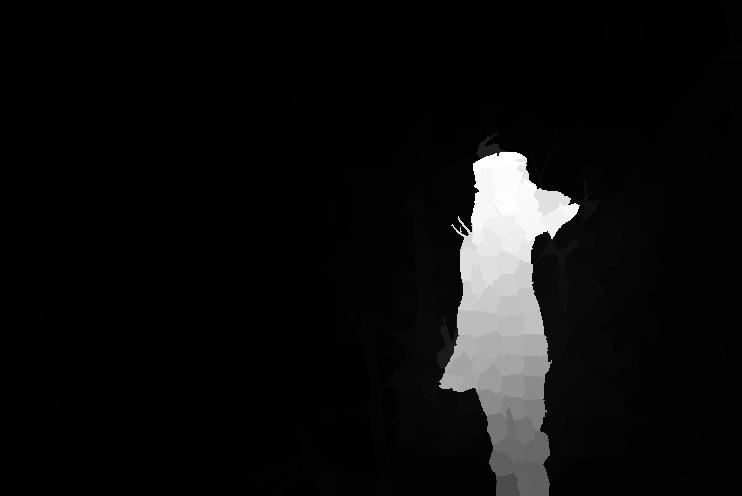}} \hspacefigure
\subfloat{ \includegraphics[width=\widthNne]{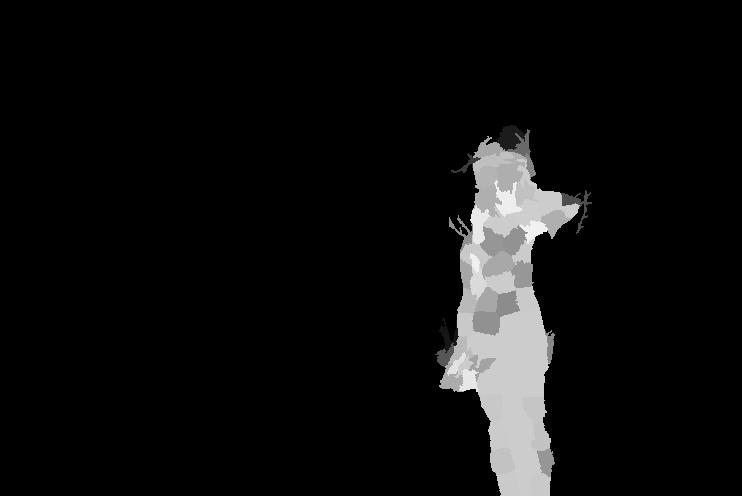}} \hspacefigure
\subfloat{ \includegraphics[width=\widthNne]{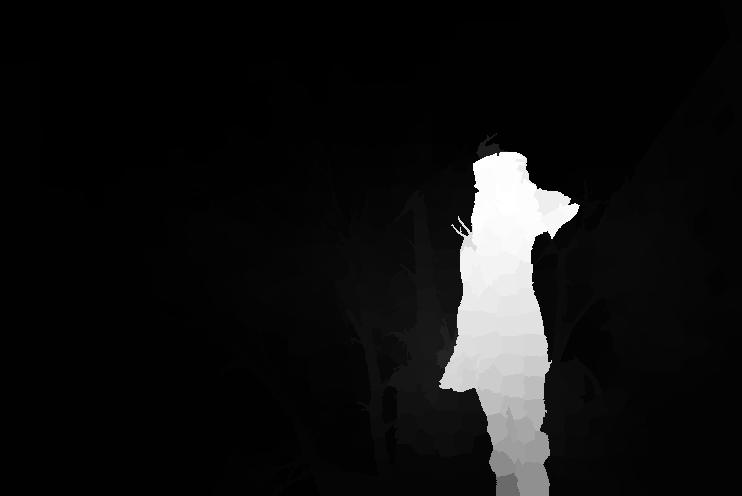}} \hspacefigure \\
\vspace{-1.5mm}
\subfloat{ \includegraphics[width=\widthNne,height=0.0859\linewidth]{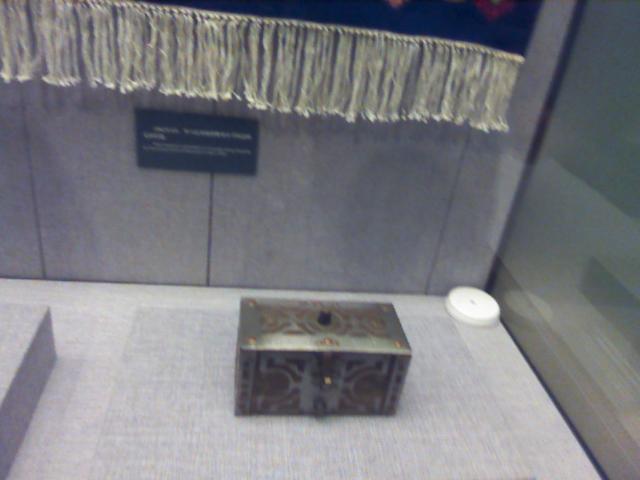}} \hspacefigure
\subfloat{ \includegraphics[width=\widthNne,height=0.0859\linewidth]{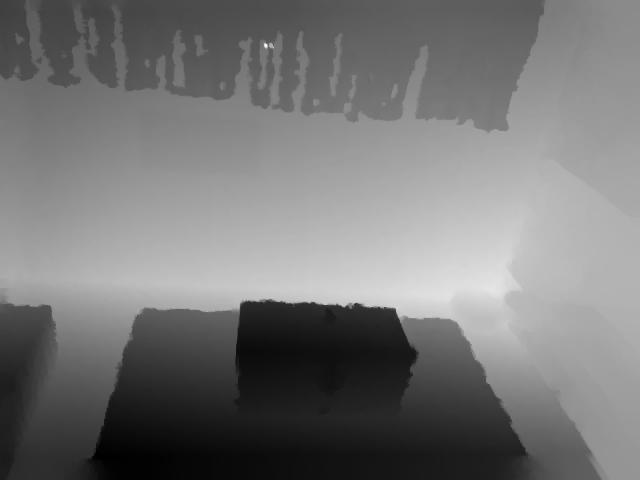}}\hspacefigure
\subfloat{ \includegraphics[width=\widthNne,height=0.0859\linewidth]{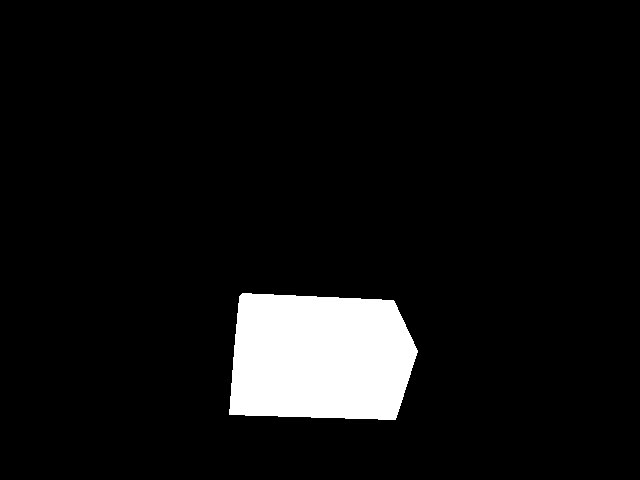}}\hspacefigure
\subfloat{ \includegraphics[width=\widthNne,height=0.0859\linewidth]{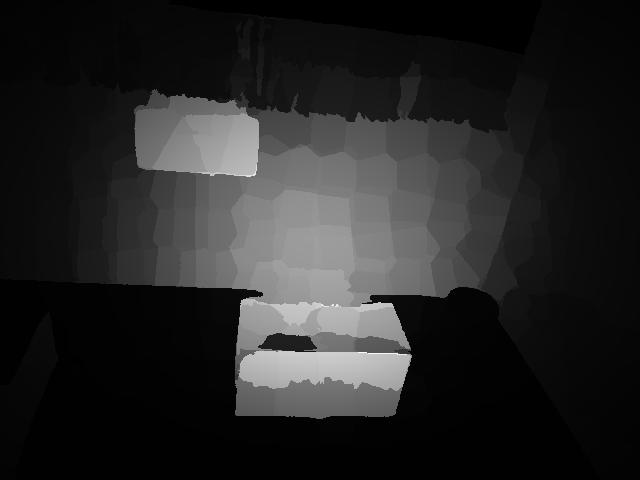}} \hspacefigure
\subfloat{ \includegraphics[width=\widthNne,height=0.0859\linewidth]{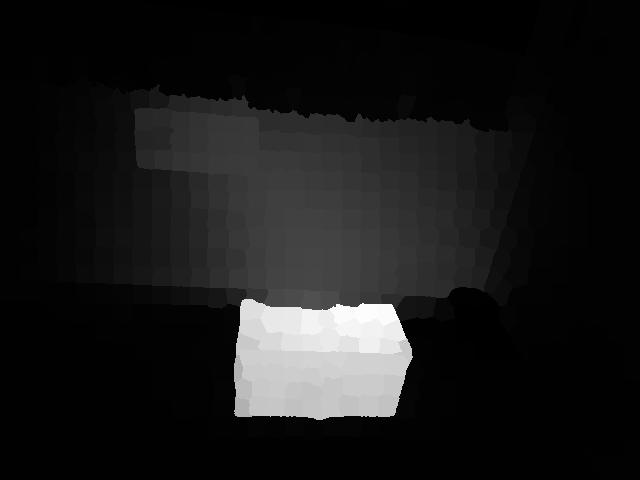}} \hspacefigure
\subfloat{ \includegraphics[width=\widthNne,height=0.0859\linewidth]{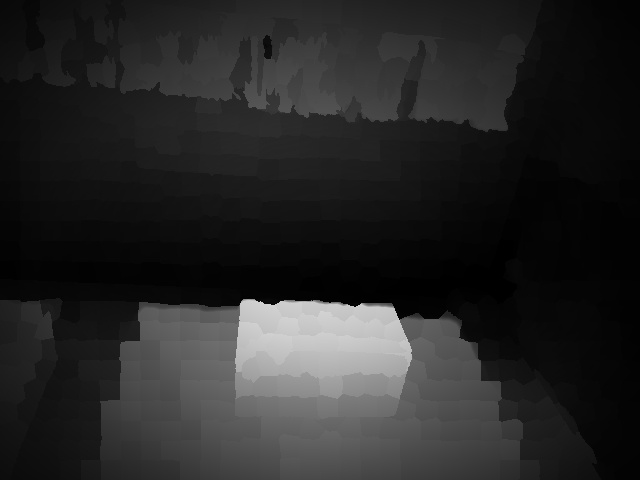}} \hspacefigure
\subfloat{ \includegraphics[width=\widthNne,height=0.0859\linewidth]{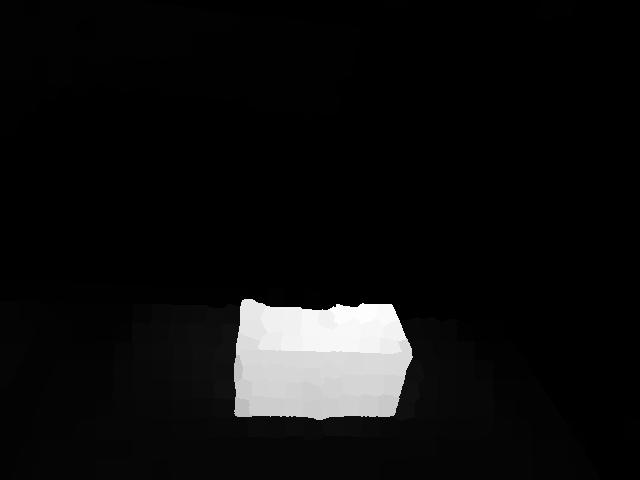}} \hspacefigure
\subfloat{ \includegraphics[width=\widthNne,height=0.0859\linewidth]{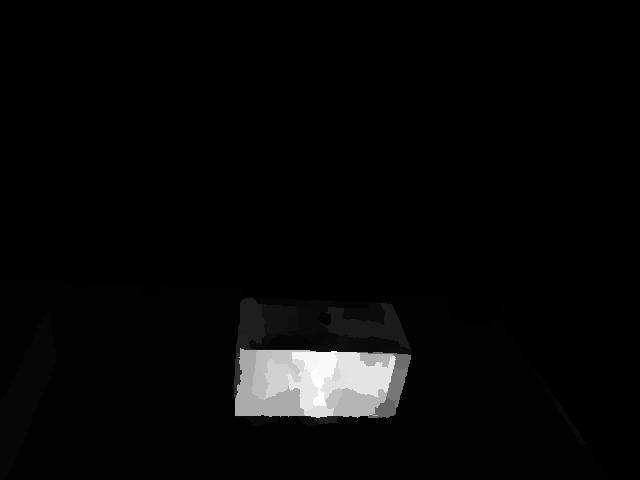}} \hspacefigure
\subfloat{ \includegraphics[width=\widthNne,height=0.0859\linewidth]{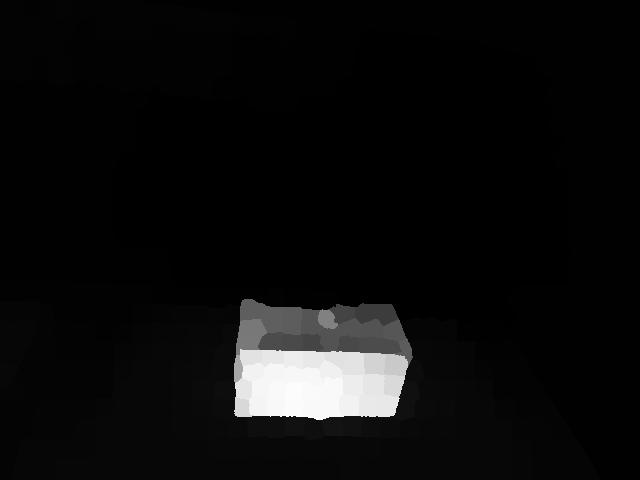}} \hspacefigure
\subfloat{ \includegraphics[width=\widthNne,height=0.0859\linewidth]{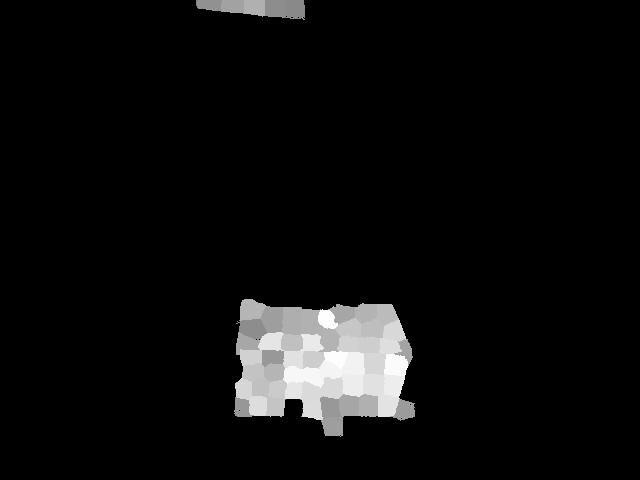}} \hspacefigure
\subfloat{ \includegraphics[width=\widthNne,height=0.0859\linewidth]{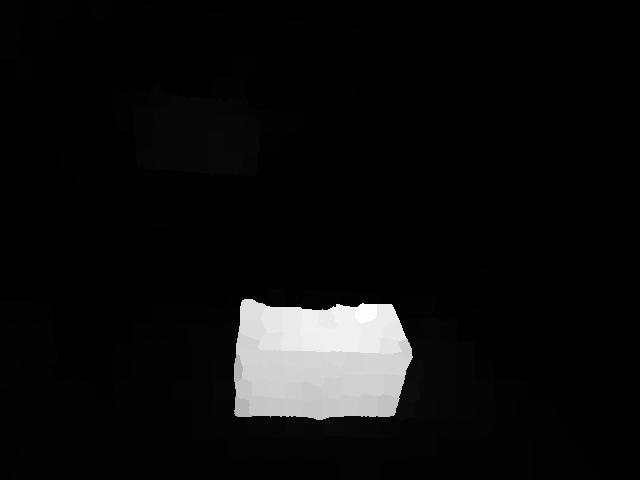}} \hspacefigure \\
\vspace{-1.5mm}
\subfloat[RGB]{ \includegraphics[width=\widthNne,height=0.0859\linewidth]{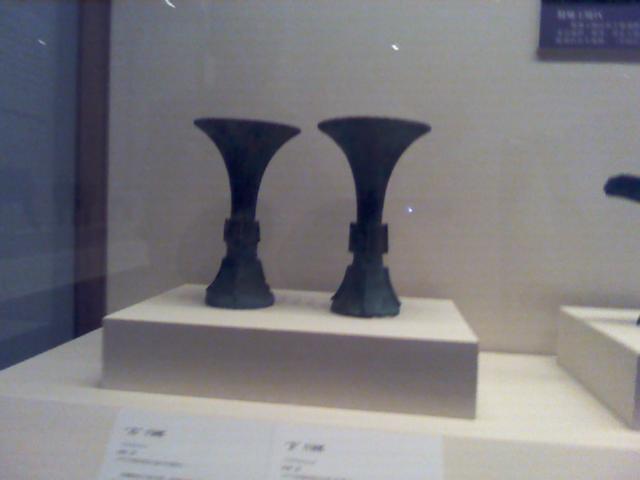}} \hspacefigure
\subfloat[Depth]{ \includegraphics[width=\widthNne,height=0.0859\linewidth]{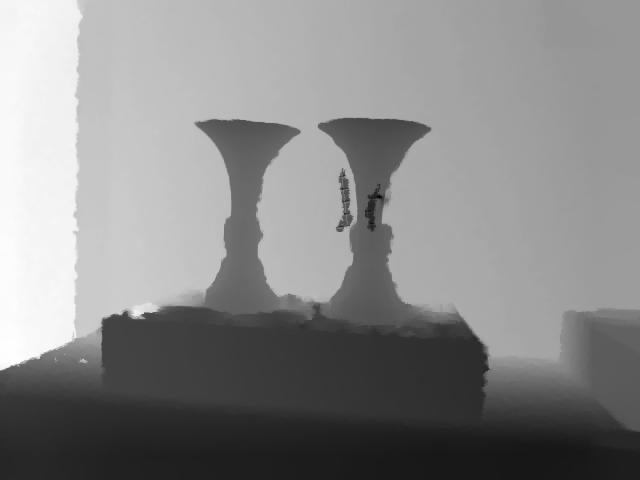}}\hspacefigure
\subfloat[GT]{ \includegraphics[width=\widthNne,height=0.0859\linewidth]{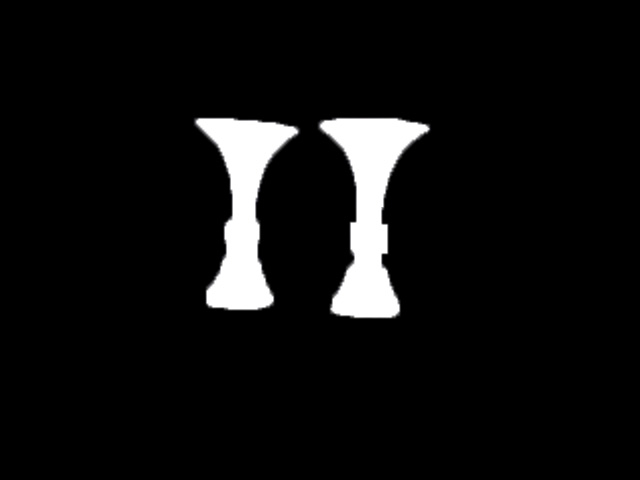}}\hspacefigure
\subfloat[LMH]{ \includegraphics[width=\widthNne,height=0.0859\linewidth]{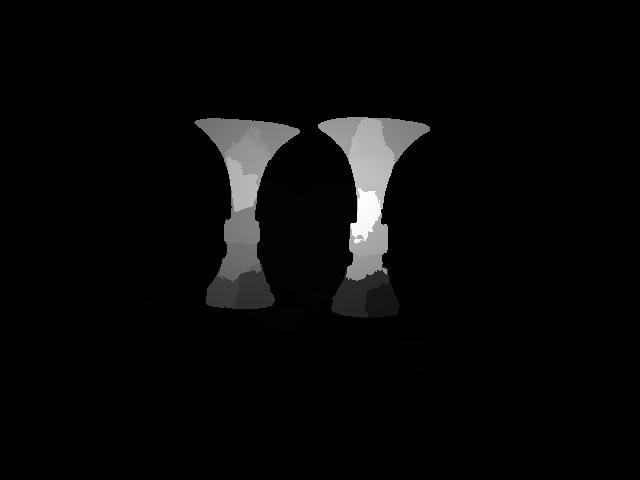}} \hspacefigure
\subfloat[LMH+LP]{ \includegraphics[width=\widthNne,height=0.0859\linewidth]{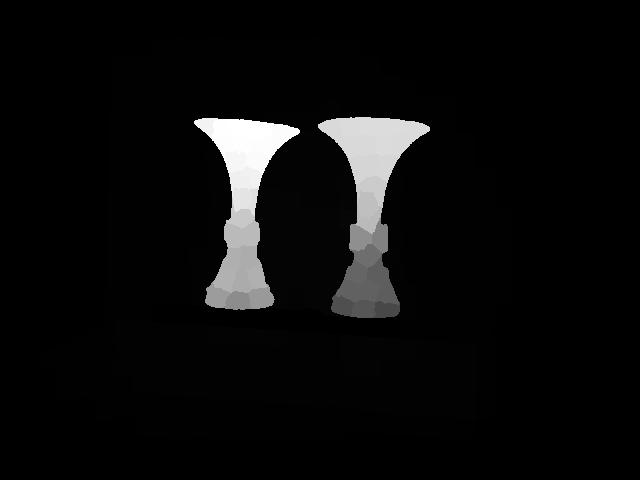}} \hspacefigure
\subfloat[ACSD]{ \includegraphics[width=\widthNne,height=0.0859\linewidth]{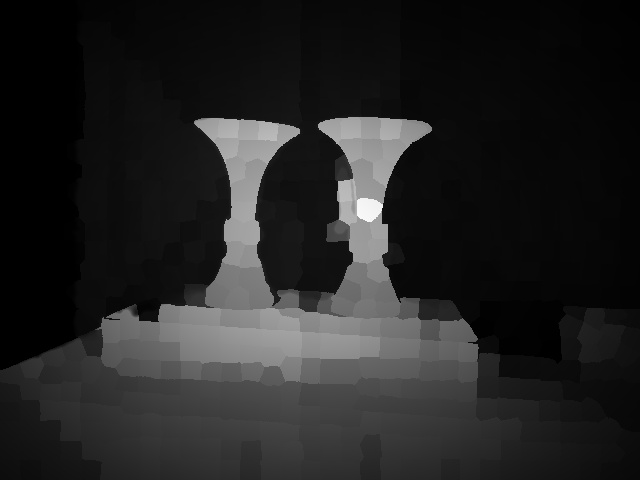}} \hspacefigure
\subfloat[ACSD+LP]{ \includegraphics[width=\widthNne,height=0.0859\linewidth]{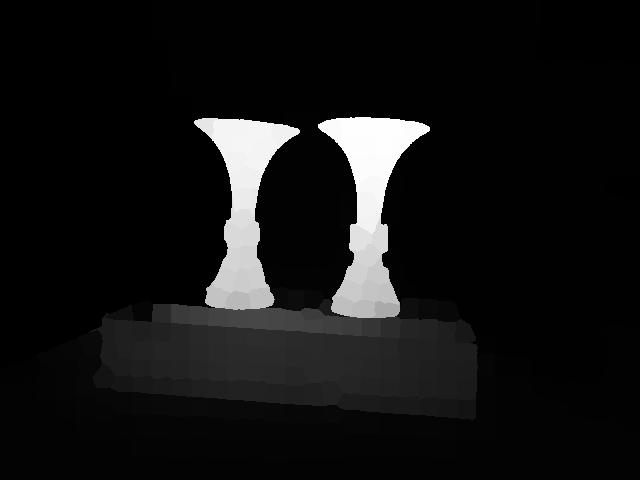}} \hspacefigure
\subfloat[GP]{ \includegraphics[width=\widthNne,height=0.0859\linewidth]{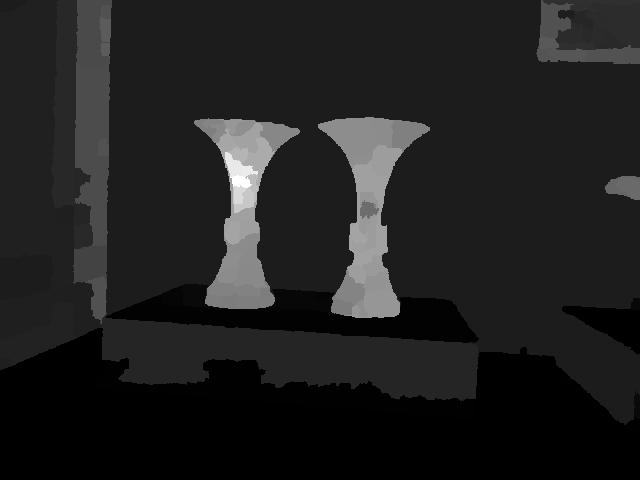}} \hspacefigure
\subfloat[GP+LP]{ \includegraphics[width=\widthNne,height=0.0859\linewidth]{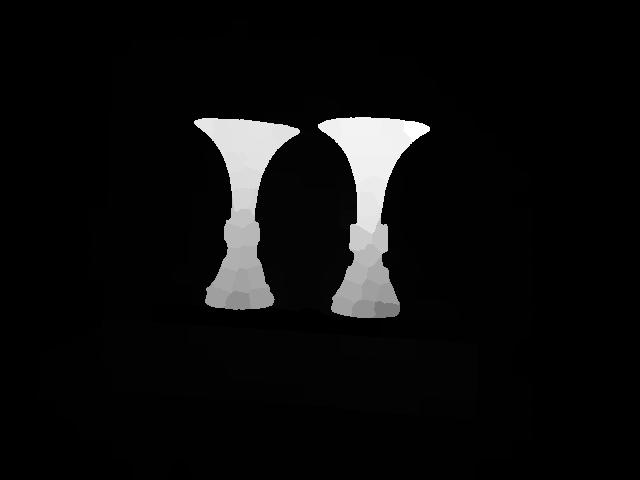}} \hspacefigure
\subfloat[Our init]{ \includegraphics[width=\widthNne,height=0.0859\linewidth]{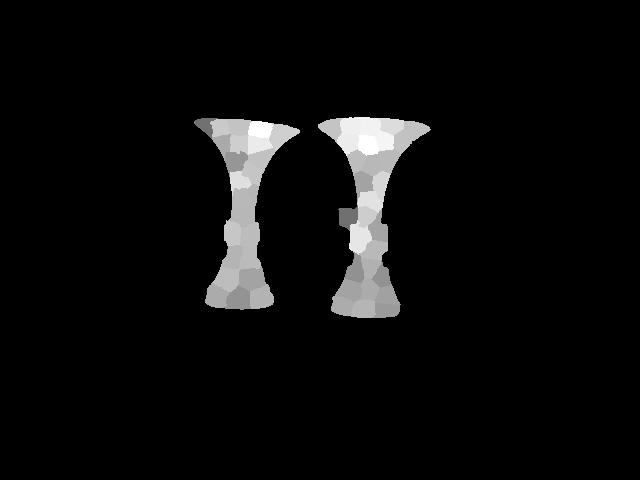}} \hspacefigure
\subfloat[Ours+LP]{ \includegraphics[width=\widthNne,height=0.0859\linewidth]{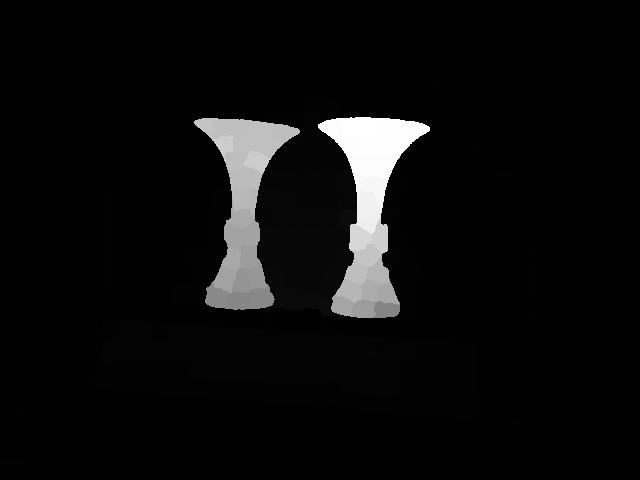}} \hspacefigure\\
\caption{ Examples to show the effectiveness of Laplacian propagation.}
\label{fig:sal_LP}
\end{figure*}

\noindent{\textbf{Analysis of Laplacian Propagation.}}
We then evaluate the effective of the proposed Laplacian propagation, and the optimized results of the existing methods using Laplacian propagation. The F-measure scores of our RGBD method without Laplacian propagation on three test dataset \cite{peng2014rgbd,ju2014depth,Li_2014_CVPR} are shown in blue in Table \ref{table:anlysis_NLPR}, Table \ref{table:anlysis}, and Table \ref{table:anlysis_LFSD}. These learned hyper-features still outperform the state-of-the-art approaches, while with LP we achieve almost 0.79, 0.79, and 0.84 F-measures. Figure \ref{fig:sal_LP} shows some examples of the optimized results of the existing methods (LMH \cite{peng2014rgbd}, ACSD \cite{ju2014depth}, and GP \cite{ren2015exploiting}) using Laplacian propagation.
These quantitative and qualitative experimental
evaluations further demonstrate that the proposed Laplacian propagation is able to refine the saliency maps of existing methods, which can be widely adopted as a post processing step.

\begin{figure}
\centering
\captionsetup[subfigure]{labelformat=empty}
\subfloat{ \label{fig:saliencyRGBvsRGBD:a}\includegraphics[width=\widtheighttwo,height=0.189\linewidth]{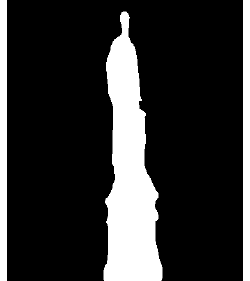}} \hspacefigure
\subfloat{ \label{fig:saliencyRGBvsRGBD:b}\includegraphics[width=\widtheighttwo,height=0.189\linewidth]{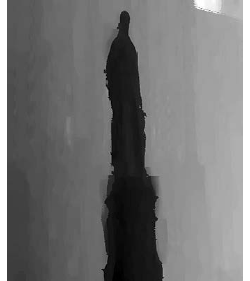}} \hspacefigure
\subfloat{ \label{fig:saliencyRGBvsRGBD:c}\includegraphics[width=\widtheighttwo,height=0.189\linewidth]{000905_left.pdf}} \hspacefigure
\subfloat{ \label{fig:saliencyRGBvsRGBD:d}\includegraphics[width=\widtheighttwo,height=0.189\linewidth]{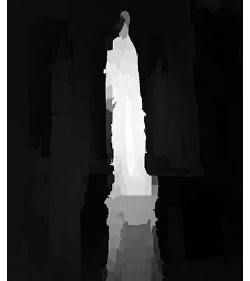}}\hspacefigure\\
\vspace{-1.5mm}
\subfloat[RGB]{ \label{fig:saliencyRGBvsRGBD:a}\includegraphics[width=\widtheighttwo,height=0.1859\linewidth]{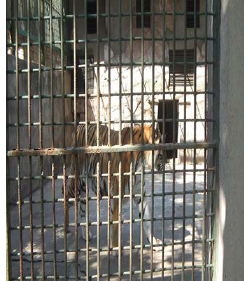}}  \hspace{-1mm}
\subfloat[Depth]{ \label{fig:saliencyRGBvsRGBD:b}\includegraphics[width=\widtheighttwo,height=0.1859\linewidth]{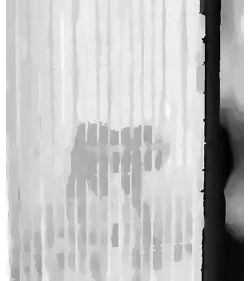}}\hspacefigure
\subfloat[GT]{ \label{fig:saliencyRGBvsRGBD:c}\includegraphics[width=\widtheighttwo,height=0.1859\linewidth]{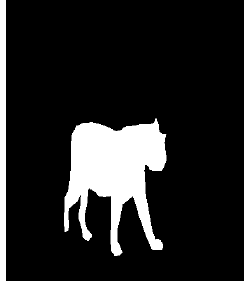}}\hspacefigure
\subfloat[Our result]{ \label{fig:saliencyRGBvsRGBD:d}\includegraphics[width=\widtheighttwo,height=0.1859\linewidth]{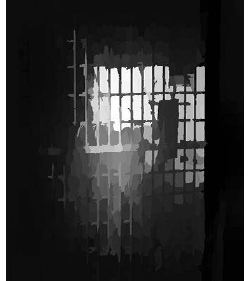}}\hspacefigure
   \caption{More visual results and some failure cases.}
    \label{fig:saliencyRGBvsRGBD} 
\end{figure}

\noindent{\textbf{Failure cases.}} Figure \ref{fig:saliencyRGBvsRGBD} gives more visual results and some failure cases of our proposed method on RGBD images.
Compared with the these two pictures, we can find that depth information is more helpful when the salient objects have high depth contrast with background or lie closer to the camera.
Our method may fail when the salient object shares a very similar color and depth information with the background.
\section{Conclusion}
In this paper, we propose a novel RGBD saliency detection method. Our framework consists of three different modules. The first module generates various low level saliency feature vectors
from the input image. The second module learns the interaction mechanism of RGB saliency features and depth-induced features and produces hyper-feature using CNN. Feeding with these hand-designed features can guide the learning process of CNN towards saliency-optimized. In the third module, we integrate a Laplacian propagation framework with CNN to obtain a spatially consistent saliency map. Both quantitative and qualitative experiment results show that the fused RGBD hyper-feature outperforms all the state-of-the-art methods.

We demonstrated that an optimized fusion leads to superior performance, and this flexible hyper-feature extraction framework can be further extended by including more saliency cues (e.g., flash cue \cite{shenfeng2014}). We aim to explore a deeper and more effective fusion network and extend it to other applications in our future work.

\begin{IEEEbiography}[{\includegraphics[width=1in,height=1.25in,clip,keepaspectratio]{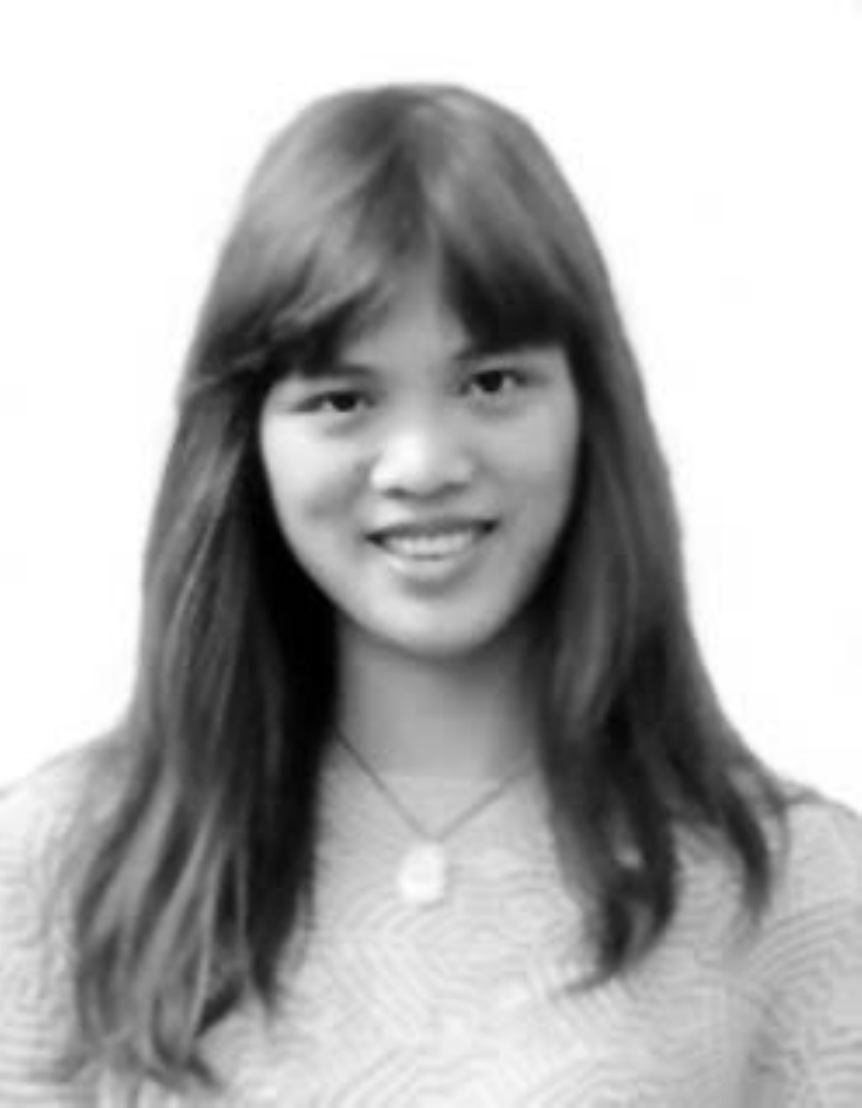}}]{Liangqiong Qu}
received the B.S. degree in automation from Central South University, China, in 2011. She is currently a joint Ph.D. student
of University of Chinese Academy of Sciences and City University of Hong Kong. Her research interests include illumination modeling, image processing, saliency detection and deep learning.
\end{IEEEbiography}

\begin{IEEEbiography}[{\includegraphics[width=1in,height=1.25in,clip,keepaspectratio]{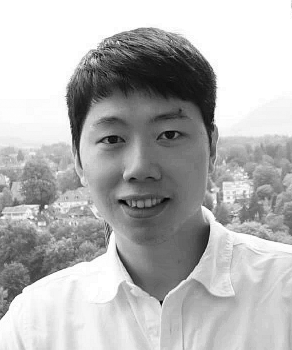}}]{Shengfeng He} obtained his B.Sc. degree and M.Sc. degree from Macau University of Science and Technology, and the Ph.D degree from City University of Hong Kong. He is currently a Research Fellow at City University of Hong Kong. His research interests include computer vision, image processing, computer graphics, and deep learning.
\end{IEEEbiography}

\begin{IEEEbiography}[{\includegraphics[width=1in,height=1.25in,clip,keepaspectratio]{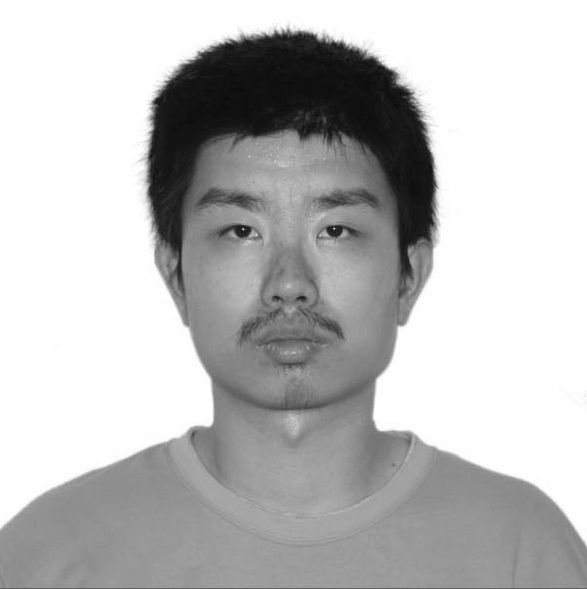}}]{Jiawei Zhang} received his BEng degree in Electronic Information Engineering from the University of
Science and Technology of China in 2011 and master degree in Institute of Acoustics, Chinese Academy of Sciences in 2014. He is currently a Computer Science PhD student in City university of Hong Kong.
\end{IEEEbiography}

\begin{IEEEbiography}[{\includegraphics[width=1in,height=1.25in,clip,keepaspectratio]{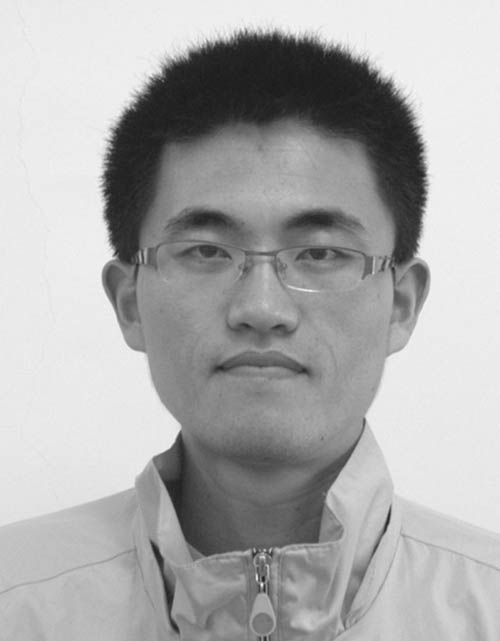}}]{Jiandong Tian}
received his B.S. Tech. degree in automation at Heilongjiang University, China, in 2005.
He received his Ph.D. degree in Pattern Recognition and Intelligent System at Chinese Academy of Sciences, China, in 2011.
He is currently an asassociate professor in computer vision at State Key Laboratory of Robotic, Shenyang Institute of Automation,
Chinese Academy of Sciences. His research interests include pattern recognition and robot vision.
\end{IEEEbiography}

\begin{IEEEbiography}[{\includegraphics[width=1in,height=1.25in,clip,keepaspectratio]{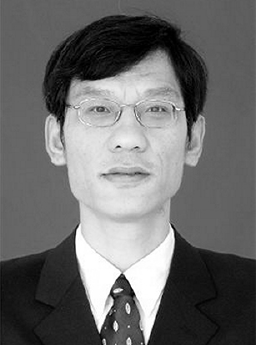}}]{Yandong Tang}
received B.S. and M.S. degrees in the department of mathematics, Shandong University in 1984 and 1987.
In 2002 he received the doctor's degree in applied mathematics from the University of Bremen, Germany. Currently he is a professor in Shenyang
Institute of Automation, Chinese Academy of Sciences. His research interests include robot vision, pattern recognition and numerical computation.
\end{IEEEbiography}

\begin{IEEEbiography}[{\includegraphics[width=1in,height=1.25in,clip,keepaspectratio]{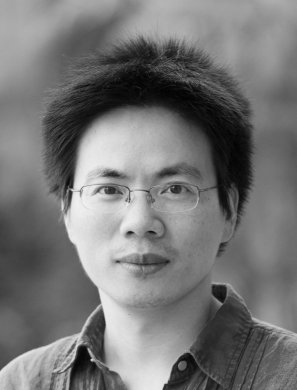}}]{Qingxiong Yang}
received the B.E. degree in
electronic engineering and information science from
the University of Science and Technology of China,
Hefei, China, in 2004, and the Ph.D. degree in electrical and computer engineering from the University
of Illinois at Urbana-Champaign, Champaign, IL,
USA, in 2010. He is currently an Assistant Professor
with the Department of Computer Science, City
University of Hong Kong, Hong Kong. His research
interests reside in computer vision and computer
graphics. He was a recipient of the Best Student
Paper Award at the 2010 International Workshop on Multimedia Signal
Processing and the Best Demo Award at the 2007 IEEE Computer Society
Conference on Computer Vision and Pattern Recognition.
\end{IEEEbiography}

\end{document}